\documentclass[11pt]{article}
\usepackage[utf8]{inputenc}
\pdfoutput=1

\usepackage{amssymb,amsmath}
\usepackage[utf8]{inputenc}
\usepackage{graphicx,url}
\usepackage{subfig}
\usepackage{lscape}
\usepackage{rotating}
\usepackage[margin=1in]{geometry}

\usepackage{booktabs}
\usepackage{tabularx}
\usepackage{float}
\usepackage{verbatim}
\usepackage{authblk}
\usepackage{minitoc}
\usepackage{amsmath}
\usepackage{amsfonts}
\usepackage{amssymb}
\usepackage{mathtools}

\usepackage{subfiles}
\usepackage{multirow}

\usepackage[numbers]{natbib}
\usepackage[american]{babel}
\usepackage{csquotes}

\usepackage{etoolbox}

\usepackage{url}
\usepackage[colorinlistoftodos]{todonotes}
\usepackage[utf8]{inputenc}
\usepackage{soul}
\usepackage{parskip}

\makeatletter
\renewcommand*{\numberline}[1]{\hb@xt@3em{#1\hfil}} 
\makeatother

\usepackage{hyperref}
\hypersetup{
    bookmarksopen,
    bookmarksdepth=2,
    breaklinks=true
    unicode=false,          % non-Latin characters in Acrobat's bookmarks
    pdftoolbar=true,        % show Acroba's toolbar?
    pdfmenubar=true,        % show Acrob's menu?
    pdffitwindow=true,     % window fit to page when opened
    pdfstartview={FitH},    % fits the width of the page to the window
    pdftitle={},    % title
    pdfauthor={},     % author
    pdfsubject={},   % subject of the document
    pdfcreator={},   % creator of the document
    pdfkeywords={}, % list of keywords
    pdfnewwindow=true,      % links in new window
    colorlinks=true,       % false: boxed links; true: colored links
    linkcolor=[rgb]{0,0,1},           % color of internal links
    citecolor=[rgb]{0,0,0},        % color of links to bibliography
    filecolor=[rgb]{0.35,0.35,0.35},      % color of file links
    urlcolor=[rgb]{0.35,0.35,0.35}           % color of external links
}

\newcommand{\methodname}{MOSAIKS}
\newcommand{\beq}{\begin{equation}}
\newcommand{\eeq}{\end{equation}}

\newcommand{\bi}{\begin{itemize}}
\newcommand{\ei}{\end{itemize}}
\newcommand{\mb}{\mathbf}

\newcommand{\RR}{\mathbb{R}}

\newcommand{\conv}{*}
\newcommand{\reLu}{\textrm{ReLU}}
\newcommand{\X}{\mb{X}}
\newcommand{\x}{\mb{x}}
\newcommand{\I}{\mb{I}}
\newcommand{\Amat}{\mb{A}}
\newcommand{\Pmat}{\mb{P}}

\newcommand{\Ak}{\Amat_k}
\newcommand{\Pk}{\Pmat_k}

\newcommand{\xk}{\x_{k}}

\newcommand{\Iell}{\I_\ell}
\newcommand{\Xell}{\x_\ell}
\newcommand{\XIell}{\x(\Iell)}
\newcommand{\XkIell}{\xk(\Iell)}
\newcommand{\AkIell}{\Ak(\Iell)}

\usepackage{bm}

\usepackage[toc,page,header]{appendix}
\usepackage{minitoc}

\usepackage{titling}

\newtoggle{arxiv}
\toggletrue{arxiv}

\makeatletter
\renewcommand*{\@fnsymbol}[1]{\ifcase#1\or$\ast$\else\@arabic{\numexpr#1-1\relax}\fi}

\title{A Generalizable and Accessible Approach to Machine Learning with Global Satellite Imagery}

\author {Esther Rolf\hspace{.1em}\thanks{Equal contribution}\hspace{.1em}
\thanks{Electrical Engineering \& Computer Science Department, UC Berkeley. esther\_rolf@berkeley.edu}, \, 
Jonathan Proctor$^{\ast}$\thanks{Center for the Environment and Data Science Initiative, Harvard University. jproctor1@fas.harvard.edu}, \,
Tamma Carleton$^{\ast}$\thanks{Bren School of Environmental Science \& Management, UC Santa Barbara. tcarleton@ucsb.edu} ,\,
Ian Bolliger$^{\ast}$\thanks{Rhodium Group. ibolliger@rhg.com} ,\,
Vaishaal Shankar$^{\ast}$\thanks{Electrical Engineering \& Computer Science Department, UC Berkeley. vaishaal@berkeley.edu},\\
Miyabi Ishihara\thanks{Statistics Department, UC Berkeley. miyabi\_ishihara@berkeley.edu} ,
Benjamin Recht\thanks{Electrical Engineering \& Computer Science Department, UC Berkeley. brecht@berkeley.edu} ,
 Solomon Hsiang\thanks{Global Policy Laboratory, Goldman School of Public Policy, UC Berkeley. shsiang@berkeley.edu}
}

\begin{document}

\doparttoc % Tell to minitoc to generate a toc for the parts
\faketableofcontents % Run a fake tableofcontents command for the partocs

\renewcommand \thepart{}
\renewcommand \partname{}

%\part*{}
%\addcontentsline{toc}{part}{} % Start the document part
%\parttoc % Insert the document TOC

%\baselineskip24pt

\maketitle

%%%%%%%%%%%% MAIN %%%%%%%%%%%%%

\iftoggle{arxiv}{
% arxiv version
\begin{abstract}
Combining satellite imagery with machine learning (SIML) has the potential to address global challenges by remotely estimating socioeconomic and environmental conditions in data-poor regions, yet the resource requirements of SIML limit its accessibility and use. We show that a single encoding of satellite imagery can generalize across diverse prediction tasks (e.g. forest cover, house price, road length). Our method achieves accuracy competitive with deep neural networks at orders of magnitude lower computational cost, scales globally, delivers label super-resolution predictions, and facilitates characterizations of uncertainty. Since image encodings are shared across tasks, they can be centrally computed and distributed to unlimited researchers, who need only fit a linear regression to their own ground truth data in order to achieve state-of-the-art SIML performance.
\end{abstract}
}{ 
% journal submission
\begin{sciabstract}
Combining satellite imagery with machine learning (SIML) has the potential to address global challenges by remotely estimating socioeconomic and environmental conditions in data-poor regions, yet the resource requirements of SIML limit its accessibility and use. We show that a single encoding of satellite imagery can generalize across diverse prediction tasks (e.g. forest cover, house price, road length). Our method achieves accuracy competitive with deep neural networks at orders of magnitude lower computational cost, scales globally, delivers label super-resolution predictions, and facilitates characterizations of uncertainty. Since image encodings are shared across tasks, they can be centrally computed and distributed to unlimited researchers, who need only fit a linear regression to their own ground truth data in order to achieve state-of-the-art SIML performance.
\end{sciabstract}
\newpage
\linenumbers
}

%\iftoggle{arxiv}{\tableofcontents}{}

\iftoggle{arxiv}{\section{Introduction}}{\section*{Introduction}}

Addressing complex global challenges---such as managing global climate changes, population movements, ecosystem transformations, or economic development---requires that many different researchers and decision-makers (hereafter \emph{users}) have access to reliable, large-scale observations of many variables simultaneously. Planet-scale ground-based monitoring systems are generally prohibitively costly for this purpose, but satellite imagery presents a viable alternative for gathering globally comprehensive data, with over 700 earth observation satellites currently in orbit \cite{scientists_2019}. Further, application of machine learning is proving to be an effective approach for transforming these vast quantities of unstructured imagery data into structured estimates of ground conditions. For example, combining satellite imagery and machine learning (SIML) has enabled better characterization of forest cover \cite{hansen2013}, land use \cite{Inglada2017}, poverty rates \cite{Jean2016} and population densities \cite{Robinson2017}, thereby supporting research and decision-making. We refer to such predictions of an individual variable as a single \textit{task}. Demand for SIML-based estimates is growing, as indicated by the large number of private service-providers specializing in predicting one or a small number of these tasks. 

The resource requirements for deploying SIML technologies, however, limit their accessibility and usage. Satellite-based measurements are particularly under-utilized in low-income contexts, where the technical capacity to implement SIML may be low, but where such measurements would likely convey the greatest benefit \cite{Yu2014,Haack2016}.
For example, government agencies in low-income settings might want to understand local waterway pollution, illegal land uses, or mass migrations. SIML, however, remains largely out of reach to these and other potential users because current approaches require a major resource-intensive enterprise, involving a combination of task-specific domain knowledge, remote sensing and engineering expertise, access to imagery, customization and tuning of sophisticated machine learning architectures, and large computational resources.

To remove many of these barriers, we develop a new approach to SIML that enables non-experts to obtain state-of-the-art performance without manipulating imagery, using specialized computational resources, or developing a complex prediction procedure. We design a one-time, task-agnostic encoding that transforms each satellite image into a vector of variables (hereafter \textit{features}). We then show that these features ($x$) perform well at predicting ground conditions ($y$) across diverse  tasks, using only a linear regression implemented on a personal computer. Prior work has similarly sought an unsupervised encoding of satellite imagery \cite{Romero2016,Cheriyadat2014,penatti2015deep,Jean2019}; however, to the best of our our knowledge, we are the first to demonstrate that a single set of features both achieves performance competitive with deep-learning methods across a variety of tasks and scales globally.

We focus here on the problem of predicting properties of small regions (e.g. {average house price}), using high-resolution daytime satellite imagery as the only input. We develop a simple yet high-performing system that is tailored to address the challenges and opportunities specific to SIML applications, taking a fundamentally different approach from leading designs. We achieve large computational gains in model training and testing, relative to leading deep neural networks, through algorithmic simplifications that take advantage of the fact that satellite images are collected from a fixed distance and viewing angle and capture repeating patterns and objects. This contrasts with deep-learning approaches to SIML that use techniques originally developed for natural images (e.g. photos taken from handheld cameras), where inconsistency in many key factors, such as subject or camera perspective, require complex solutions that our results suggest are mostly unnecessary for SIML applications.

A key contribution of our analysis is the demonstration that a single set of general purpose features can encode rich information in satellite images. We utilize an unsupervised learning methodology, which separates feature construction from model-fitting. This approach dramatically increases computational speed for any given researcher and delivers large computational gains at the research-system level by reorganizing how imagery is processed and distributed. Traditionally, hundreds or thousands of researchers use the same images to solve different and unrelated tasks (e.g. Fig.~\ref{Fig:Cartoon}A). Our approach allows common sources of imagery to be converted into one centralized  set of features that can be accessed by many researchers, each solving different tasks. This isolates future users from the  costly steps of obtaining, storing, manipulating, and processing imagery themselves. The magnitude of the resulting benefits grow with the size of the expanding SIML user community and the scale of global imagery data, which currently increases by more than $80$TB/day~\cite{littlepage}.

\iftoggle{arxiv}{
\begin{figure}
\centering
\includegraphics[width=\textwidth]{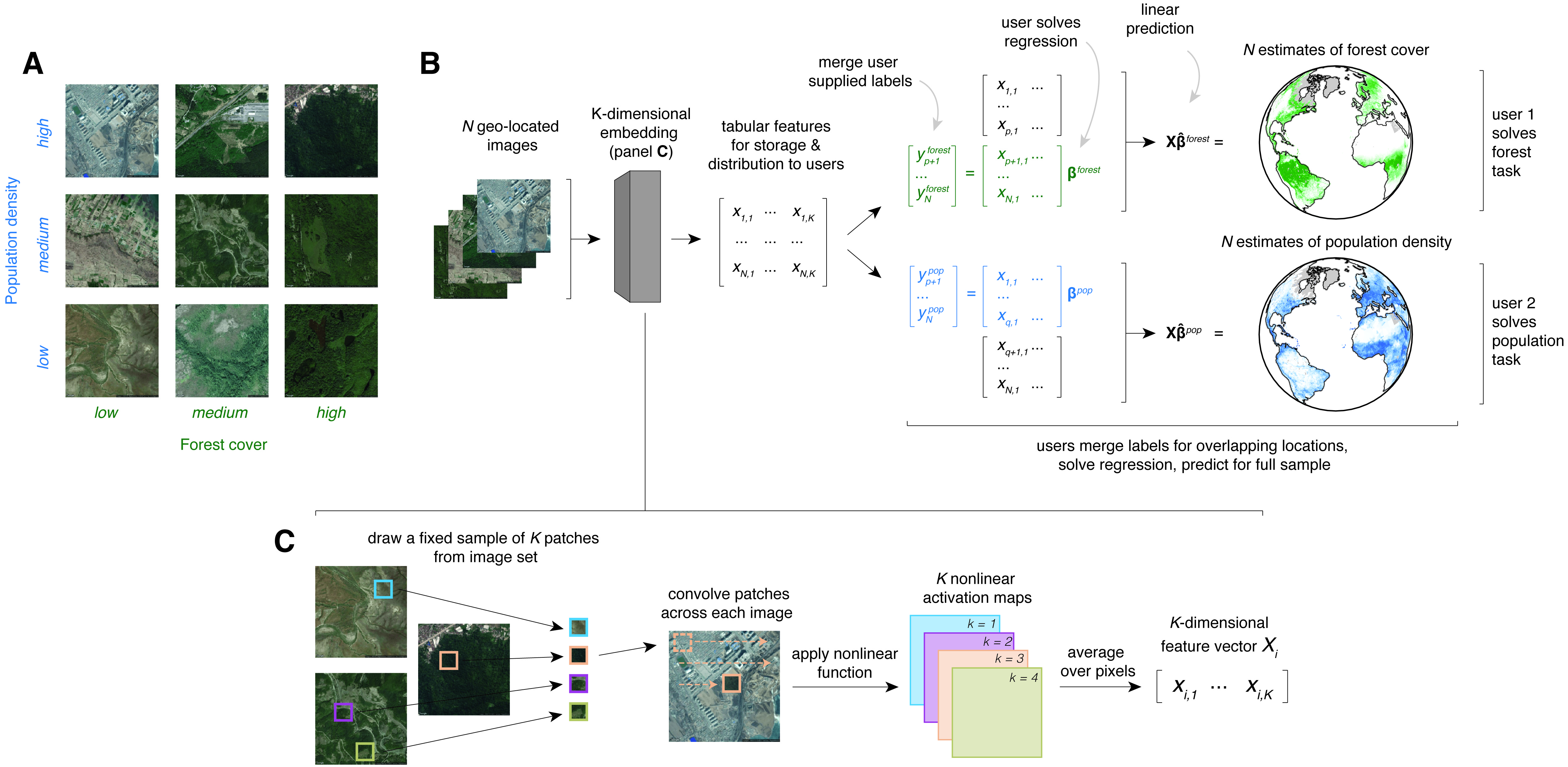}
\caption{\footnotesize \textbf{A generalizable approach to combining satellite imagery with machine learning (SIML) without users handling images.}
\methodname\ is designed to solve an unlimited number of tasks at planet-scale quickly. After a one-time unsupervised image featurization using random convolutional features, \methodname\ centrally stores and distributes task-agnostic features to users, each of whom generates predictions in a new context.
(A) Satellite imagery is shared across multiple potential tasks. For example, nine images from the US sample are ordered based on population density and forest cover, both of which have distinct identifying features that are observable in each image.  (B) Schematic of the \methodname\ process. $N$ images are transformed using random convolutional features into a compressed and highly descriptive $K$-dimensional feature vector before labels are known.  Once features are computed, they can be stored in tabular form (matrix $\textbf{X}$) and used for unlimited tasks without recomputation. Users interested in a new task ($s$) merge their own labels ($y^s$) to features for training. Here, \textit{User 1} has forest cover labels for locations $p+1$ to $N$ and \textit{User 2} has population density labels for locations $1$ to $q$. Each user then solves a single linear regression for $\bm\beta^s$. Linear prediction using $\bm\beta^s$ and the full sample of \methodname\ features $\textbf{X}$ then generates SIML estimates for label values at all locations. Generalizability allows different users to solve different tasks using an identical procedure and the same table of features---differing only in the user-supplied label data for training. Each task can be solved by a user on a desktop computer in minutes without users ever manipulating the imagery. (C) Illustration of the one-time unsupervised computation of random convolutional features (Supplementary Materials Sections \ref{sec:main-methods} and \ref{sec:featurization}). $K$ patches are randomly sampled from across the $N$ images. Each patch is convolved over each image, generating a nonlinear activation map for each patch. Activation maps are averaged over pixels to generate a single $K$-dimensional feature vector for each image.}
\label{Fig:Cartoon}
\end{figure}}{% non-arxiv version figures at the end
}

\iftoggle{arxiv}{\section{Multi-task Observation using Satellite Imagery \& Kitchen Sinks}}
{\section*{Multi-task Observation using Satellite Imagery \& Kitchen Sinks}}

Our objective is to enable any user with basic resources to predict ground conditions using only satellite imagery and a limited sample of task-specific ground truth data which they possess. Our SIML system, ``Multi-task Observation using Satellite Imagery and Kitchen Sinks'' (\methodname, see Supplementary Materials \ref{sec:main-methods}), makes SIML accessible and generalizable by separating the prediction procedure into two independent steps: a fixed ``featurization step'' which translates satellite imagery into succinct vector representations ($images \rightarrow x$), and a ``regression step'' which learns task-specific coefficients that map these features to outcomes for a given task ($x\rightarrow y$). The unsupervised featurizaton step can be centrally executed once, producing one set of outputs that are used to solve many different tasks through repeated application of the regression step by multiple independent users (Fig.~\ref{Fig:Cartoon}B). Because the regression step is computationally efficient,  \methodname\ scales nearly costlessly across unlimited users and tasks.

The \emph{accessibility} of our approach stems from the simplicity and computational efficiency of the regression step for potential users, given features which are already computed once and stored centrally (Fig.~\ref{Fig:Cartoon}B).
To generate SIML predictions, a user of \methodname\ (i) queries these tabular data for a vector of $K$ features for each of their $N$ locations of interest; (ii) merges these features $x$ with label data $y$, i.e. the user's independently collected ground truth data; (iii) implements a linear regression of $y$ on $x$ to obtain coefficients $\beta$ -- below, we use ridge regression; (iv) uses coefficients $\beta$ and and features $x$ to predict labels $\hat{y}$ in new locations where imagery and features are available but ground truth data are not.

The \emph{generalizability} of our approach means that a single mathematical summary of satellite imagery ($x$) performs well across many prediction tasks ($y_1, y_2,...$) without any task-specific modification to the procedure. The success of this generalizability relies on how images are encoded as features. We design a featurization function by building on the theoretically grounded machine learning concept of ``random kitchen sinks'' \cite{Rahimi2008}, which we apply to satellite imagery by constructing ``random convolutional features" (RCFs) (Fig. \ref{Fig:Cartoon}C, Supplementary Materials \ref{sec:main-methods}). RCFs are suitable for the structure of satellite imagery and have established performance encoding genetic sequences \cite{Morrow2017}, classifying photographs \cite{Coates2012}, and predicting solar flares\cite{Jonas2018} (see Supplementary Materials Section \ref{sec:featurization}). RCFs capture a flexible measure of similarity between every sub-image across every pair of images without using contextual or task-specific information. The regression step in \methodname\ then treats these features $x$ as an overcomplete basis for predicting any $y$, which may be a nonlinear function of image elements (see Supplementary Materials Section \ref{sec:main-methods}).

In contrast to many recent alternative approaches to SIML, \methodname\ does not require training or using the output of a deep neural network and encoding images into unsupervised features requires no labels. 
Nonetheless, \methodname\ achieves competitive performance at a large computational advantage that grows linearly with the number of SIML users and tasks, due to shared computation and storage. In principle, any unsupervised featurization would enable these computational gains. However, to date, a single set of unsupervised features has neither achieved accuracy competitive with supervised CNN-based approaches across many SIML tasks, nor at the scale that we study. Below, we show that \methodname\ achieves a practical level of generalization in real world contexts.

\iftoggle{arxiv}{\section{Results}}{\section*{Results}}

We design a battery of experiments to test whether and under what settings \methodname\ can provide access to high-performing, computationally-efficient, global-scale SIML predictions. Specifically, we 1) demonstrate generalization across tasks, and compare \methodname's performance and cost to existing state-of-the-art SIML models; 2) assess its performance when data are limited and when predicting far from observed labels; 3) scale the analysis to make global predictions and try recreating the results of a national survey; and 4) detail additional properties of \methodname, such as the ability to make predictions at finer resolution than the provided labels. 
 
\iftoggle{arxiv}{\subsection{Generalization across tasks}}{
\subsubsection*{Generalization across tasks}}

\iftoggle{arxiv}{
\begin{figure}
\vspace{-2.5cm}
\centering
\includegraphics[height=.9\textheight]{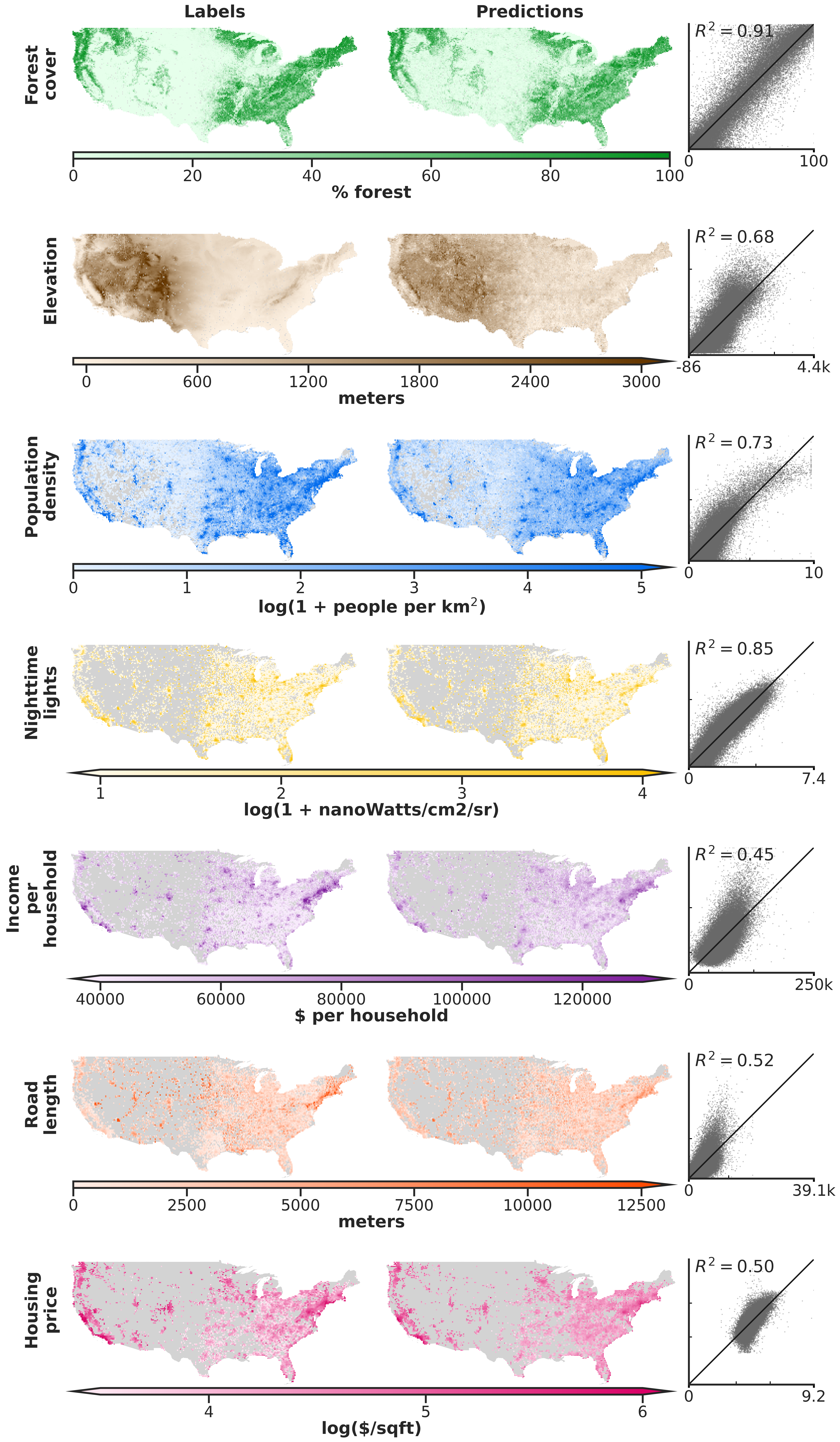}
\caption{\footnotesize \textbf{1km $\times$ 1km resolution prediction of many tasks across the continental US using daytime images processed once, before tasks were chosen.}  100,000 daytime images were each converted to 8,192 features and stored. Seven tasks were then selected based on coverage and diversity and predictions were generated for each task using the same procedure. Left maps: 80,000 observations used for training and validation, aggregated up to 20km$\times$20km cells for display. Right maps: 
concatenated validation set estimates from 5-fold cross-validation for the same 80,000 grid cells (observations are never used to generate their own prediction), identically aggregated for display. Scatters: Validation set estimates (vertical axis) vs. ground-truth (horizontal axis); each point is a $\sim$1km$\times$1km grid cell. Black line is at $45^\circ$. Test set and validation set performance are essentially identical (Table \ref{SI:testSetTable}); validation set values are shown for display purposes only, as there are more observations. The tasks in the top three rows are uniformly sampled across space, the tasks in the bottom four rows are sampled using population weights (Supplementary Materials Section \ref{sec:grid_and_sampling}); grey areas were not sampled in the experiment.}
\label{Fig:MapArray}
\end{figure}}{}

We first test whether \methodname\ achieves a practical level of generalization by applying it to a diverse set of pre-selected tasks in the United States (US). 
While many applications of interest for SIML are in remote and/or data-limited environments where ground-truth may be unavailable or inaccurate, systematic evaluation and validation of SIML methods are most reliable in well-observed and data-rich environments \cite{Blumenstock2018}.

\iftoggle{arxiv}{\subsubsection{Multi-task performance of \methodname\ in the US}}
{\paragraph{Multi-task performance of \methodname\ in the US}} We sample daytime images 
\iftoggle{arxiv}{}{using the Google Static Maps API} from across the continental US ($N=100,000$), each covering $\sim$1km$\times$1km (256-by-256 pixels) (Supplementary Materials Sections \ref{sec:grid_and_sampling}-\ref{sec:label_agg}). We first implement the featurization step, passing these images through MOSAIKS' feature extraction algorithm to produce $K=8,192$ features per image (Supplementary Materials Section \ref{sec:featurization}). Using only the  resulting matrix of features ($\mb{X}$), we then repeatedly implement the regression step by solving a cross-validated ridge regression for each task and predict
forest cover ($R^2=0.91$), 
elevation ($R^2=0.68$), 
population density ($R^2=0.72$), 
nighttime lights ($R^2=0.85$), 
average income ($R^2=0.45$), 
total road length ($R^2=0.53$), 
and average house price ($R^2=0.52$)\footnote{Performance observed for housing using our published data will be higher ($R^2=0.60$) because privacy concerns mandate the withholding of a subset of this data (see Supplementary Materials Section~\ref{sec:labeldata}).} in a holdout test sample (Fig.~\ref{Fig:MapArray}, Table \ref{SI:testSetTable}, Supplementary Materials Sections \ref{sec:data_separation_uncertainty}-\ref{sec:test_set_performance}). Computing the feature matrix $\mathbf{X}$ from imagery took less than 2 hours on a cloud computing node (Amazon EC2 p3.2xlarge instance, Tesla V100 GPU).
Subsequently, solving a cross-validated ridge regression for each task took 6.8 minutes to compute on a local workstation with ten cores (Intel Xeon CPU E5-2630) (Supplementary Materials Section~\ref{sec:cost_analysis}). These seven outcomes are not strongly correlated with one another (Fig.~\ref{fig:corrys}) and no attempted tasks in this experiment are omitted. These results indicate that \methodname\ is skillful for a wide range of possible applications without changing the procedure or features and without task-specific expertise. 

\iftoggle{arxiv}{\subsubsection{Comparison to state-of-the-art SIML approaches}}{\paragraph{Comparison to state-of-the-art SIML approaches}} We contextualize this performance by comparing \methodname\ to existing deep-learning based SIML approaches.  First, we retrain end-to-end a commonly-used deep convolutional neural network (CNN) architecture\cite{He2016} (ResNet-18) using identical imagery and labels for the seven tasks above. This training took 7.9 hours per task on a cloud computing node (Amazon EC2 p3.xlarge instance, Tesla V100 GPU). We find that \methodname\ exhibits predictive accuracy competitive with the CNN for all seven tasks (mean $R^2_{CNN}-R^2_{MOSAIKS}= 0.04$; smallest $R^2_{CNN}-R^2_{MOSAIKS}= -0.03$ for housing; largest $R^2_{CNN}-R^2_{MOSAIKS}= 0.12$ for elevation) in addition to being approximately $250$ to $10,000 \times$ faster to train, depending on whether the regression step is performed on a laptop (2018 Macbook Pro) or on the same cloud computing node used to train the CNN (Fig.~\ref{Fig:Waffle}A, Supplementary Materials Section~\ref{sec:cnn_benchmark} and Table~\ref{table:cost_table}). 

Second, we apply ``transfer learning''  \cite{Pan2010} using the ResNet-152 CNN pre-trained on natural images to featurize the same satellite images. We then apply ridge regression to the CNN-derived features. The speed of this approach is similar to \methodname, but its performance is dramatically lower on all seven tasks (Fig.~\ref{Fig:Waffle}A, Supplementary Materials Section~\ref{sec:cnn_benchmark}). 

Third, we compare \methodname\ to an approach from prior studies\cite{Jean2016,xie2016transfer,Head2017} where a deep CNN (VGG16 \cite{Simonyan2015} pretrained on the ImageNet dataset) is trained end-to-end on night lights and then each task is solved via transfer learning (Supplementary Materials Section \ref{sec:cnn_benchmark}). We apply \methodname\ to the imagery from Rwanda, Haiti, and Nepal used in ref. \cite{Head2017} to solve all eleven development-oriented tasks they analyze. We find \methodname\ matches prior performance across tasks in Rwanda and Haiti, and has slightly lower performance (average $\Delta R^2= 0.08$) on tasks in Nepal (Fig. \ref{fig:headRep}). The regression step of this transfer learning approach and \methodname\ are similarly fast, but the transfer learning approach requires country-specific retraining of the CNN, limiting its accessibility and reducing its generalizability. 

Together, these three experiments illustrate that with a single set of task-independent features, \methodname\ predicts outcomes across a diverse set of tasks, with performance and speed that favorably compare to existing SIML approaches.  However, throughout this set of experiments, we find that some sources of variation in labels are not recovered by \methodname.  For example, extremely high elevations ($>$3,000m) are not reliably distinguished from high elevations (2,400-3,000m) that appear visually similar (Fig. \ref{fig:us_error_scatter}). Additionally, roughly half the variation in incomes and housing prices is unresolved, presumably because they depend on factors not observable from orbit, such as tax policies or school districts (Fig.~\ref{Fig:MapArray}). We also show that patterns of predictability across tasks are strikingly similar across \methodname\ and alternative SIML approaches (Figs. \ref{fig:cnn_comparison} and \ref{fig:headRep}). Together, these findings are consistent with the hypothesis that there exists some performance ceiling for each task, due to some factors not being observable from satellite imagery.

\iftoggle{arxiv}{\subsection{Evaluation of model sensitivity}}{
\subsubsection*{Evaluation of model sensitivity}}

\iftoggle{arxiv}{
\begin{figure}
\centering
\vspace{-1.5cm}
\includegraphics[width = .9\textwidth]{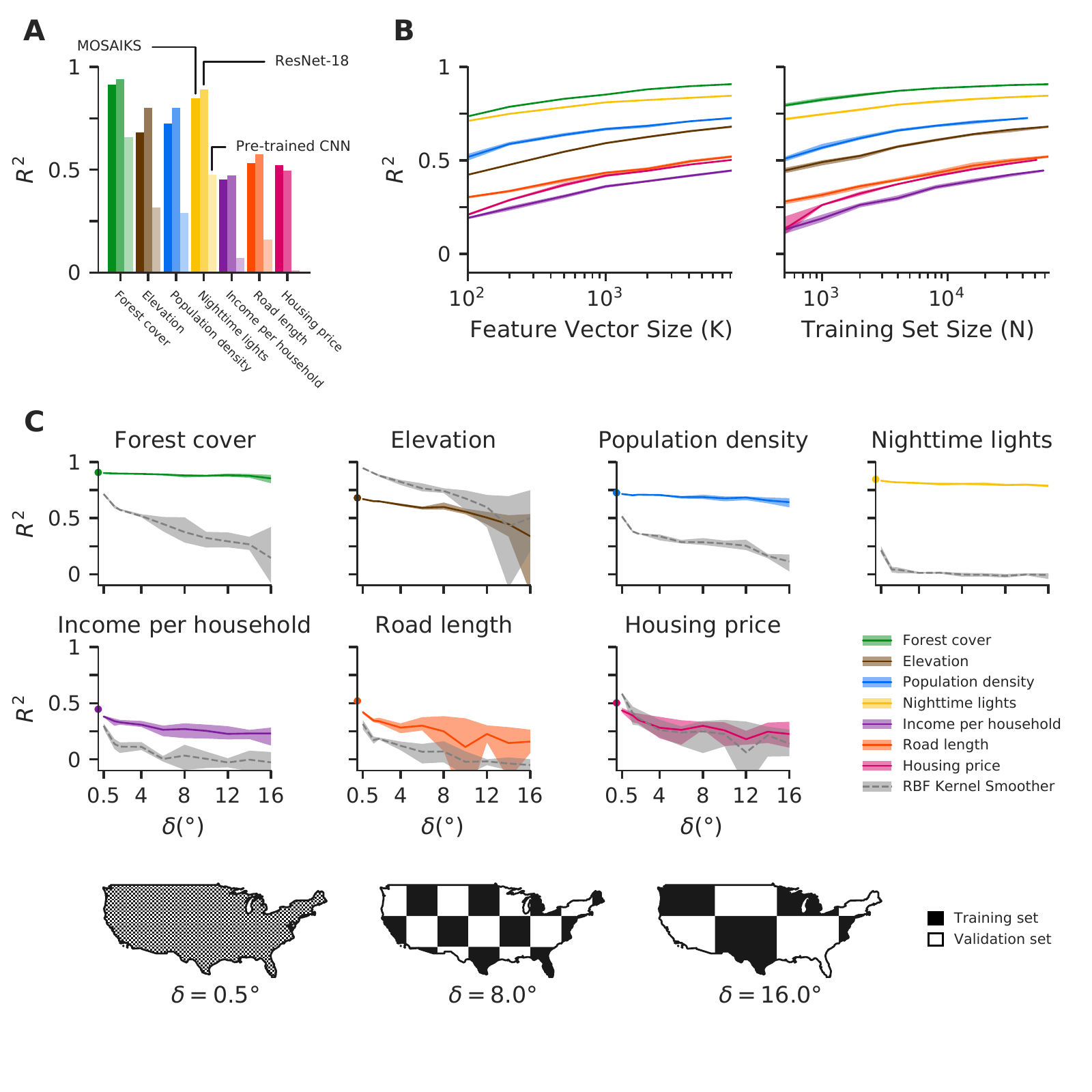}
\caption{\footnotesize \textbf{Prediction accuracy relative to a convolutional neural network and transfer learning, using smaller $K$ and $N$, and over large contiguous regions with no ground truth data.} (A) Task-specific \methodname\ test-set performance (dark bars) in contrast to: an 18-layer variant of the ResNet Architecture (ResNet-18) trained end-to-end for each task (middle bars); and an unsupervised featurization using the last hidden layer of a 152-layer ResNet variant pre-trained on natural imagery and applied using ridge regression (lightest bars). See Supplementary Materials Section \ref{sec:cnn_benchmark} for details. (B) Validation set $R^2$ performance for all seven tasks while varying the number of random convolutional features $K$ and holding $N=64,000$ (left) and while varying $N$ and holding $K=8,192$ (right). Shaded bands indicate the range of predictive skill across 5 folds. Lines indicate average accuracy across folds. (C) Evaluation of performance over regions of increasing size that that are excluded from training sample. Data are split using a ``checkerboard'' partition, where the width and height of each square is $\delta$ (measured in degrees). Example partitions with $\delta=0.5^\circ$, $8^\circ$, 16$^\circ$ are shown in maps. For a given $\delta$, training occurs using data sampled from ``black squares'' and performance is evaluated in ``white squares.'' Plots show colored lines representing average performance of \methodname\ in the US across $\delta$ values for each task. Benchmark performance from Fig.~\ref{Fig:MapArray} are indicated as circles at $\delta=0$. Grey dashed lines indicate corresponding performance using only spatial interpolation with an optimized radial basis function kernel instead of \methodname\ (Supplementary Materials Section \ref{sec:secondary_analysis_spatial}). To moderate the influence of the exact placement of square edges, training and test sets are resampled four times for each $\delta$ with the checkerboard position re-initialized using offset vertices (see Supplementary Materials Section \ref{sec:secondary_analysis_spatial} and Fig.~\ref{fig:jitter_explanation}). The ranges of performance are plotted as colored or grey bands.}
\label{Fig:Waffle}
\end{figure}}{}

There is growing recognition that understanding the accuracy, precision, and limits of SIML predictions is important, since consequential decisions increasingly depend on these outputs, such as which households should receive financial assistance~\cite{Athey2017b,Blumenstock2018}. However, historically, the high costs of training deep-learning models have generally prevented the stress-testing and bench-marking that would ensure accuracy and constrain uncertainty. To characterize the performance of \methodname\, we test its sensitivity to the number of features ($K$) and training observations ($N$), as well as  the extent of spatial extrapolation.

\iftoggle{arxiv}{\subsubsection{Changes to training data}}{\paragraph{Changes to training data}}
 Unlike some featurization methods, these is no measure of ``importance'' for individual features in \methodname\, so the computational complexity of the regression step can be manipulated by simply including more or fewer features. Repeatedly re-solving the linear regression step in \methodname\ with a varied number of features indicates that increasing $K$ above 1,000 features provides minor predictive gains (Fig.~\ref{Fig:Waffle}B). A majority of the observable signal in the baseline experiment using $K=8,192$ is recovered using $K=200$ (min 55\% for income, max 89\% for nighttime lights), reducing each 65,536-pixel tri-band image to just 200 features ($\sim250\times$ data compression). 
% vary N
Similarly, re-solving \methodname\ predictions with a different number of training observations demonstrates that models trained with fewer samples may still exhibit high accuracy (Fig.~\ref{Fig:Waffle}B). A majority of the available signal is recovered for many outcomes using only $N=500$ (55\% for road length to 87\% for forest cover), with the exception of income (28\%) and housing price (26\%) tasks, which require larger samples. Together, these experiments suggest that users with computational, data acquisition, or data storage constraints can easily tailor \methodname\ to match available resources and can reliably estimate the performance impact of these alterations (Supplementary Material Section \ref{sec:secondary_analysis_nk}). 

\iftoggle{arxiv}{\subsubsection{Spatial cross-validation}}{\paragraph{Spatial cross-validation}}
To systematically evaluate the ability of \methodname\ to make accurate predictions in large contiguous areas where labels are not available, we conduct a spatial cross-validation experiment by partitioning the US into a checkerboard pattern (Fig.~\ref{Fig:Waffle}C), training on the ``black squares'' and testing on the ``white squares'' (Supplementary Materials Section \ref{sec:secondary_analysis_spatial}). Increasing the width of squares ($\delta$) in the checkerboard increases the average distances between train and test observations, simulating increasingly large spatial extrapolations. We find that for three of seven tasks (forest cover, population density, and nighttime lights), performance declines minimally regardless of distance (maximum $R^2$ decline of 10\% at $\delta = 16^\circ$ for population density). For income, road length, and housing price, performance falls moderately at small degrees of spatial extrapolation (19\%, 33\%, and 35\% decline at $\delta =4^\circ$, respectively), but largely stabilizes thereafter. Note that the poor performance of road length predictions is possibly due to missing labels and data quality (Supplementary Materials Section \ref{par:road_data} and Fig. \ref{fig:private_roads}). Finally, elevation exhibits steady decline with increasing distances between training and testing data (49\% decline at $\delta = 16^\circ$).

To contextualize this performance, we compare \methodname\ to spatial interpolation of observations, a widely used approach to fill in regions of missing data (Supplementary Materials Section \ref{sec:secondary_analysis_spatial}). Using the same samples, \methodname\ substantially outperforms spatial interpolation (Fig.~\ref{Fig:Waffle}C, grey dashed lines) across all tasks except for elevation, where interpolation performs almost perfectly over small ranges ($\delta = 0.5^\circ: R^2=0.95$), and housing price, where interpolation slightly outperforms \methodname\ at small ranges. For both, interpolation performance converges to that of \methodname\ over larger distances. Thus, in addition to generalizing across tasks, \methodname\ generalizes out-of-sample across space, outperforming spatial interpolation of ground-truth in 5 of 7 tasks.

The above sensitivity tests are enabled by the speed and simplicity of training \methodname. These computational gains also enable quantification of uncertainty in model performance within each diagnostic test. As demonstrated by the shaded bands in Figs.~\ref{Fig:Waffle}B-C, uncertainty in \methodname\ performance due to variation in splits of training-validation data remains modest under most conditions.

\iftoggle{arxiv}{\subsection{Applications}}{\subsubsection*{Applications}}
Having evaluated \methodname\ systematically in the data-rich US, we test its performance at planetary scale and its ability to recreate results from a national survey. 

\iftoggle{arxiv}{\subsubsection{Global observation}}{\paragraph{Global observation}}
We test the ability of \methodname\ to scale globally using the four tasks for which global labels are readily available. Using a random sub-sample of global land locations (training and validation: $N=$ 338,781, test: $N=$ 84,692; Supplementary Materials Section~\ref{sec:global_analysis}), we construct the first planet-scale, multi-task estimates using a single set of label-independent features ($K=2048$, Fig.~\ref{Fig:Global}A), predicting the distribution of forest cover ($R^2=0.85$), elevation ($R^2=0.45$), population density ($R^2=0.62$), and nighttime lights ($R^2=0.49$). Note that inconsistent image and label quality across the globe are likely partially responsible for lowering performance relative to the US-only experiments above (Supplementary Materials Section~\ref{sec:global_analysis}). 

\iftoggle{arxiv}{\subsubsection{National survey ``field test''}}{\paragraph{National survey ``field test''}}
It has been widely suggested that SIML could be used by resource-constrained governments to reduce the cost of surveying their citizens \cite{Jean2016, Head2017, Reed2018, Bedi2007, DeSherbinin2017}. To demonstrate \methodname's performance in this theoretical use-case, we simulate a `field test' with the goal of recreating results from an existing nationally representative survey. Using the pre-computed features from the first US experiment above, we generate predictions for 12 pre-selected questions in the 2015 American Community Survey (ACS) conducted by the US Census Bureau \cite{acs-income}. We obtain $R^2$ values ranging from 0.06 (\textit{percent household income spent on rent}, an outlier) to 0.52 (\textit{building age}), with an average $R^2$ of 0.34 across 12 tasks (Fig.~\ref{Fig:Global}B).
%
%Generating these new predictions consisted of four  steps: (1) downloading the ACS data and MOSAIKS feature matrix and merging these datasets on location, (2) splitting this merged dataset into training and holdout sets, (3) solving a ridge regression model on the training set using cross-validation to pick the model hyperparameter, and (4) evaluating this model on the holdout set. We note that variants of steps (1), (2), and (4) are essential to any SIML pipeline, and (3) is a built-in function to many programming platforms. 
%
Compared to a baseline of ``no ground survey,'' or a costly survey extension, these results suggest that \methodname\ predictions could provide useful information to a decision-maker for almost all tasks at low cost; noting that, in contrast, the ACS costs $>\$200$ million to deploy annually \cite{ACSBudget2021}. However, some variables (e.g. \textit{percent household income spent on rent}) may continue to be retrievable only via ground survey. 
%lowest $R^2=0.21$ for \textit{total housing units},
%highest $R^2=0.52$ for \textit{building age}),
%though there are some variables for which \methodname\ recovers essentially no signal (lowest $R^2=0.06$, \textit{percent household income spent on rent}). 

%\subsection*{Useful model properties}
\iftoggle{arxiv}{\subsection{Extensions}}{\subsection*{Extensions}}

\iftoggle{arxiv}{
\begin{figure}
\centering
\vspace{-1.5cm}
\includegraphics[width = \textwidth]{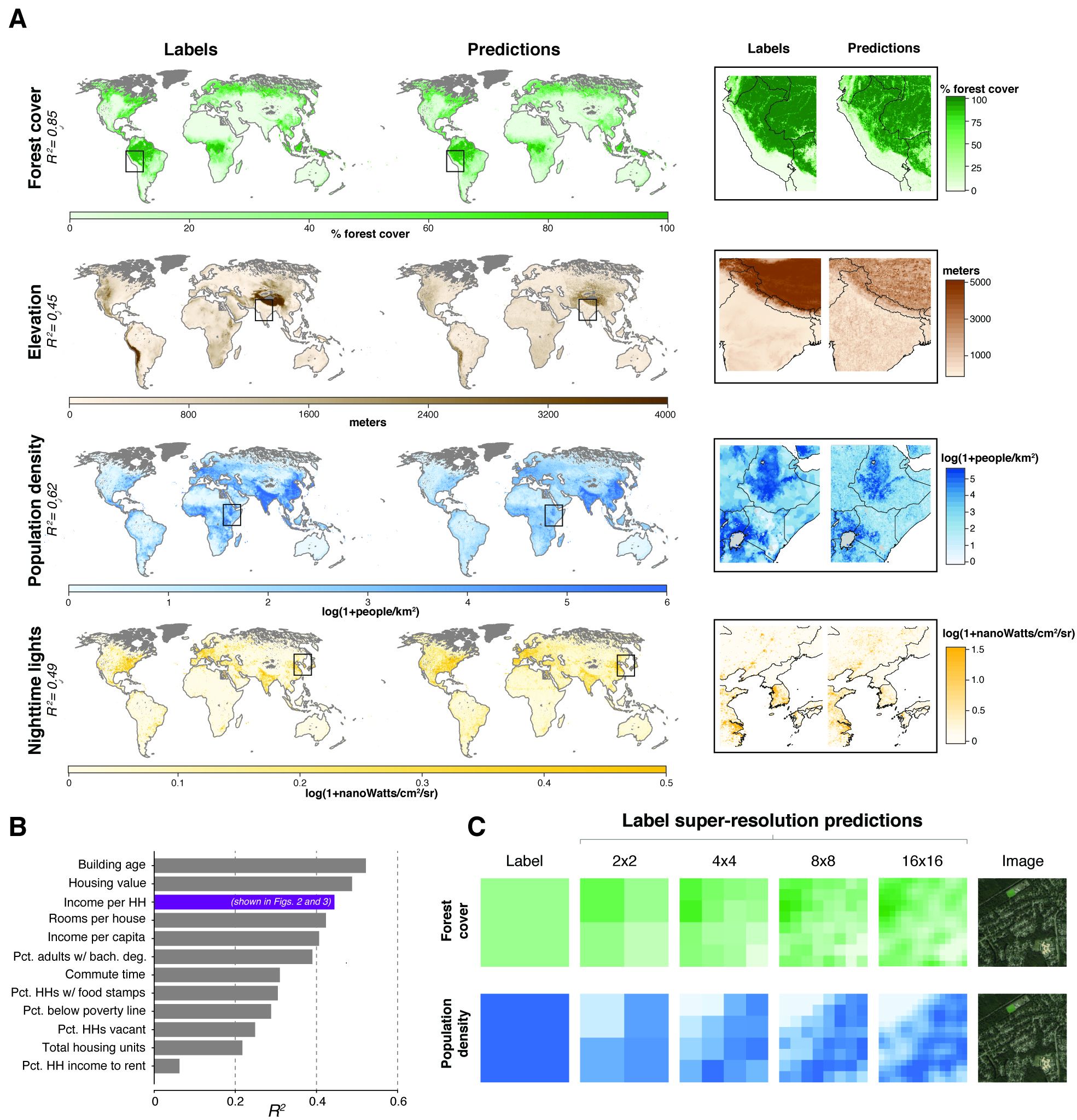}
\caption{\footnotesize \textbf{A single featurization of imagery predicts multiple variables at planet-scale, predicts results from a national survey, and achieves label super-resolution.} (A) Training data (left maps) and estimates using a single featurization of daytime imagery (right maps). Insets (far right) marked by black squares in global maps.  Training sample is a uniform random sampling of 1,000,000 land grid cells, 498,063 for which imagery were available and could be matched to task labels. Out-of-sample predictions are constructed using 5-fold cross-validation. For display purposes only, maps depict $\sim$50km $\times$ 50km average values (ground truth and predictions at $\sim$1km $\times$ 1km). (B) Test-set performance in the US shown for 12 variables from the 2015 American Community Survey (ACS) conducted by the US Census Bureau \cite{acs-income}. Income per household (in purple) is also shown in Figs. \ref{Fig:MapArray} and \ref{Fig:Waffle}, and was selected as an outcome for the analysis in these figures before this ACS experiment was run. (C) Both labels and features in \methodname\ are linear combinations of sub-image ground-level conditions, allowing optimized regression weights to be applied to imagery of any spatial extent (Supplementary Materials Section \ref{sec:superresolution}). MOSAIKS thus achieves label super-resolution by generating skillful estimates at spatial resolutions finer than the labels used for training. Shown are example label super-resolution estimates at $2\times 2$, $4\times 4$, $8\times 8$, and $16\times 16$ the original $1\times 1$ label resolution (See Fig. \ref{fig:superresolution_all10} for additional examples). Systematic evaluation of within-image $R^2$ across the entire sample is reported in Supplementary Materials Section~\ref{sec:superresolution} for the forest cover task.}
\label{Fig:Global}
\end{figure}
}{}

The design of \methodname\ naturally provides two additional useful properties.   

\iftoggle{arxiv}{\subsubsection{Incorporating multiple sensors}}{\paragraph{Incorporating multiple sensors}}
Available satellites exhibit a diversity of properties (e.g. wavelength, timing of sampling) that can be used to improve SIML predictions \cite{Tsagkatakis2019}. While most SIML approaches, including the above analysis, use a single sensor, the design of \methodname\ allows seamless integration of data from additional satellites because the regression step is linear in the features. To demonstrate this, we include nighttime lights as a second data source in the analysis of survey data from Rwanda, Haiti, and Nepal discussed above (Supplementary Materials \ref{sec:cnn_benchmark}). In all 36 tasks, predictions either improved or were unchanged when nighttime imagery was added to daytime imagery in the model (average $\Delta R^2= 0.03)$. This approach naturally optimizes how data from all sensors are used without requiring that users possess expertise on each technology.

\iftoggle{arxiv}{\subsubsection{Predicting at sub-image resolution}}{\paragraph{Predicting at sub-image resolution}}
Many use cases would benefit from SIML predictions at finer resolution than is available in training data \cite{Tsagkatakis2019,malkin2018label}. Here we show that \methodname\ can estimate the relative contribution of sub-regions within an image to overall image-level labels, even though only aggregated image-level labels are used in training (See Fig.~\ref{Fig:Global}C and Fig.~\ref{fig:superresolution_all10}). Such ``label super-resolution" prediction follows from the functional form of the featurization and linear regression steps in \methodname, allowing it to be analytically derived for labels that represent nearly linear combinations of ground-level conditions (Supplementary Materials Section~\ref{sec:superresolution} and Fig.~\ref{fig:superresolution_cartoon}). 
%Essentially, \methodname\ ``works'' because it estimates the relative contribution of sub-regions within an image to overall image-level labels, a property we exploit to achieve super-resolution. 
We numerically assess label super-resolution predictions of \methodname\ for the forest cover task, since raw label data are available at much finer resolution than our image labels. Provided only a single label per image, \methodname\ recovers substantial within-image signal when predicting forest cover in 4 to 1,024 sub-labels per label (within-image $R^2 = $ 0.54-0.32, see Fig.~\ref{Fig:superreslineplot} and Supplementary Materials Section~\ref{sec:superresolution}).

\iftoggle{arxiv}{\section{Discussion}}{\section*{Discussion}}

We develop a new approach to SIML that achieves practical generalization across tasks while exhibiting performance that is competitive with deep learning models optimized for a single task. Crucial to planet-scale analyses, \methodname\ requires orders of magnitude less computation time to solve a new task than CNN-based approaches and it allows 1km-by-1km image data to be compressed $\sim$6-500 times before storage/transmission (Supplementary Materials Section \ref{sec:main-methods}). Such compression is a deterministic operation that could theoretically be implemented in satellite hardware. We hope these computational gains, paired with the relative simplicity of using \methodname, will democratize access to global-scale SIML technology and accelerate its application to solving pressing global challenges. We hypothesize that there exist hundreds of variables observable from orbit whose application could improve human well-being if measurements were made accessible.

While we have shown that in many cases \methodname\ is a faster and simpler alternative to existing  deep learning methods, there remain contexts in which custom-designed SIML pipelines will continue to play a key role in research and decision-making. Existing ground-based surveys will also remain important. In both cases we expect \methodname\ can complement these systems, especially in resource constrained settings. For example, \methodname\ can provide fast assessments to guide slower SIML systems or extend the range and resolution of ground-based surveys.

As real-world policy actions increasingly depend on SIML predictions, it is crucial to understand the accuracy, precision and sensitivity of these measurements. The low cost and high speed of re-training \methodname\ enables unprecedented stress tests that can support robust SIML-based decision systems. 
Here, we tested the sensitivity of MOSAIKS to model parameters, number of training points, and degree of spatial extrapolation, and expect that many more tests can be developed and implemented to analyze model performance and prediction accuracies in context.
To aid systematic bench-marking and comparison of SIML architectures, the labels and features used in this study are made publicly available; to our knowledge this represents the largest multi-label benchmark dataset for SIML regression tasks. The high performance of RCF, a relatively simple featurization, suggests that developing and benchmarking other unsupervised SIML methods across tasks at scale may be a rich area for future research.

%%%--- We can make MOSAIKS better ---%%%

By distilling SIML to a pipeline with simple and mathematically interpretable components, \methodname\ facilitates development of methodologies for additional SIML use cases and enhanced performance.
%The high accuracy of a simple featurization (RCF) suggests there is room to improve unsupervised featurization for SIML, without sacrificing complexity or training time.
For example, the ability of \methodname\ to achieve label super-resolution is easily derived analytically (Supplementary Materials Section \ref{sec:superresolution}). Furthermore, while we have focused here on tri-band daytime imagery, we showed that \methodname\ can seamlessly integrate data from multiple sensors through simple concatenation, extracting useful information from each source to maximize performance. We conjecture that integrating new diverse data, from both satellite and non-satellite sources, may substantially increase  predictive accuracy for tasks not entirely resolved by daytime imagery alone. 

We hope that \methodname\ lays the foundation for 
an accessible and democratized system of global information sharing, where imagery from all available global sensors is continuously encoded as features and stored in a single table of data, which is then distributed and used planet-wide. As a step in this direction, we make a global cross-section of features publicly available using 2019 imagery from Planet Labs, Inc.  Such a unified global system may enhance our collective ability to observe and understand the world, a necessary condition for tackling pressing global challenges. 

%\newpage

\iftoggle{arxiv}{}{\singlespacing
\section*{Code and data availability} 
Code, data, a configured computing environment, and free cloud computing for this analysis is provided via Code Ocean. The editor will be provided with a link to the Code Ocean ``capsule'', which will be distributed to reviewers.

All data used in this analysis is from free, publicly available sources and is available for download other than the house price data. House price data is provided by Zillow through the Zlllow Transaction and Assessment Dataset (ZTRAX). More information on accessing the data can be found at \url{http://www.zillow.com/ztrax}. The results and opinions are those of the author(s) and do not reflect the position of Zillow Group. The house price dataset we release publicly is a subset of that which used in the analysis, where grid cells containing $<$30 observations of recent property sales are removed to preserve privacy. Instructions for downloading the replication data are included in the Readme file within the project's Code Ocean capsule.

At the time of submission, interested users can obtain random convolutional features in areas of interest by emailing the authors a csv file with locations (latitude, longitude) along with a short description of the intended use. Concurrent with review, we are automating this process to further simplify the user experience.
}

\section*{Acknowledgements}
We thank 
Patrick Baylis,
Joshua Blumenstock,
Jennifer Burney,
Hannah Druckenmiller,
Jonathan Kadish,
Alyssa Morrow,
James Rising,
Geoffrey Schiebinger,
and participants in seminars at 
UC Berkeley,
University of Chicago,
Harvard,
American Geophysical Union,
The World Bank,
The United Nations Development Program \& Environment Program,
Planet Inc.,
The Potsdam Institute for Climate Impact Research, and
The Workshop in Environmental Economics and Data Science
for helpful comments and suggestions.
We acknowledge funding from 
the NSF Graduate Research Fellowship Program (Grant DGE 1752814),
the US Environmental Protection Agency Science To Achieve Results Fellowship Program (Grant FP91780401),
the NSF Research Traineeship Program Data Science for the 21st Century, the Harvard Center for the Environment, the Harvard Data Science Initiative,
the Sloan Foundation,
and a gift from the Rhodium Group. 
The authors declare no conflicts of interest.

% Authors must make available upon request, to editors and reviewers, any previously unreported custom computer code or algorithm used to generate results that are reported in the paper and central to its main claims. Any reason that would preclude the need for code or algorithm sharing will be evaluated by the editors who reserve the right to decline the paper if important code is unavailable.
% this section needs to house a link to the github repo.

%\printbibliography[resetnumbers=true]
\iftoggle{arxiv}{\bibliographystyle{unsrt}}{\bibliographystyle{Science_w_title}}
\bibliography{references_mendeley_ER2} % appendix_bib.bib

\begin{thebibliography}{10}

\bibitem{scientists_2019}
{Union of Concerned Scientists}.
\newblock {UCS Satellite Database}, 1 2019.

\bibitem{hansen2013}
M~C Hansen, P~V Potapov, R~Moore, M~Hancher, S~A Turubanova, A~Tyukavina,
  D~Thau, S~V Stehman, S~J Goetz, T~R Loveland, A~Kommareddy, A~Egorov,
  L~Chini, C~O Justice, and J~R~G Townshend.
\newblock {High-resolution global maps of 21st-century forest cover change.}
\newblock {\em Science (New York, N.Y.)}, 342(6160):850--3, 11 2013.

\bibitem{Inglada2017}
Jordi Inglada, Arthur Vincent, Marcela Arias, Benjamin Tardy, David Morin, and
  Isabel Rodes.
\newblock {Operational high resolution land cover map production at the country
  scale using satellite image time series}.
\newblock {\em Remote Sensing}, 9(1):95, 2017.

\bibitem{Jean2016}
Neal Jean, Marshall Burke, Michael Xie, W.~Matthew Davis, David~B. Lobell, and
  Stefano Ermon.
\newblock {Combining satellite imagery and machine learning to predict
  poverty}.
\newblock {\em Science}, 353(6301):790--794, 2016.

\bibitem{Robinson2017}
Caleb Robinson, Fred Hohman, and Bistra Dilkina.
\newblock {A Deep Learning Approach for Population Estimation from Satellite
  Imagery}.
\newblock In {\em Proceedings of the 1st ACM SIGSPATIAL Workshop on Geospatial
  Humanities - GeoHumanities 2017}, pages 47--54, New York, New York, USA,
  2017. ACM Press.

\bibitem{Yu2014}
Le~Yu, Lu~Liang, Jie Wang, Yuanyuan Zhao, Qu~Cheng, Luanyun Hu, Shuang Liu,
  Liang Yu, Xiaoyi Wang, Peng Zhu, Xueyan Li, Yue Xu, Congcong Li, Wei Fu,
  Xuecao Li, Wenyu Li, Caixia Liu, Na~Cong, Han Zhang, Fangdi Sun, Xinfang Bi,
  Qinchuan Xin, Dandan Li, Donghui Yan, Zhiliang Zhu, Michael~F. Goodchild, and
  Peng Gong.
\newblock {Meta-discoveries from a synthesis of satellite-based land-cover
  mapping research}.
\newblock {\em International Journal of Remote Sensing}, 35(13):4573--4588, 7
  2014.

\bibitem{Haack2016}
Barry Haack and Robert Ryerson.
\newblock {Improving remote sensing research and education in developing
  countries: Approaches and recommendations}.
\newblock {\em International Journal of Applied Earth Observation and
  Geoinformation}, 45:77--83, 3 2016.

\bibitem{Romero2016}
Adriana Romero, Carlo Gatta, and Gustau Camps-Valls.
\newblock {Unsupervised deep feature extraction for remote sensing image
  classification}.
\newblock {\em IEEE Transactions on Geoscience and Remote Sensing},
  54(3):1349--1362, 2016.

\bibitem{Cheriyadat2014}
Anil~M. Cheriyadat.
\newblock {Unsupervised feature learning for aerial scene classification}.
\newblock {\em IEEE Transactions on Geoscience and Remote Sensing},
  52(1):439--451, 2014.

\bibitem{penatti2015deep}
Otávio A~B Penatti, Keiller Nogueira, Jefersson~A Dos~Santos, and Jefersson
  A~Dos Santos.
\newblock {Do Deep Features Generalize from Everyday Objects to Remote Sensing
  and Aerial Scenes Domains?}
\newblock {\em Proceedings of the IEEE conference on computer vision and
  pattern recognition workshops}, pages 44--51, 2015.

\bibitem{Jean2019}
Neal Jean, Sherrie Wang, Anshul Samar, George Azzari, David Lobell, and Stefano
  Ermon.
\newblock {Tile2Vec: Unsupervised Representation Learning for Spatially
  Distributed Data}.
\newblock In {\em Proceedings of the AAAI Conference on Artificial
  Intelligence}, volume~33, pages 3967--3974, 2019.

\bibitem{littlepage}
Jay Littlepage.
\newblock {DigitalGlobe moves to the cloud with AWS Snowmobile}.

\bibitem{Rahimi2008}
Ali Rahimi and Benjamin Recht.
\newblock {Weighted sums of random kitchen sinks: Replacing minimization with
  randomization in learning}.
\newblock {\em Advances in neural information processing {\ldots}},
  1(1):1313--1320, 2008.

\bibitem{Morrow2017}
Alyssa Morrow, Vaishaal Shankar, Devin Petersohn, Anthony Joseph, Benjamin
  Recht, and Nir Yosef.
\newblock {Convolutional Kitchen Sinks for Transcription Factor Binding Site
  Prediction}.
\newblock {\em arXiv preprint}, 2017.

\bibitem{Coates2012}
Adam Coates and Andrew~Y Ng.
\newblock {Learning feature representations with K-means}.
\newblock In {\em Neural networks: Tricks of the trade}. Springer, Berlin,
  Heidelberg, 2012.

\bibitem{Jonas2018}
Eric Jonas, Monica Bobra, Vaishaal Shankar, J.~Todd~Hoeksema, and Benjamin
  Recht.
\newblock {Flare Prediction Using Photospheric and Coronal Image Data}.
\newblock {\em Solar Physics}, 293(3):1--22, 2018.

\bibitem{Blumenstock2018}
Joshua Blumenstock.
\newblock {Don’t forget people in the use of big data for development}, 2018.

\bibitem{He2016}
Kaiming He, Xiangyu Zhang, Shaoqing Ren, and Jian Sun.
\newblock {Deep residual learning for image recognition}.
\newblock In {\em Proceedings of the IEEE conference on computer vision and
  pattern recognition}, pages 770--778, 2016.

\bibitem{Pan2010}
Sinno~Jialin Pan and Qiang Yang.
\newblock {A survey on transfer learning}.
\newblock {\em IEEE Transactions on Knowledge and Data Engineering},
  22(10):1345--1359, 2010.

\bibitem{xie2016transfer}
Michael Xie, Neal Jean, Marshall Burke, David Lobell, and Stefano Ermon.
\newblock {Transfer learning from deep features for remote sensing and poverty
  mapping}.
\newblock In {\em Thirtieth AAAI Conference on Artificial Intelligence}, 2016.

\bibitem{Head2017}
Andrew Head, Mélanie Manguin, Nhat Tran, and Joshua~E Blumenstock.
\newblock {Can human development be measured with satellite imagery?}
\newblock In {\em ICTD}, pages 1--8, 2017.

\bibitem{Simonyan2015}
Karen Simonyan and Andrew Zisserman.
\newblock {Very deep convolutional networks for large-scale image recognition}.
\newblock In {\em International Conference on Learning Representations}, 2015.

\bibitem{Athey2017b}
Susan Athey.
\newblock {Beyond prediction: Using big data for policy problems}.
\newblock {\em Science}, 355(6324):483--485, 2 2017.

\bibitem{Reed2018}
Fennis Reed, Andrea Gaughan, Forrest Stevens, Greg Yetman, Alessandro
  Sorichetta, and Andrew Tatem.
\newblock {Gridded Population Maps Informed by Different Built Settlement
  Products}.
\newblock {\em Data}, 3(3):33, 9 2018.

\bibitem{Bedi2007}
Tara Bedi, Aline Coudouel, and Kenneth Simler, editors.
\newblock {\em {More than a pretty picture: Using poverty maps to design better
  policies and interventions}}.
\newblock The World Bank, Washington, DC, 2007.

\bibitem{DeSherbinin2017}
A.~M. De~Sherbinin, G.~Yetman, K.~MacManus, and S.~Vinay.
\newblock {Improved Mapping of Human Population and Settlements through
  Integration of Remote Sensing and Socioeconomic Data}.
\newblock {\em AGUFM}, 2017:IN51H--06, 2017.

\bibitem{acs-income}
{U.S. Census Bureau}.
\newblock {2015 American Community Survey 5-Year Estimates, Table B19013}.

\bibitem{ACSBudget2021}
{U.S. Census Bureau}.
\newblock {Budget Estimates, Fiscal Year 2021}, 2021.

\bibitem{Tsagkatakis2019}
Grigorios Tsagkatakis, Anastasia Aidini, Konstantina Fotiadou, Michalis
  Giannopoulos, Anastasia Pentari, and Panagiotis Tsakalides.
\newblock {Survey of deep-learning approaches for remote sensing observation
  enhancement}.
\newblock {\em Sensors (Switzerland)}, 19(18):1--39, 2019.

\bibitem{malkin2018label}
Kolya Malkin, Caleb Robinson, Le~Hou, Rachel Soobitsky, Jacob Czawlytko,
  Dimitris Samaras, Joel Saltz, Lucas Joppa, and Nebojsa Jojic.
\newblock {Label super-resolution networks}.
\newblock In {\em International Conference on Learning Representations}, 2019.

\bibitem{awstiles}
{Amazon Web Services}.
\newblock {Terrain Tiles}, 2018.

\bibitem{gpw}
{Center for International Earth Science Information Network (CIESIN)}.
\newblock {Gridded Population of the World, Version 4}, 2016.

\bibitem{viirs-nl}
{NOAA National Centers for Environmental Information}.
\newblock {Version 1 VIIRS Day/Night Band Nighttime Lights}, 2019.

\bibitem{tiger-roads}
{U.S. Census Bureau}.
\newblock {TIGER/Line Geodatabases}, 2016.

\bibitem{ztrax}
{Zillow}.
\newblock {ZTRAX: Zillow Transaction and Assessor Dataset}, 2017.

\bibitem{Perez2017}
Anthony Perez, Christopher Yeh, George Azzari, Marshall Burke, David Lobell,
  and Stefano Ermon.
\newblock {Poverty prediction with public Landsat 7 satellite imagery and
  machine learning}.
\newblock In {\em NIPS 2017 Workshop on Machine Learning for the Developing
  World}, 2017.

\bibitem{RahimiRecht2008}
Ali Rahimi and Benjamin Recht.
\newblock {Uniform approximation of functions with random bases}.
\newblock In {\em 46th Annual Allerton Conference on Communication, Control,
  and Computing}, pages 555--561. IEEE, 9 2008.

\bibitem{PerezSuay2017}
Adrián P{\'{e}}rez-Suay, Julia Amor{\'{o}}s-L{\'{o}}pez, Luis
  G{\'{o}}mez-Chova, Valero Laparra, Jordi Mu{\~{n}}oz-Mar{\'{i}}, and Gustau
  Camps-Valls.
\newblock {Randomized kernels for large scale Earth observation applications}.
\newblock {\em Remote Sensing of Environment}, 202:54--63, 12 2017.

\bibitem{Alkama2016}
Ramdane Alkama and Alessandro Cescatti.
\newblock {Biophysical climate impacts of recent changes in global forest
  cover}.
\newblock {\em Science}, 351(6273):600--604, 2 2016.

\bibitem{Carlson2018}
Kimberly~M Carlson, Robert Heilmayr, Holly~K Gibbs, Praveen Noojipady, David~N
  Burns, Douglas~C Morton, Nathalie~F Walker, Gary~D Paoli, and Claire Kremen.
\newblock {Effect of oil palm sustainability certification on deforestation and
  fire in Indonesia.}
\newblock {\em Proceedings of the National Academy of Sciences of the United
  States of America}, 115(1):121--126, 1 2018.

\bibitem{acsPackage}
Ezra~Haber Glenn.
\newblock {acs: Download, Manipulate, and Present American Community Survey and
  Decennial Data from the US Census}, 2019.

\bibitem{Moulton2018}
Jeremy Moulton and Scott Wentland.
\newblock {Monetary Policy and the Housing Market}.
\newblock In {\em Annual Meeting of the American Economic Association},
  Philadelphia, PA, 2018.

\bibitem{Gindelsky2019}
Marina Gindelsky, Jeremy Moulton, and Scott Wentland.
\newblock {Valuing Housing Services in the Era of Big Data: A User Cost
  Approach Leveraging Zillow Microdata}.
\newblock In {NBER}, editor, {\em Big Data for 21st Century Economic
  Statistics}. University of Chicago Press, 2019.

\bibitem{UnionofConcernedScientists2018}
{Union of Concerned Scientists}.
\newblock {Underwater: Rising Seas, Chronic Floods, and the Implications for US
  Coastal Real Estate}.
\newblock Technical report, Union of Concerned Scientists, 2018.

\bibitem{ZillowResearch}
{Zillow Research}.
\newblock {zillow-research/ztrax}.

\bibitem{krizhevsky2009learning}
Alex Krizhevsky and Geoffrey Hinton.
\newblock {Learning multiple layers of features from tiny images}.
\newblock Technical report, Citeseer, 2009.

\bibitem{Coates2011}
Adam Coates, Ann Arbor, and Andrew~Y Ng.
\newblock {An Analysis of Single-Layer Networks in Unsupervised Feature
  Learning}.
\newblock {\em International Conference on Articial Intelligence and
  Statistics}, pages 215--223, 2011.

\bibitem{recht2019imagenet}
Benjamin Recht, Rebecca Roelofs, Ludwig Schmidt, and Vaishaal Shankar.
\newblock {Do ImageNet Classifiers Generalize to ImageNet?}
\newblock In {\em International Conference on Machine Learning}, pages
  5389--5400, 2019.

\bibitem{agarwal2013least}
Alekh Agarwal, Sham~M Kakade, Nikos Karampatziakis, Le~Song, and Gregory
  Valiant.
\newblock {Least squares revisited: Scalable approaches for multi-class
  prediction}.
\newblock In {\em International Conference on Machine Learning}, pages
  541--549, 2014.

\bibitem{Rahimi}
Ali Rahimi and Ben Recht.
\newblock {Random Features for Large-Scale Kernel Machines}.
\newblock In {\em Advances in Neural Information Processing Systems}, 2007.

\bibitem{Daniely2016}
Amit Daniely, Roy Frostig, and Yoram Singer.
\newblock {Toward deeper understanding of neural networks: The power of
  initialization and a dual view on expressivity}.
\newblock {\em Advances in Neural Information Processing Systems}, pages
  2261--2269, 2016.

\bibitem{alber2017}
Maximilian Alber, Pieter-Jan Kindermans, Kristof~T. Sch{\"{u}}tt, Klaus-Robert
  M{\"{u}}ller, and Fei Sha.
\newblock {An empirical study on the properties of random bases for kernel
  methods}.
\newblock In {\em Neural Information Processing Systems}, 2017.

\bibitem{zhou2016learning}
Bolei Zhou, Aditya Khosla, Agata Lapedriza, Aude Oliva, and Antonio Torralba.
\newblock {Learning deep features for discriminative localization}.
\newblock In {\em Proceedings of the IEEE conference on computer vision and
  pattern recognition}, pages 2921--2929, 2016.

\bibitem{Firat2014}
Orhan Firat, Gulcan Can, and Fatos T.~Yarman Vural.
\newblock {Representation learning for contextual object and region detection
  in remote sensing}.
\newblock {\em Proceedings - International Conference on Pattern Recognition},
  pages 3708--3713, 2014.

\bibitem{Volpi2017}
Michele Volpi and Devis Tuia.
\newblock {Dense semantic labeling of subdecimeter resolution images with
  convolutional neural networks}.
\newblock {\em IEEE Transactions on Geoscience and Remote Sensing},
  55(2):881--893, 2017.

\bibitem{vincent2010stacked}
Pascal Vincent, Hugo Larochelle, Isabelle Lajoie, Yoshua Bengio, and
  Pierre-Antoine Manzagol.
\newblock {Stacked denoising autoencoders: Learning useful representations in a
  deep network with a local denoising criterion}.
\newblock {\em Journal of machine learning research}, 11(Dec):3371--3408, 2010.

\bibitem{Bolliger2017}
Ian Bolliger, Tamma Carleton, Solomon Hsiang, Jonathan Kadish, Jonathan
  Proctor, Benjamin Recht, Esther Rolf, and Vaishaal Shankar.
\newblock {Ground Control to Major Tom: the importance of field surveys in
  remotely sensed data analysis}.
\newblock {\em arXiv}, 2017.

\bibitem{krizhevsky2012imagenet}
Alex Krizhevsky, Ilya Sutskever, and Geoffrey~E Hinton.
\newblock {Imagenet classification with deep convolutional neural networks}.
\newblock In {\em Advances in neural information processing systems}, pages
  1097--1105, 2012.

\bibitem{Gechter2018a}
Michael Gechter, Nick Tsivanidis, Russell Cooper, Jonathan Eaton, Pablo
  Fajgelbaum, Jon Hersh, Chang-Tai Hsieh, Dan Keniston, Kala Krishna, Abhay
  Pethe, Vidhadyar Phatak, Debraj Ray, Steve Redding, Han Chen, Inhyuk Choi,
  JD~Costantini, Dongyang He, Suraj Jacob, Peiquan Lin, Alejandra Lopez, Rucha
  Pandit, Pankti Sanghvi, Christine Suhr, and Sneha Thube.
\newblock {The Welfare Consequences of Formalizing Developing Country Cities:
  Evidence from the Mumbai Mills Redevelopment}.
\newblock {\em Working paper}, 2018.

\bibitem{Zhong2017}
Yanfei Zhong, Feng Fei, Yanfei Liu, Bei Zhao, Hongzan Jiao, and Liangpei Zhang.
\newblock {SatCNN: satellite image dataset classification using agile
  convolutional neural networks}.
\newblock {\em Remote Sensing Letters}, 8(2):136--145, 2 2017.

\bibitem{Hu2019}
Wenjie Hu, Jay~Harshadbhai Patel, Zoe-Alanah Robert, Paul Novosad, Samuel
  Asher, Zhongyi Tang, Marshall Burke, David Lobell, and Stefano Ermon.
\newblock {Mapping Missing Population in Rural India: A Deep Learning Approach
  with Satellite Imagery}.
\newblock In {\em Conference on Artificial Intelligence, Ethics, and Society},
  2019.

\bibitem{Maggiori2017}
Emmanuel Maggiori, Yuliya Tarabalka, Guillaume Charpiat, and Pierre Alliez.
\newblock {Convolutional Neural Networks for Large-Scale Remote-Sensing Image
  Classification}.
\newblock {\em IEEE Transactions on Geoscience and Remote Sensing},
  55(2):645--657, 2 2017.

\bibitem{Cheng2017}
Gong Cheng, Junwei Han, and Xiaoqiang Lu.
\newblock {Remote sensing image scene classification: Benchmark and state of
  the art}.
\newblock {\em Proceedings of the IEEE}, 105(10):1865--1883, 10 2017.

\bibitem{Zoph2016NeuralLearning}
Barret Zoph and Quoc~V Le.
\newblock {Neural architecture search with reinforcement learning}.
\newblock In {\em International Conference on Learning Representations}, 2017.

\bibitem{strubell2019energy}
Emma Strubell, Ananya Ganesh, and Andrew McCallum.
\newblock {Energy and Policy Considerations for Deep Learning in NLP}.
\newblock In {\em Proceedings of the 57th Annual Meeting of the Association for
  Computational Linguistics}, pages 3654--3650, 2019.

\bibitem{Ruder2017}
Sebastian Ruder.
\newblock {An Overview of Multi-Task Learning in Deep Neural Networks}.
\newblock {\em arXiv preprint}, 6 2017.

\end{thebibliography}

\iftoggle{arxiv}{}{
\newpage
%\todomisc{Make sure all figures are updated to latest version.}
\begin{sidewaysfigure}
\centering
\includegraphics[width=\textwidth]{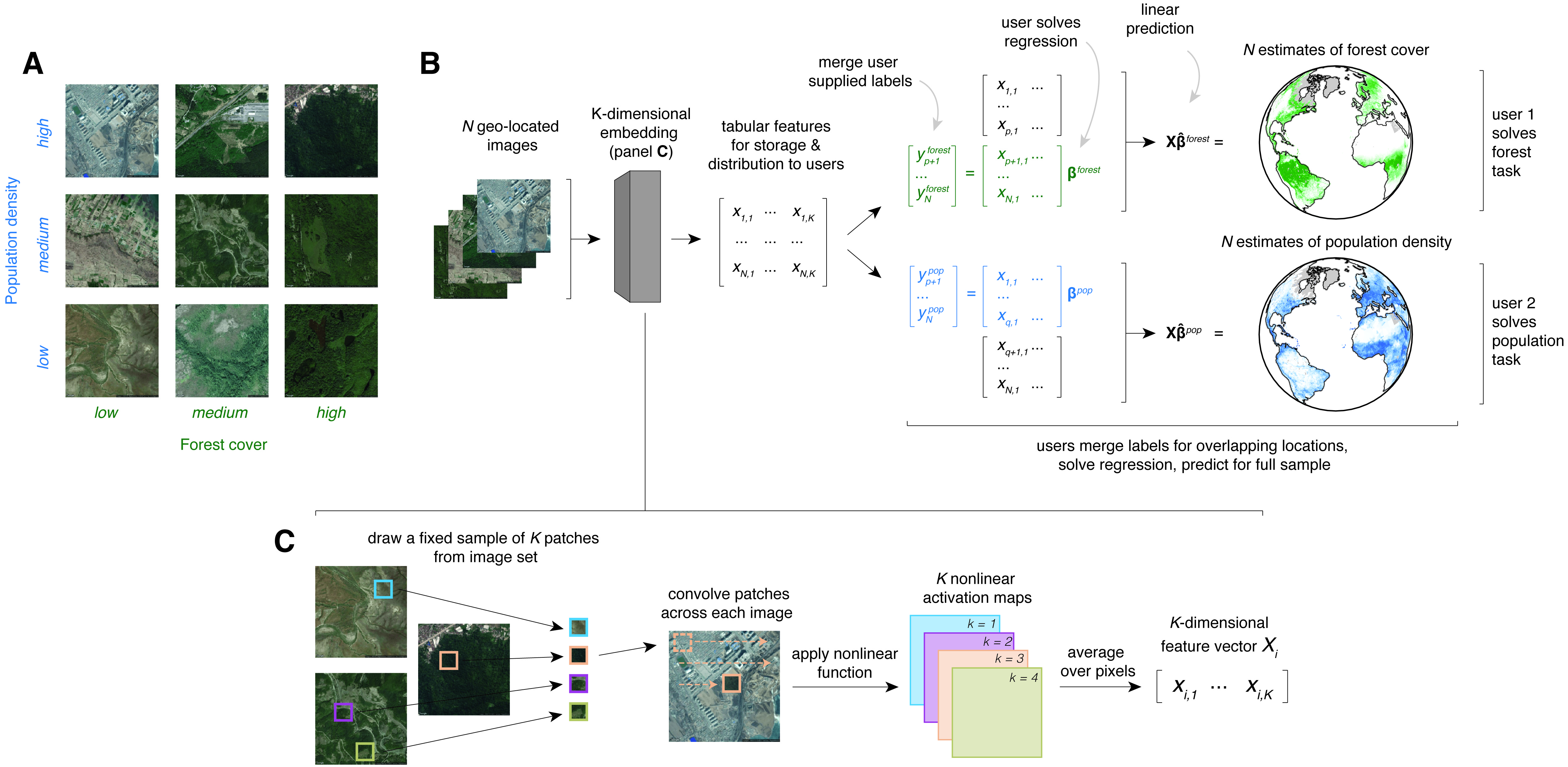}
\caption{\footnotesize \textbf{A generalizable approach to combining satellite imagery with machine learning (SIML) without users handling images.}
\methodname\ is designed to solve an unlimited number of tasks at planet-scale quickly. After a one-time unsupervised image featurization using random convolutional features, \methodname\ centrally stores and distributes task-agnostic features to users, each of whom generates predictions in a new context.
(A) Satellite imagery is shared across multiple potential tasks. For example, nine images from the US sample are ordered based on population density and forest cover, both of which have distinct identifying features that are observable in each image.  (B) Schematic of the \methodname\ process. $N$ images are transformed using random convolutional features into a compressed and highly descriptive $K$-dimensional feature vector before labels are known.  Once features are computed, they can be stored in tabular form (matrix $\textbf{X}$) and used for unlimited tasks without recomputation. Users interested in a new task ($s$) merge their own labels ($y^s$) to features for training. Here, \textit{User 1} has forest cover labels for locations $p+1$ to $N$ and \textit{User 2} has population density labels for locations $1$ to $q$. Each user then solves a single linear regression for $\bm\beta^s$. Linear prediction using $\bm\beta^s$ and the full sample of \methodname\ features $\textbf{X}$ then generates SIML estimates for label values at all locations. Generalizability allows different users to solve different tasks using an identical procedure and the same table of features---differing only in the user-supplied label data for training. Each task can be solved by a user on a desktop computer in minutes without users ever manipulating the imagery. (C) Illustration of the one-time unsupervised computation of random convolutional features (Supplementary Materials Sections \ref{sec:main-methods} and \ref{sec:featurization}). $K$ patches are randomly sampled from across the $N$ images. Each patch is convolved over each image, generating a nonlinear activation map for each patch. Activation maps are averaged over pixels to generate a single $K$-dimensional feature vector for each image.}
\label{Fig:Cartoon}
\end{sidewaysfigure}

\begin{figure}
\vspace{-2.5cm}
\centering
\includegraphics[height=.9\textheight]{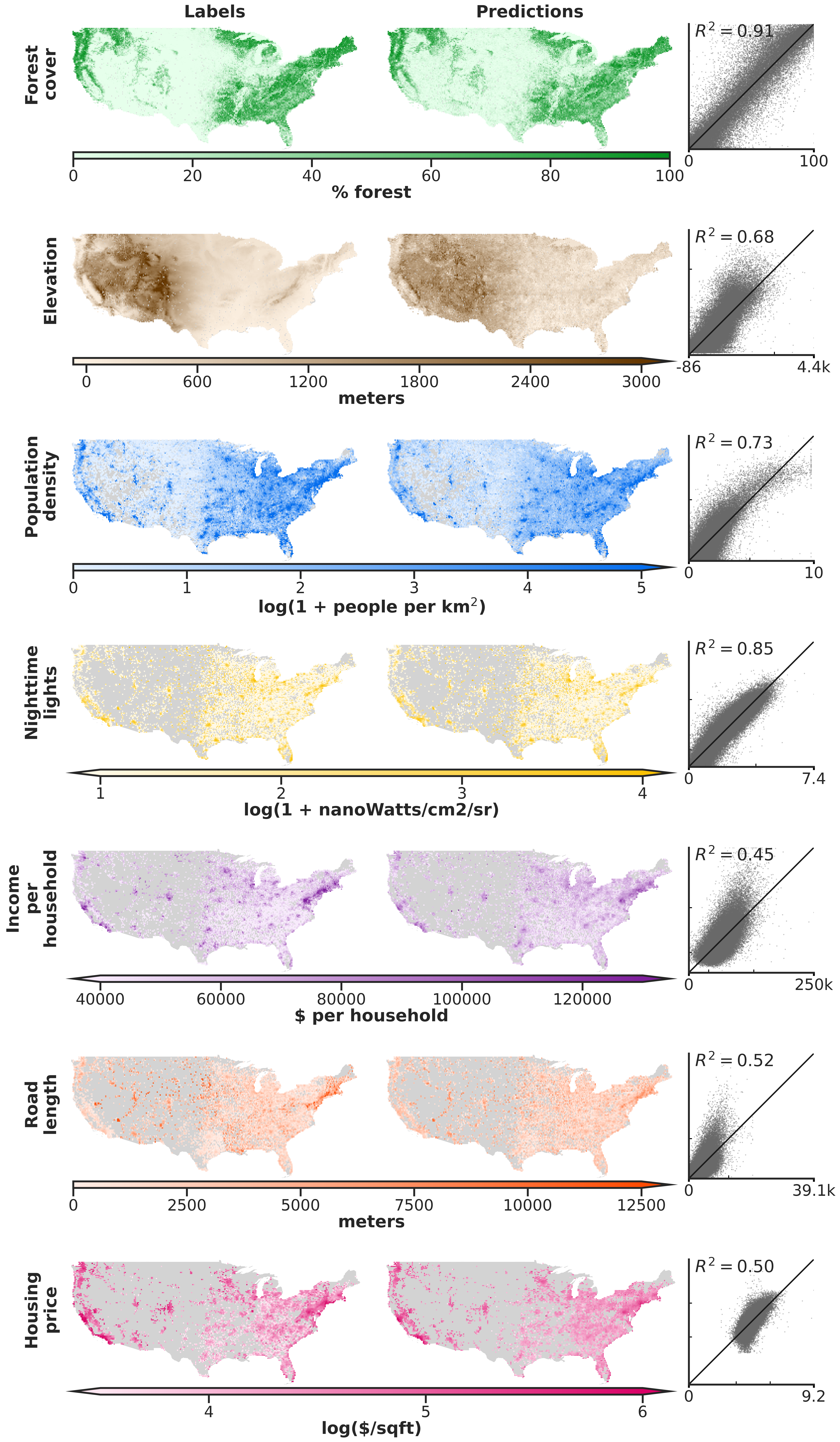}
\caption{\footnotesize \textbf{1km $\times$ 1km resolution prediction of many tasks across the continental US using daytime images processed once, before tasks were chosen.}  100,000 daytime images were each converted to 8,192 features and stored. Seven tasks were then selected based on coverage and diversity and predictions were generated for each task using the same procedure. Left maps: 80,000 observations used for training and validation, aggregated up to 20km$\times$20km cells for display. Right maps: 
concatenated validation set estimates from 5-fold cross-validation for the same 80,000 grid cells (observations are never used to generate their own prediction), identically aggregated for display. Scatters: Validation set estimates (vertical axis) vs. ground-truth (horizontal axis); each point is a $\sim$1km$\times$1km grid cell. Black line is at $45^\circ$. Test set and validation set performance are essentially identical (Table \ref{SI:testSetTable}); validation set values are shown for display purposes only, as there are more observations. The tasks in the top three rows are uniformly sampled across space, the tasks in the bottom four rows are sampled using population weights (Supplementary Materials Section \ref{sec:grid_and_sampling}); grey areas were not sampled in the experiment.}
\label{Fig:MapArray}
\end{figure}

\begin{figure}
\centering
\vspace{-1.5cm}
\includegraphics[width = .9\textwidth]{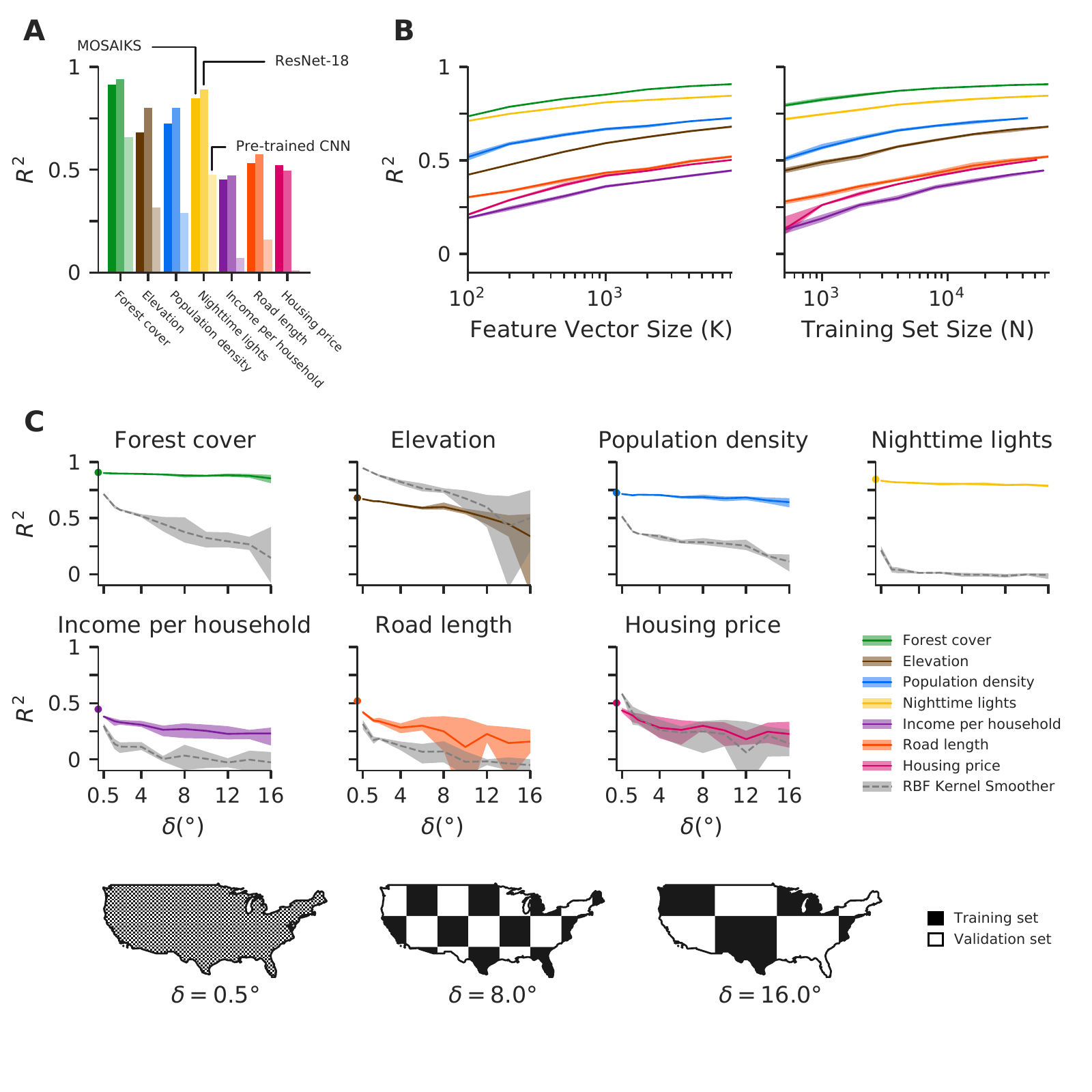}
\caption{\footnotesize \textbf{Prediction accuracy relative to a convolutional neural network and transfer learning, using smaller $K$ and $N$, and over large contiguous regions with no ground truth data.} (A) Task-specific \methodname\ test-set performance (dark bars) in contrast to: an 18-layer variant of the ResNet Architecture (ResNet-18) trained end-to-end for each task (middle bars); and an unsupervised featurization using the last hidden layer of a 152-layer ResNet variant pre-trained on natural imagery and applied using ridge regression (lightest bars). See Supplementary Materials Section \ref{sec:cnn_benchmark} for details. (B) Validation set $R^2$ performance for all seven tasks while varying the number of random convolutional features $K$ and holding $N=64,000$ (left) and while varying $N$ and holding $K=8,192$ (right). Shaded bands indicate the range of predictive skill across 5 folds. Lines indicate average accuracy across folds. (C) Evaluation of performance over regions of increasing size that that are excluded from training sample. Data are split using a ``checkerboard'' partition, where the width and height of each square is $\delta$ (measured in degrees). Example partitions with $\delta=0.5^\circ$, $8^\circ$, 16$^\circ$ are shown in maps. For a given $\delta$, training occurs using data sampled from ``black squares'' and performance is evaluated in ``white squares.'' Plots show colored lines representing average performance of \methodname\ in the US across $\delta$ values for each task. Benchmark performance from Fig.~\ref{Fig:MapArray} are indicated as circles at $\delta=0$. Grey dashed lines indicate corresponding performance using only spatial interpolation with an optimized radial basis function kernel instead of \methodname\ (Supplementary Materials Section \ref{sec:secondary_analysis_spatial}). To moderate the influence of the exact placement of square edges, training and test sets are resampled four times for each $\delta$ with the checkerboard position re-initialized using offset vertices (see Supplementary Materials Section \ref{sec:secondary_analysis_spatial} and Fig.~\ref{fig:jitter_explanation}). The ranges of performance are plotted as colored or grey bands.}
\label{Fig:Waffle}
\end{figure}

\begin{figure}
\centering
\vspace{-1.5cm}
\includegraphics[width = \textwidth]{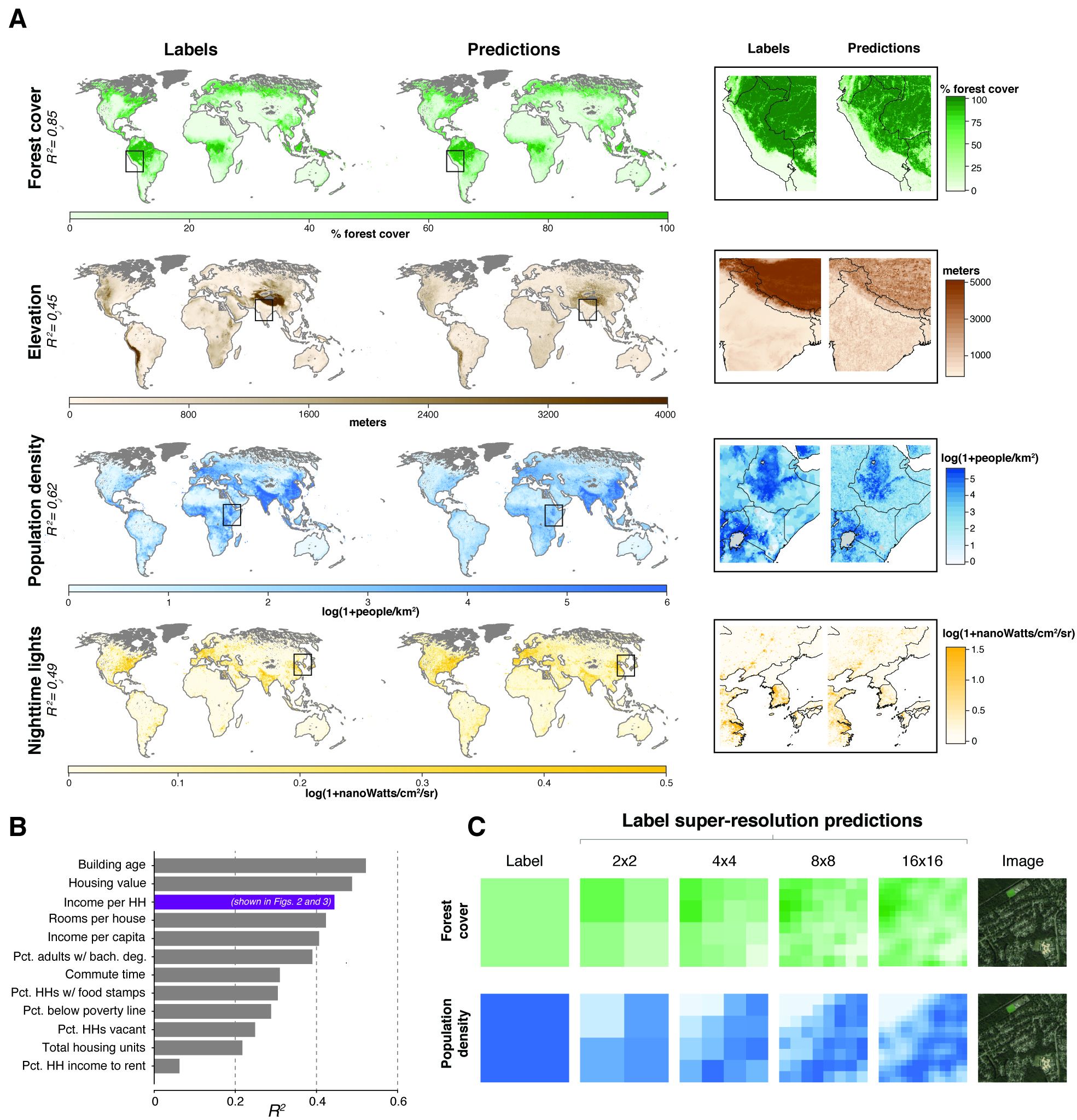}
\caption{\footnotesize \textbf{A single featurization of imagery predicts multiple variables at planet-scale, predicts results from a national survey, and achieves label super-resolution.} (A) Training data (left maps) and estimates using a single featurization of daytime imagery (right maps). Insets (far right) marked by black squares in global maps.  Training sample is a uniform random sampling of 1,000,000 land grid cells, 498,063 for which imagery were available and could be matched to task labels. Out-of-sample predictions are constructed using 5-fold cross-validation. For display purposes only, maps depict $\sim$50km $\times$ 50km average values (ground truth and predictions at $\sim$1km $\times$ 1km). (B) Test-set performance in the US shown for 12 variables from the 2015 American Community Survey (ACS) conducted by the US Census Bureau \cite{acs-income}. Income per household (in purple) is also shown in Figs. \ref{Fig:MapArray} and \ref{Fig:Waffle}, and was selected as an outcome for the analysis in these figures before this ACS experiment was run. (C) Both labels and features in \methodname\ are linear combinations of sub-image ground-level conditions, allowing optimized regression weights to be applied to imagery of any spatial extent (Supplementary Materials Section \ref{sec:superresolution}). MOSAIKS thus achieves label super-resolution by generating skillful estimates at spatial resolutions finer than the labels used for training. Shown are example label super-resolution estimates at $2\times 2$, $4\times 4$, $8\times 8$, and $16\times 16$ the original $1\times 1$ label resolution (See Fig. \ref{fig:superresolution_all10} for additional examples). Systematic evaluation of within-image $R^2$ across the entire sample is reported in Supplementary Materials Section~\ref{sec:superresolution} for the forest cover task.}
\label{Fig:Global}
\end{figure}

}

%\end{refsection}

\clearpage
\newpage

%%%%%%%%%%%% SUPPLEMENT %%%%%%%%%%%%%

% \renewcommand \thepart{}
% \renewcommand \partname{}

\appendix
%\addcontentsline{toc}{part}{}
\addcontentsline{toc}{section}{Appendix} % Add the appendix text to the document TOC
\part{Supplementary Materials Appendix} % Start the appendix part
%\documentclass[Full_Paper.tex]{subfiles}
%\begin{document}
\renewcommand{\thefigure}{S\arabic{figure}}
\renewcommand{\thetable}{S\arabic{table}}
\renewcommand{\thesection}{S.\arabic{section}}
\setcounter{figure}{0}
\setcounter{table}{0}

%\begin{refsection}
\iftoggle{arxiv}{}{\linenumbers}

The primary goal of our analysis is to develop, evaluate and contextualize the performance of \methodname. In the following four supplementary sections we first summarize the methods used in this evaluation and then describe the data, the experiments conducted, and how \methodname\ compares to other approaches in the literature in greater depth. We also describe the intuition behind and the mechanics of \methodname's algorithms in greater detail. 

\iftoggle{arxiv}{
\parttoc % Insert the appendix TOC
}
{ \tableofcontents }
%\newpage
%%%%%%%%%%%%%%%%%%%%%%%%%%%%%%%%%%%%%%%%%%%%%%%%%%%%%%%%%%%%%%%%%%
                % SUPPLEMENT: DATA %
%%%%%%%%%%%%%%%%%%%%%%%%%%%%%%%%%%%%%%%%%%%%%%%%%%%%%%%%%%%%%%%%%%

% \textbf{NOTE: This is the order in which supplementary material are referenced in Sol's main text right now}
% \begin{enumerate}
%     \item Budgets of other remote sensing research projects
%     \item The solve
%     \item Costs of competing approaches
%     \item Featurization
%     \item Figure with errors from US predictions
%     \item Road data quality appendix section
%     \item Sensitivity to N and K
%     \item Checkerboard
%     \item Global model: solve, data quality, comparison of performance to the US model
%     \item Unclear what he wants here... ``We have examined performance of Goldenye over regular square images, but it can be implemented using training or testing data that is either irregular or inconsistently shaped (e.g. districts) by simply adjusting observation weights''. Maybe we just refer readers to the income section, given that those were irregular polygons? [Esther here - I get the point of this, it's that we could convolve over irregular shapes and then take the average and it would still make sense. Not sure where this should go, seems like more of a musing and not something we'll really want to do.]
%     \item api
% \end{enumerate}

%\clearpage 

\section{Methods summary}
\label{sec:main-methods}

Here we provide additional information on our implementation of \methodname\ and experimental procedures, as well as a description of the theoretical foundation underlying \methodname. Full details on the methodology behind \methodname\ can be found throughout Section \ref{sec:methods}. %Additional details are contained in the Supplementary Materials.

\subsection*{Implementation of \methodname}

%Each region centered at location index $\ell$ is observed via a satellite image $\Iell$.

%some prediction task $s$ (e.g. population density) 

%over a set of locations indexed by $\ell = \{1,\ldots, N\}$.

%available data on ground conditions $y_\ell^s$:

We begin with a set of images $\{\mathbf{\Iell\}}_{\ell=1}^{N}$, each of which is centered at locations indexed by $\ell = \{1,\ldots, N\}$. \methodname\ generates task-agnostic feature vectors $\bf{x}(\Iell)$ for each satellite image $\Iell$ by convolving an $M\times M \times S$ ``patch'', $\Pk$, across the entire image. $M$ is the width and height of the patch in units of pixels and $S$ is number of spectral bands. In each step of the convolution, the inner product of the patch and an $M\times M \times S$ sub-image region is taken, and a ReLU activation function with bias $b_k=1$ is applied. Each patch is a randomly sampled sub-image from the set of training images $\{\mathbf{\Iell\}}_{\ell=1}^{N}$ (Fig.~\ref{fig:featurization}). In our main analysis, we use patches of width and height $M=3$ (Fig.~\ref{fig:patch_size}) and $S=3$ bands (red, green, and blue). To create a single summary metric for the image-patch pair, these inner product values are then averaged across the entire image, generating the $k$th feature $\XkIell$, derived from patch $\Pk$. The dimension of the resulting feature space is equal to $K$, the number of patches used, and in all of our main analyses we employ $K = 8,192$ (i.e. $2^{13}$). Both images and patches are whitened according to a standard image preprocessing procedure before convolution (Section~\ref{sec:featurization}).

In practice, this one-time featurization can be centrally computed and then features $\XkIell$ distributed to users in tabular form. A user need only (i) obtain and link the subset of these features that match spatially with their own labels and then (ii) solve \textit{linear} regressions of the labels on the features to  learn \textit{nonlinear} mappings from the original image pixel values to the labels (the nonlinearity of the mapping between image pixels and labels stems from the nonlinearity of the ReLU activation function). 
We show strong performance across seven different tasks using ridge regression to train the relationship between labels $y_{\ell}$ and features $\XkIell$ in this second step, although future work may demonstrate that other fitting procedures yield similar or better results for particular tasks.

Implementation of this one-time unsupervised featurization takes about the same time to compute as a single forward pass of a CNN. With $K=8,912$ features, featurization results in a roughly 6 to 1 compression of stored and transmitted imagery data in the cases we study.\footnote{This is calculated as: $(256 * 256 * 3) / (8192 * 4) = 6\times$ compression, where 256 * 256 * 3 integer values per image are compressed into 8192 float32 features, each of which takes 4$\times$ the storage of an integer. Using 100 features gives a 500$\times$ compression.} Notably, storage and computational cost can be traded off with performance by using more or fewer features from each image (Fig.~\ref{Fig:Waffle}B). Since features are random, there is no natural value for $K$ that is specifically preferable.

\subsection*{Experimental procedures}

\paragraph*{Task selection and data}

Tasks were selected based on diversity and data availability, with the goal of evaluating the generalizability of \methodname\ (Section~\ref{sec:labeldata}). Results for all tasks evaluated are reported in the paper. We align image and label data by projecting imagery and label information onto a $\sim$1km $\times$ 1km grid, which was designed to ensure zero spatial overlap between observations (Sections~\ref{sec:grid_and_sampling} and \ref{sec:aggregatingRawLabelsToGC}). 
\iftoggle{arxiv}{}{Images are obtained from the Google Static Maps API (Section \ref{sec:imagary}) \cite{GoogleMaps}, and labels for the seven tasks are obtained from refs.} \cite{hansen2013, awstiles,gpw, viirs-nl, acs-income, tiger-roads, ztrax}. Details on data are described in Table~\ref{tab:labeldata} and Section~\ref{sec:data}.

\paragraph*{US experiments}

From this grid we sample 20,000 hold-out test cells and 80,000 training and validation cells from within the continental US (Section~\ref{sec:data_separation_uncertainty}). To span meaningful variation in all seven tasks, we generate two of these 100,000-sample data sets according to different sampling methods. First, we sample uniformly at random across space for the forest cover, elevation, and population density, tasks which exhibit rich variation across the US. Second, we sample via a population-weighted scheme for nighttime lights, income, road length, and housing price, tasks for which meaningful variation lies within populated areas of the US.  Some sample sizes are slightly reduced due to missing label data ($N=91,377$ for income, $80,420$ for housing price, and $67,968$ for population density). We model labels whose distribution is approximately log-normal using a log transformation (Section~\ref{sec:training_testing_model} and Table \ref{tab:logvlevel}). 

Because fitting a linear model is computationally cheap, relative to many other SIML approaches, it is feasible to conduct numerous sensitivity tests of predictive skill. We present cross-validation results from a random sample, while also systematically evaluating the behavior of the model with respect to: (a) geographic distance between training and testing samples, i.e. spatial cross-validation, (b) the dimension $K$ of the feature space, and (c) the size $N$ of the training set (Fig.~\ref{Fig:Waffle}, Sections~\ref{sec:secondary_analysis_nk} and \ref{sec:secondary_analysis_spatial}). We represent uncertainty in each sensitivity test by showing variance in predictive performance across different training and validation sets. We also benchmark model performance and computational expense against an 18-layer variant of the ResNet Architecture, a common deep network architecture that has been used in satellite based learning tasks \cite{Perez2017}, trained end-to-end for each task and a transfer learning approach \cite{Pan2010} utilizing an unsupervised featurization based on the last hidden layer of a 152-layer ResNet variant pre-trained on natural imagery and applied using ridge regression (Sections~\ref{sec:cnn_benchmark} and \ref{sec:cost_analysis}).

\paragraph*{Global experiment}

To demonstrate performance at scale, we apply the same approach used within the data-rich US context to global imagery and labels. We employ a target sample of $N=1,000,000$, which drops to a realized sample of $N=423,476$ due to missing imagery and label data outside the US (Fig.~\ref{Fig:Global}). We generate predictions for all tasks with labels that are available globally (forest cover, elevation, population density, and nighttime lights) (Section~\ref{sec:global_analysis}). 

\paragraph*{Label super-resolution experiment}

Predictions at label super-resolution (i.e. higher resolution than that of the labels used to train the model), shown in Fig.~\ref{Fig:Global}B, are generated for forest cover and population density by multiplying the trained ridge regression weights by the un-pooled feature values for each sub-image and applying a Gaussian filter to smooth the resulting predictions 
%using a kernel bandwidth of $\sigma$ = 16 pixels. 
(Section~\ref{sec:superresolution}). Additional examples of label super-resolution performance are shown in Fig. \ref{fig:superresolution_all10}. We quantitatively assess label super-resolution performance (Fig. \ref{Fig:superreslineplot}) using forest cover, as raw forest cover data are available at substantially finer resolution than our common $\sim$1km x 1km grid. Performance is evaluated by computing the fraction of variance ($R^2$) within each image that is captured by \methodname, across the entire sample.

\subsection*{Theoretical foundations}

\methodname\ is motivated by the goal of enabling generalizable and skillful SIML predictions. It achieves this by embedding images in a basis that is both descriptive (i.e. models trained using this single basis achieve high skill across diverse labels) and efficient (i.e. such skill is achieved using a relatively low-dimensional basis). The approach for this embedding relies on the theory of ``random kitchen sinks''~\cite{Rahimi2008}, a method for feature generation that enables the linear approximation of arbitrary well-behaved functions. This is akin to the use of polynomial features or discrete Fourier transforms for function approximation generally, such as functions of one dimension. 
When users apply these features in linear regression, they identify linear weightings of these basis vectors important for predicting a specific set of labels. 
With inputs of high dimension, such as the satellite images we consider, it has been shown experimentally~\cite{Morrow2017, Coates2012, Jonas2018} and theoretically~\cite{RahimiRecht2008} that a randomly selected subspace of the basis often performs as well as the entire basis for prediction problems.

\paragraph{Convolutional random kitchen sinks} Random kitchen sinks approximate arbitrary functions by creating a finite series of features generated by passing the input variables $z$ through a set of $K$ nonlinear functions $g(z; \Theta_k)$, each paramaterized by draws of a random vector $\Theta$. The realized vectors $\Theta_k$ are drawn independently from a pre-specified distributions for each of $k = 1...K$ features. Given an expressive enough function $g$ and infinite $K$, such a featurization would be a universal function approximator~\cite{RahimiRecht2008}. In our case, such a function $g$ would encode interactions between all subsets of pixels in an image. Unfortunately, for an image of size $256 \times 256 \times 3$, there are $2^{256 \times 256 \times 3}$ such subsets. Therefore, the fully-expressive approach is inefficient in generating predictive skill with reasonably concise $K$ because each feature encodes more pixel interactions than are empirically useful.

%Though such a featurization would be a universal function approximator given infinite data and features \cite{RahimiRecht2008}, it is overly-expressive, in the sense that the mapping $\Theta_k$ would encode more pixel interactions than are empirically useful.

%it would be both computationally infeasible and 
%Such a featurization would be both computationally infeasible and overly-expressive, in the sense that the model would have far more parameters than is empirically useful. %(the superior performance of more sparse $\Theta_k$ is shown in Supplementary Information \ref{sec:featurization}).

To adapt random kitchen sinks for satellite imagery, we use convolutional random features, making the simplifying assumption that most information contained within satellite imagery is represented in \textit{local} image structure. Random convolutional features have been shown to provide good predictive performance across a variety of tasks from predicting DNA binding sites~\cite{Morrow2017} and solar flares~\cite{Jonas2018} to clustering photographs~\cite{Coates2012} (kitchen sinks have also been used in a non-convolutional approach to classify individual pixels of hyper-spectral satellite data~\cite{PerezSuay2017}). Applied to satellite images, random convolutional features reduce the number of effective parameters in the function by considering only local spatial relationships between pixels. This results in a highly expressive, yet computationally tractable, model for prediction.  

Specifically, we create each $\Theta_k$ by extracting a small sub-image patch from a randomly selected image within our image set, as described above. These patches are selected independently, and in advance, of any of the label data. The convolution of each patch across the satellite image being featurized captures information from the entire $\RR^{256 \times 256 \times 3}$ image using only $3 * M^2$ free parameters for each $k$. Creating and subsequently averaging over the activation map (after a ReLU nonlinearity) defines our instantiation of the kitchen sinks function $g(z; \Theta_k)$ as $g(\Iell; \Pk, b_k) = \XkIell$, where $b_k$ is a scalar bias term. Our choice of this functional form is guided by both the structural properties of satellite imagery and the nature of common SIML prediction tasks, and it is validated by the performance demonstrated across tasks.
  
%The choice of a ReLU function for $g$ is driven by the demonstrated function approximation capabilities of linear combinations of ReLU transforms. Linear combinations of ReLUs applied to random affine functions of the input can reconstruct any well-behaved function $f$ as the number of functions goes to infinity \todomisc{find citation}. In the context of imagery and kitchen sinks, an affine function could be equivalently represented as the inner product of an image vector (i.e. the flattened array of values for all bands at all pixels) with a vector of the same length, plus a bias term. In other words, $\Theta_k$ would be a row vector, and the elements of both $\Theta_k$ and $b$ would be constructed randomly and independently.
%\textbf{}
%In practice, taking random ReLU/affine transforms of the entire image is an inefficient way to approximate any recoverable $f$ (from Eq.~\ref{Eq:RSProblem}), largely because the majority of spatial relationships that would be uncovered through the random affine transforms are not informative of the variables we wish to predict in SIML tasks. Similar to the use of convolutions to reduce the parameter space of Neural Networks, \methodname\ uses convolutions to impose structure to $\Theta_k$. While this reduces the space of possible functions that can be approximated by a kitchen sink approach, it greatly increases the efficiency of our ability to represent spatial and spectral relationships within the image that are relevant for common SIML tasks.  

\paragraph{Relevant structural properties of satellite imagery and SIML tasks} Three particular properties provide the the motivation for our choice of a convolution and average-pool mapping to define~$g$.

First, we hypothesize that convolutions of small patches will be sufficient to capture nearly all of the relevant spatial information encoded in images because objects of interest (e.g. a car or a tree) tend to be contained in a small sub-region of the image. This is particularly true in satellite imagery, which has a much lower spatial resolution that most natural imagery (Fig. \ref{fig:patch_size}).

Second, we expect a single layer of convolutions to perform well because satellite images are taken from a constant perspective (from above the subject) at a constant distance and are (often) orthorectified to remove the effects of image perspective and terrain. Together, these characteristics mean that a given object will tend to appear the same when captured in different images. This allows for \methodname's relatively simple, translation invariant featurization scheme to achieve high performance, and avoids the need for more complex architectures designed to provide robustness to variation in object size and orientation.

Third, we average-pool the convolution outputs because most labels in the types of problems we study can be approximately decomposed into a sum of sub-image characteristics. For example, forest cover is measured by the percent of total image area covered in forest, which can equivalently be measured by averaging the percent forest cover across sub-regions of the image. Labels that are strictly averages, totals, or counts of sub-image values (such as forest cover, road length, population density, elevation, and night lights) will all exhibit this decomposition. While this is not strictly true of all SIML tasks, for example income and average housing price, we demonstrate that \methodname\ still recovers strong predictive skill on these tasks. This suggests that some components of the observed variance in these labels may still be decomposable in this way, likely because they are well-approximated by functions of sums of observable objects.

%In summary, \methodname\ leverages the expressiveness of random kitchen sinks~\cite{Rahimi2008} to generate descriptive features, encoding much of the spatial structure contained within satellite imagery that is relevant to SIML prediction tasks. Because sub-image-scale convolutions succinctly encode much of the spatial structure relevant to SIML prediction, these features are found to achieve high performance across a diversity of SIML tasks.

\paragraph*{Additional interpretations}
The full \methodname\ platform, encompassing both featurization and linear prediction, bears similarity to a few related approaches. Namely, it can be interpreted as a computationally feasible approximation of kernel ridge regression for a fully convolutional kernel or, alternatively, as a two-layer CNN with an incredibly wide hidden layer generated with untrained filters. A discussion of these interpretations and how they can help to understand \methodname's predictive skill can be found in Section~\ref{sec:featurization}.

\section{Data}
\label{sec:data}
    This section describes the datasets we use to construct our ground truth labels across all seven of our tasks: forest cover, elevation, population density, nighttime lights, income, road length, and housing price.
\iftoggle{arxiv}{}{In addition, we describe the imagery used in the analysis.} In Section \ref{sec:label_agg} we detail our method for linking the labeled data for each outcome to the imagery (Fig. \ref{fig:aggregationOfLabels}).

In evaluating the ability of \methodname\ to generalize, we are interested in its ability to recover different types of variables, including: (i) variables that are averages of sub-image properties, (ii) variables that not directly observable through daytime imagery but are a function of visible objects in the image, such as nighttime lights, and (iii) variables that are an underlying factor that determines what material appears in the image, such as elevation. Labels may also be a combinations of (i)-(iii), such as housing price or household income. An advantage of \methodname\ is that it solves all these cases without any alteration of method. In the main text, we use the the same set of image features to predict all seven outcomes and, in principle, this set of features can be used to predict an unlimited number of outcomes (Section~\ref{sec:featurization}, so long as the outcomes and the images are aligned as described in Section ~\ref{sec:label_agg}).

For each task, we obtain an up-to-date and geographically complete publicly available datasource to match with the images. Most of these data are based on measurements from 2010 - 2015, though our data on population density draws from sources that date back as far as 2005 in order to achieve global coverage. 
\iftoggle{arxiv}{}{Our imagery data, from the Google Static Maps API (Section \ref{sec:imagary}), was mostly acquired in 2018, though in some cases images may be a few years older.} %It is possible that these minor differences in timing between measurement of our labels and imagery imagery explains a portion of the error in our model performance. 

    % table of data sources for all domains
    \begin{table}[!ht]
\centering
 \resizebox{\textwidth}{!}{
\begin{tabular}{ l l l p{5cm}}
\textbf{Task} & \textbf{Units}  & \textbf{Native resolution} & \textbf{Data source} \\
 \hline\hline
%&&& \\
Forest cover & \% forest cover & $\sim$30m $\times$ 30m  & \cite{hansen2013} \\ 
Elevation & meters  & $\sim$611.5m $\times$ 611.5m & \cite{awstiles} \\ 
Population density & people per sq. km. & $\sim$1km $\times$ 1km & \cite{gpw} \\ 
Nighttime lights & nanoWatts/cm$^2$/sr & $\sim$500m $\times$ 500m & \cite{viirs-nl} \\ 
Income & USD per household & census block group & \cite{acs-income}\\ 
Road length & meters & polyline & \cite{tiger-roads} \\ 
Housing price & USD per sq. ft. & geocoded point data & \cite{ztrax} \\
\hline
\end{tabular}}
  \caption{ \textbf{Data sources for all tasks.} Note that for all raster data sets (forest cover, elevation, population density, and nighttime lights) stated resolutions apply to grid cells located at the equator; raster size in Euclidean distance will vary with latitude.} 
  \label{tab:labeldata}
\end{table}

\iftoggle{arxiv}{\subsection*{Labels}}{\subsection{Labels}}
    \label{sec:labeldata}
        
Tasks were chosen to represent outcomes of classes (i)-(iii) above subject to the condition that high resolution and up-to-date label data is available across the US.  Below we describe these data sources. See Section \ref{sec:aggregatingRawLabelsToGC} and Fig.~\ref{fig:aggregationOfLabels} for a description of how we assign raw label data to images. 

\paragraph{\textbf{Forest cover}}

To measure forest cover, we use globally comprehensive raster data from ref. \cite{hansen2013}, which is designed to accurately measure forest cover in 2010. This dataset is commonly used to measure forest cover when ground-based measurements are not available \cite{Alkama2016,Carlson2018}. Forest in these data is defined as vegetation greater than 5m in height, and measurements of forest cover are given at a raw resolution of roughly 30m by 30m. These estimates of annual maximum forest cover are derived from a model based on Landsat imagery captured during the growing season. Specifically, the authors train a pixel-level bagged decision tree using three types of features: ``(i) reflectance values representing maximum, minimum and selected percentile values (10, 25, 50, 75 and 90\% percentiles); (ii) mean reflectance values for observations between selected percentiles (for the max-10\%, 10-25\%, 25-50\%, 50-75\%, 75-90\%, 90\%-max, min-max, 10-90\%, and 25-75\% intervals); and (iii) slope of linear regression of band reflectance value versus image date.'' These estimates of forest cover were derived using different spectral bands than we observe in our imagery, and using information about how surface reflectance changes over the growing season, which we did not observe. This gives us confidence that we are indeed learning to map visual, static, high-resolution imagery to forest cover, rather than simply recovering the model used in ref. \cite{hansen2013}.\footnote{These data can be accessed at: \\ \url{https://landcover.usgs.gov/glc/TreeCoverDescriptionAndDownloads.php}.} 

\paragraph{\textbf{Elevation}}

We use data on elevation provided by Mapzen, and accessed via the Amazon Web Services (AWS) Terrain Tile service. These Mapzen terrain tiles provide global elevation coverage in raster format. The underlying data behind the Mapzen tiles comes from the Shuttle Radar Topography Mission (SRTM) at NASA's Jet Propulsion Laboratory (JPL), in addition to other open data projects.

These data can be accessed through AWS at different zoom levels, which range from 1 to 14 and, along with latitude, determine the resolution of the resulting raster. To align with the resolution of our satellite imagery, we use zoom level 8, which leads to a raw resolution of 611.5 meters at the equator.\footnote{We accessed these data via the R function \texttt{get\_aws\_terrain} from the \texttt{elevatr} package. Code and documentation can be found here: \url{https://www.github.com/jhollist/elevatr}.}

\paragraph{\textbf{Population density}}

We use data on population density from the Gridded Population of the World (GPW) dataset \cite{gpw}. The GPW data estimates population on a global 30 arc-second (roughly 1 km at the equator) grid using population census tables and geographic boundaries. It compiles, grids, and temporally extrapolates population data from 13.5 million administrative units. It draws primarily from the 2010 Population and Housing Censuses, which collected data between 2005 and 2014. GPW data in the US comes from the 2010 census.\footnote{These data can be accessed at \url{http://sedac.ciesin.columbia.edu/data/collection/gpw-v4}}

\paragraph{\textbf{Nighttime lights}}

We use luminosity data generated from nighttime satellite imagery, which is provided by the Earth Observations Group at the National Oceanic and Atmospheric Administration (NOAA) and the National Geogphysical Data Center (NGDC). The values we use are Version 1.3 annual composites representing the average radiance captured from satellite images taken at night by the Visible Infrared Imaging Radiometer Suite (VIIRS). We use values from 2015, the most recent annual composite available. 

This composite is created after the Day/Night VIIRS band is filtered to remove the effects of stray light, lightening, lunar illumination, lights from aurora, fires, boats, and background light. Cloud cover is removed using the VIIRS Cloud Mask product. These values are provided across the globe from a latitude of 75N to 65S at a resolution of 15 arc-seconds. The radiance units are nW cm$^{-2}$ sr$^{-1}$ (nanowatts per square centimeter per steradian).

Like forest cover, these labels are themselves derived from satellite imagery. However, because they capture luminosity at night, while our satellite imagery is taken during the day, the labels for luminosity and the imagery used to predict luminosity represent independent data sources. %Many social, economic, and environmental characteristics of a given location may be visible during the day while not being captured at all in nighttime lights, and vice versa. 
Our ability to predict nighttime lights depends on how well objects visible during the day are indicative of light emissions at night.\footnote{These data can be accessed at \url{https://www.ngdc.noaa.gov/eog/viirs/download\_dnb\_composites.html\#NTL\_2015}.}

\paragraph{\textbf{Income}}

We use the American Community Survey (ACS) 5-year estimates of median annual household income in 2015. These data are publicly available at the census block group level, of which there are 211,267 in the US, including Puerto Rico. On average, block groups are around 38 km$^2$, though block groups are smaller in more densely populated areas.\footnote{These data are accessible using the \texttt{acs} package in R \cite{acsPackage}, table number B19013.}
%When multiple block groups lie within a grid cell, we calculate the "average income" of the grid cell as the area-weighted average of the median income within that grid cell (section \ref{sec:aggregatingRawLabelsToGC}).

\paragraph{\textbf{Road length}} \label{par:road_data}

We use road network data from the United States Geological Survey (USGS) National Transportation Dataset, which is based on TIGER/Line data provided by US Census Bureau in 2016. Shapefiles for each state provide the road locations and types, including highways, local neighborhood roads, rural roads, city streets, unpaved dirt trails, ramps, service drives, and private roads. Road types are indicated by a 5-digit code Feature Class Code which is assigned by the Census Bureau.\footnote{\url{https://www.census.gov/geo/reference/mtfcc.html}} The variable we predict is road length (in meters), which is computed as the total length of all types of roads that are recorded in a given grid cell. 

\begin{figure}[!t]
    \centering
    % Change the file to roadSegment_private_axis for a figure with axis
    \includegraphics[width=.78\textwidth]{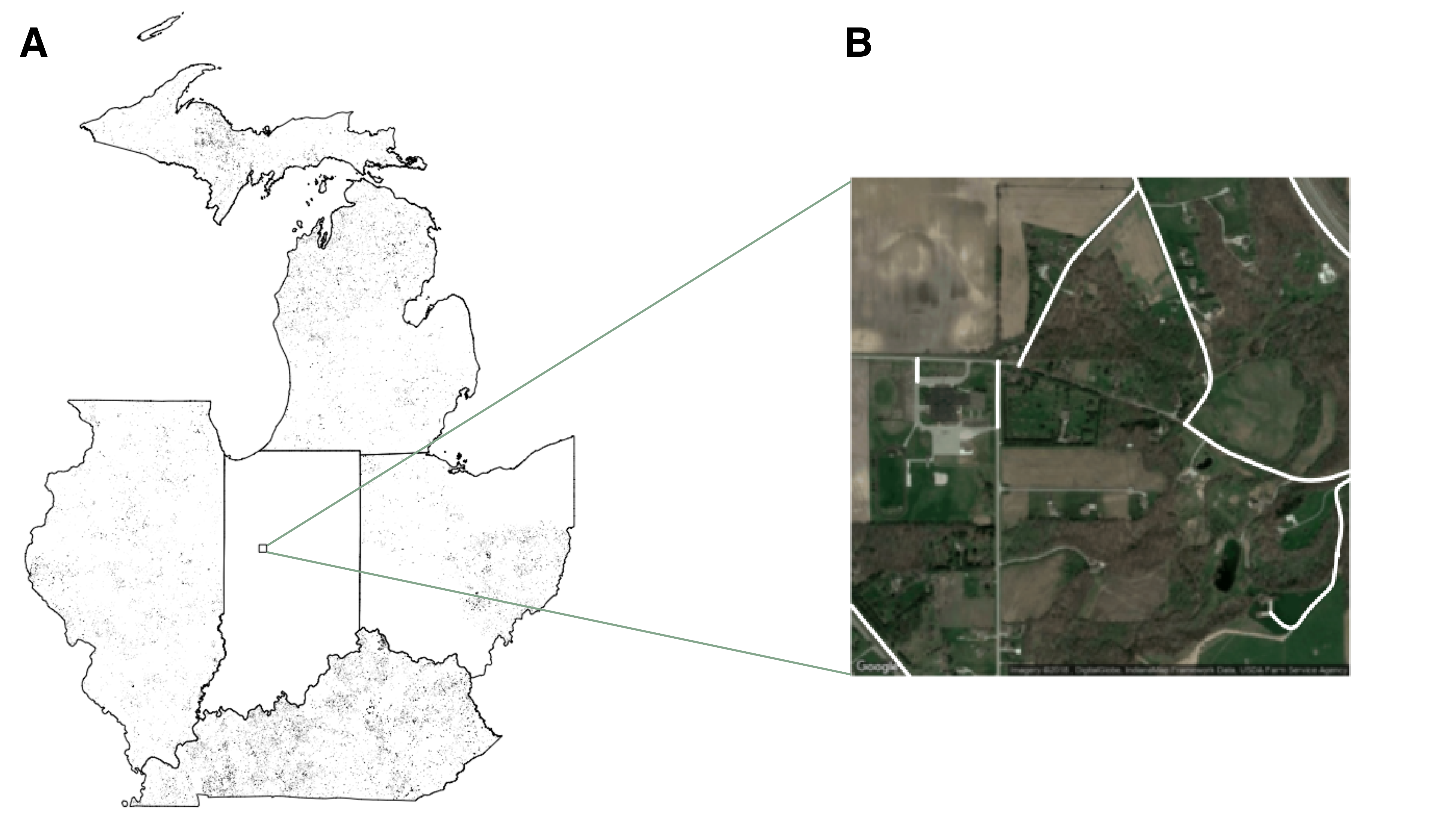}
    \caption{\textbf{Quality of ground truth road data varies by region.} (A) Private roads in the northern Midwest recorded in the USGS National Transportation Dataset. The conspicuous lack of recorded private roads in Indiana and sections of Ohio suggests that road data quality in certain regions may be lacking. (B) Overlaying recorded roads of all types (shown in white) over a single satellite image in Indiana, demonstrates that some roads that are easily visible from satellite imagery are missing in the available data that we use to construct labels.
    %For example, USGS reports no private roads in the area inside the box on the left figure, despite such roads being visible in satellite imagery (right figure).
    } %in 40.314, -86.754 (right), }
    \label{fig:private_roads}
\end{figure}

The Census Bureau database is created and corrected via a combination of partner supplied data, aerial images, and fieldwork. The spatial accuracy of linear features of roads and coordinates vary by source materials used. The accuracy also differs by region, causing cases in which some regions lack recordings of certain road types, the most common one being private roads and dirt trails. For example, private roads are rarely recorded in Indiana and some regions in Ohio (Fig. \ref{fig:private_roads}A), despite satellite images that suggest they are present (Fig \ref{fig:private_roads}B).\footnote{The data can be accessed at: \url{https://prd-tnm.s3.amazonaws.com/index.html?prefix=StagedProducts/Tran/Shape/}}

% reference: https://www2.census.gov/geo/pdfs/maps-data/data/tiger/tgrshp2017/TGRSHP2017_TechDoc.pdf

\paragraph{\textbf{Housing price}}

We estimate housing price per square foot using sale price and assessed square footage values for residential buildings. Data are provided by Zillow through the Zlllow Transaction and Assessment Dataset (ZTRAX). This dataset aggregates transaction and assessment data across the United States, combining reported values from states and counties with widely varying regulations and standards. Thus, significant data cleaning is required. Furthermore, because some states do not require mandatory disclosure of the sale price, we currently have limited data for the following states: Idaho, Indiana, Kansas, Mississippi, Missouri, Montana, New Mexico, North Dakota, South Dakota, Texas, Utah, and Wyoming. To address data quality issues, we develop a quality assurance and quality control (QA/QC) approach that is based on approaches employed in previous work~\cite{Moulton2018,Gindelsky2019,UnionofConcernedScientists2018} but adapted for our case.

 \begin{figure}[!t]
        \centering
        \includegraphics[width=.8\textwidth]{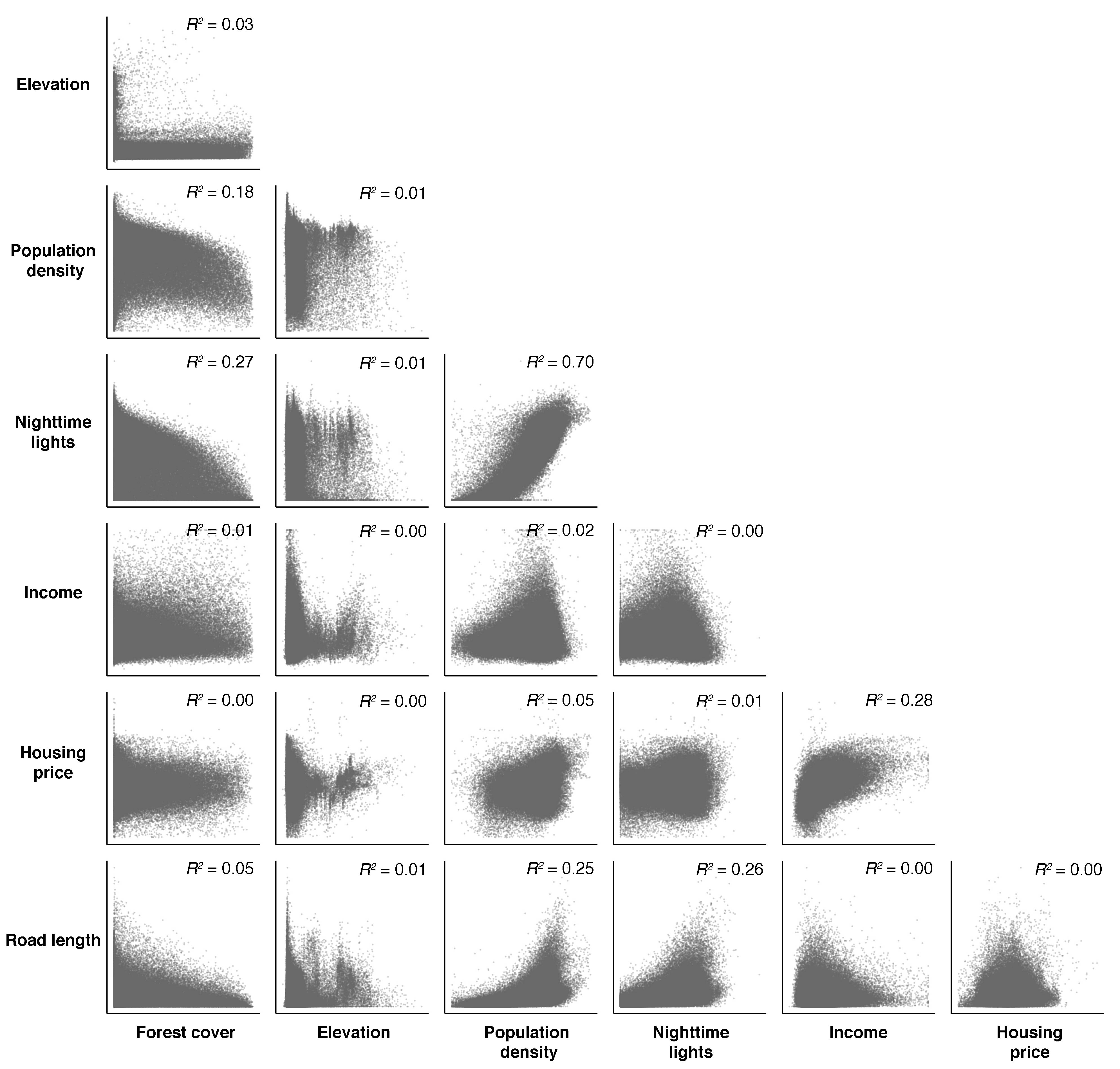}
        \caption{\textbf{Correlation of labels across tasks.} Each figure shows a scatter plot of labeled outcomes for one of our seven tasks against another. All points come from a population-weighted random sampling of grid cells (as described in Section \ref{sec:grid_and_sampling}) across the US. Scatters and $R^2$ values are shown across approximately 100,000 grid cell labels, depending on the data availability for each task.}
        \label{fig:corrys}
\end{figure}

ZTRAX contains data on the majority of buildings in the United States, initially comprising 374 million detailed records of transactions across more than 2,750 counties. The data is organized into two components - \textit{transaction data} and \textit{assessment data}. These two datasets are linked, allowing us to merge the latest sale price of a property to the latest assessment data. To minimize the effect of nation-wide trends in housing price that would be unobservable from our cross-sectional satellite imagery, we limit our dataset to sales occurring in 2010 or later. Further, we restrict our analysis to buildings coded as ``residential" or ``residential income - multi-family'' and drop any sale that was coded as an intra-family transfer. To obtain a square footage value, we follow the example in Zillow Research's GitHub repository~\cite{ZillowResearch} and take the maximum reported square footage for a given improvement, and then sum over all improvements on a given property.

To reduce the number of potentially miscoded outliers at the bottom end of the distribution of sale price and property size, we drop any remaining sales that fall under \$10,000 USD, any properties that fall under 100 sq. ft., and any \$/sq. ft. values under \$10. To address outliers on the high end of the distribution, we take this restricted sample and further cut our dataset at the 99th percentile of \$/sq. ft. by state. Afterwards, we select the most recent recorded sale price for each property (divided by the most recent assessed square footage). We then average across all of the remaining units within each grid cell to comprise our final dataset of housing price per square foot.

\begin{figure}[!t]
        \centering
        \includegraphics[width=.8\textwidth]{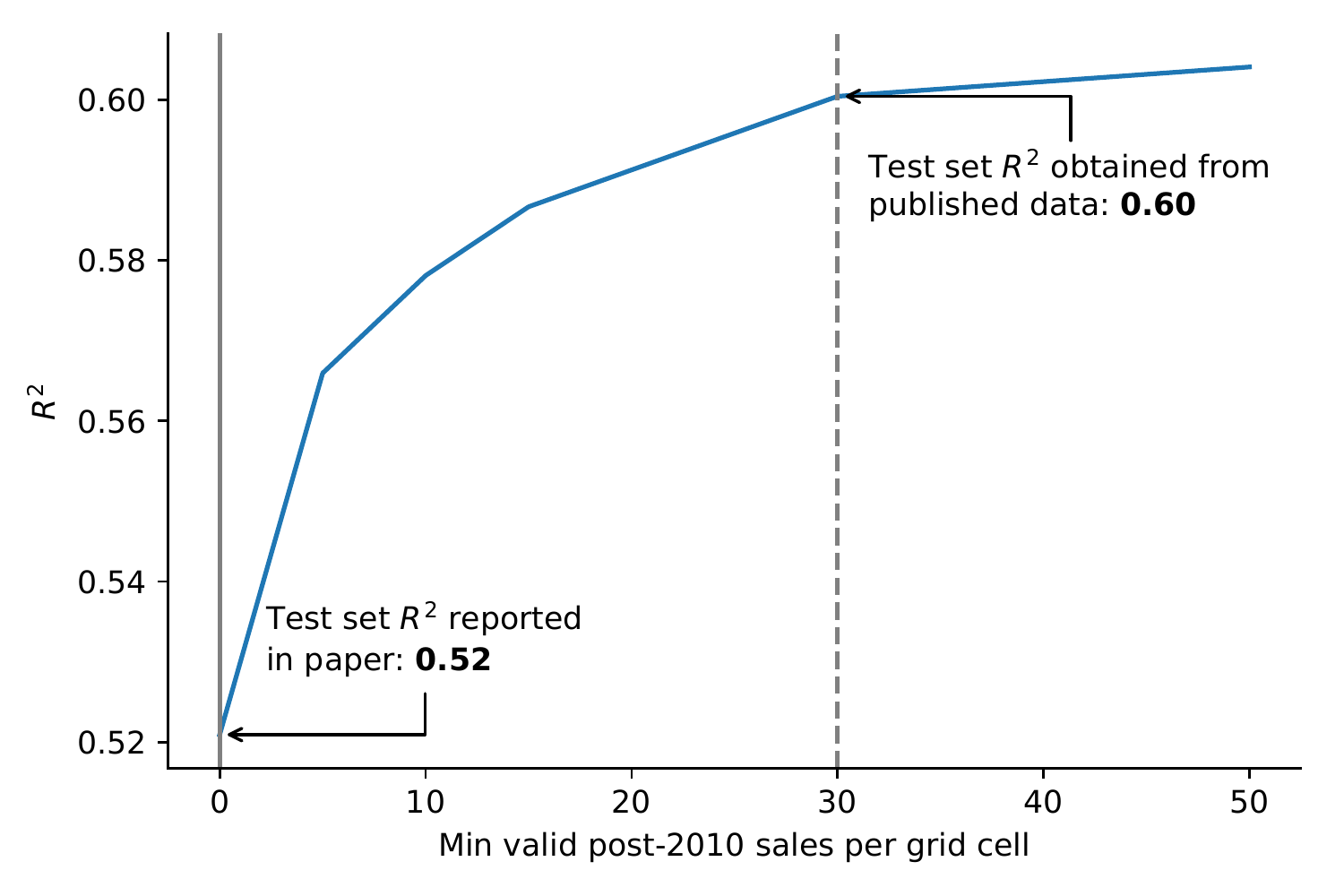}
        \caption{\textbf{Test set performance when restricting the dataset of mean housing price labels}. Curve shows the $R^2$ obtained in the test set for the housing price task in our main prediction experiment (Table ~\ref{SI:testSetTable}), when removing data with low numbers of valid, recent sales of buildings within the associated grid cell. The dashed line indicates the restriction applied to the publicly available dataset.}
        \label{fig:housing-threshold}
\end{figure}

To protect potentially identifiable information, our public data release contains housing price labels only for grid cells that contain 30 or more sales meeting the aforementioned criteria. This reduces the size of the dataset from $N=80,420$ to $N=52,355$ and makes the model performance obtainable by users better than that stated in the main text. For example, the public dataset will yield a test set $R^2$ of 0.60, rather than 0.52 (Table~\ref{SI:testSetTable}). This could be due to the fact that the \textit{average housing price} label we train on is noisier when estimated in a grid cell with few valid sales prices. It could also be because the \textit{average housing price} of areas with few recent sales may be inherently harder to predict via satellite imagery than that of areas with a greater number of recent sales. Fig.~\ref{fig:housing-threshold} empirically demonstrates the performance effect of removing grid cells with few recent sales.

\paragraph{\textbf{Correlation of outcomes across tasks}}

The seven tasks described above were selected in order to evaluate the performance of \methodname\ across many diverse contexts. Figure \ref{fig:corrys} evaluates the extent to which this was achieved, by plotting label values against one another. A few of the labels are moderately correlated, most notably population density and nighttime lights, but in general there is substantial orthogonal variation across these seven tasks. %\methodname\ captures meaningful signal across each outcome variable

%\clearpage

\iftoggle{arxiv}{}{
    \subsection{Imagery}
    \label{sec:imagary}
        \input{appendix_files/imagery.tex}}

%%%%%%%%%%%%%%%%%%%%%%%%%%%%%%%%%%%%%%%%%%%%%%%%%%%%%%%%%%%%%%%%%%
                % SUPPLEMENT: METHODS %
%%%%%%%%%%%%%%%%%%%%%%%%%%%%%%%%%%%%%%%%%%%%%%%%%%%%%%%%%%%%%%%%%%
%\clearpage 
 \section{Methods}
 \label{sec:methods}
    \phantomsection
     This section describes the methods that we use to define samples (Section \ref{sec:grid_and_sampling}), to construct labels (Section \ref{sec:label_agg}), and to construct features (Section \ref{sec:featurization}) for each image. It then describes how we separate data for training and evaluation (Section \ref{sec:data_separation_uncertainty}), train models (Section \ref{sec:training_testing_model}),  test predictive skill (Section \ref{sec:test_set_performance}), test sensitivity to the dataset size (Section \ref{sec:secondary_analysis_nk}) and test model extrapolation performance (Section \ref{sec:secondary_analysis_spatial}). Next, we describe tests of model performance at sub-label or ``super'' resolution as well as at the global scale (Sections \ref{sec:superresolution} and \ref{sec:global_analysis}).
%Finally, we discuss how the remote sensing algorithms used in this paper compare to other state of the art approaches in the literature (Section~\ref{sec:cnn_benchmark}). 
    
     \subsection{Grid definition and sampling strategy}
     \label{sec:grid_and_sampling}
         \paragraph{\textbf{Grid definition:}} To evaluate the generalizability of \methodname\ performance across tasks we need a standardized unit of observation to link raw labels for all tasks and imagery.  To do this, we construct a single global grid onto which we project both satellite imagery and labeled data. We design the grid to match our source of satellite imagery to ensure adjacent images do not overlap.  Each element of the grid, i.e. each ``grid cell,'' was designed to be a square in physical space. Because the earth is a sphere, the angular extent of grid cells changes across latitudes.\footnote{For the continental US (spanning 25 to 50 degrees latitude and -125 to -66 longitude), the grid cells are 0.0138 degrees in width (1.39 km) at the southern edge of the grid, and 0.0138 degrees in width (0.98 km) at the northern edge of the grid. The grid cells are 0.012 degrees in height (1.39 km) at the southern edge of the grid, and 0.0089 degrees in height (.98 km) at the northern edge of the grid.}

\paragraph{Sampling strategy:} For our primary experiment in the continental US we subsample sets of 100,000 observations, roughly 1.25\%  of the grid cells in the continental US, using two distinct sampling strategies.\footnote{We discard marine grid cells, but do not discard grid cells that are composed only of lakes or smaller inland bodies of water.}
 First, we sample uniformly-at-random (UAR) from all grid cells within the continental US. This sampling strategy is most appropriate for tasks like forest cover, where there is meaningful variation in most regions of the country. Second, we implement a population-weighted (POP) sampling strategy. To generate this sample, each grid cell is weighted by population density values taken from Version 4 of the Gridded Population of the World dataset, which provides a raster of population density estimates for the year 2015.\footnote{These data are available at \url{http://sedac.ciesin.columbia.edu/data/collection/gpw-v4/sets/bro}} This weighted sampling strategy is most applicable to tasks like housing price, where the most meaningful variation lies in more populated regions of the US. We use the UAR grid when sampling population density to avoid any issues that might arise from sampling a task using the same variable as sampling weights. In both the UAR and POP samples, we randomly sample just once; all results in the paper are displayed using the same two subsets of the full grid. Note that these sub-sampled grid cells, by construction, are each covered by exactly one satellite image %In principle, sampling is not crucial, but it helps us evaluate performance 
 %and focus on gross patterns of variation 
 without having to process data over the entire US.

%We discard all grid cells in this rectangular grid that contain only ocean, so that the resulting grid contains only points with some land mass in the contiguous United States. \footnote{We do not discard grid cells that are composed only of lakes or smaller inland bodies of water.} The resulting grid of the continental U.S. comprises roughly 8 million grid cells which are on average just over 1km$^2$ in size. To generate a dataset of manageable size for both training and testing our predictive model, we subsample sets of 100,000 observations -- roughly 1.25 \% -- from this grid using two distinct sampling strategies. First, we sample uniform-at-random (UAR) from all grid cells within the continental U.S.. This sampling strategy is most appropriate for applications like forest cover, elevation and population, where there is meaningful variation in most regions of the country. Second, we implement a population-weighted sampling strategy. To generate this sample, each grid cell is weighted by population density values taken from Version 4 of the Gridded Population of the World dataset, which provides a raster of population density estimates (data available at \url{http://sedac.ciesin.columbia.edu/data/collection/gpw-v4/sets/bro}) for the year 2015. This weighted sampling strategy is most valuable for applications like nighttime lights, income, roads and housing, where the most meaningful variation lies in more populated regions of the US. In both cases, we sample just once; all results in the paper are displayed using the same two subsets of the full grid. 

In our main results, we use the UAR sample for the forest cover, elevation, and population density tasks. We use the POP sample for nighttime lights, income, road length, and housing price. %Performance for each application under both sampling strategies is shown in Section \ref{sec:samplingrobustness}.
See Section \ref{sec:global_analysis} for a discussion of how we extend this grid and sampling procedure to the global scale. 

     \subsection{Assigning labeled data to sampled imagery}
     \label{sec:label_agg}
         \label{sec:aggregatingRawLabelsToGC}
\begin{figure}[t]
        \centering
        \includegraphics[width=1\textwidth]{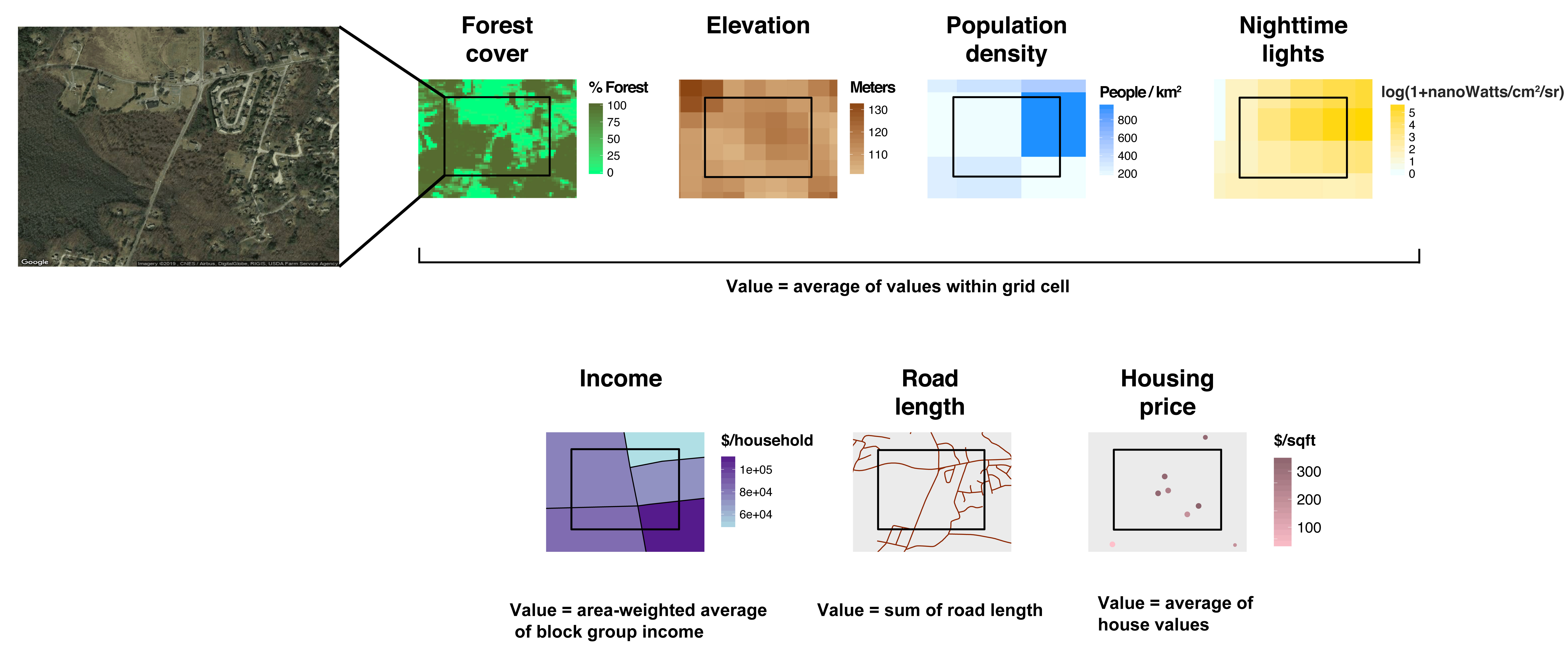}
        \caption{\textbf{Calculation of grid cell labels from raw data.} We calculate labels by spatially overlaying our grid cells and raw labeled data. We calculate labels as the average of raw label values that fall within the grid cell, except for roads where we calculate the label as the sum of road length within the grid cell.}
        \label{fig:aggregationOfLabels}
\end{figure}

To assign labels to each grid cell, we spatially overlay our raw labeled data and our custom grid. The native format and spatial resolution of the labeled data vary across the tasks studied, necessitating different aggregation or disaggregation procedures for each task. Here, we describe the approach taken in each task (Fig. \ref{fig:aggregationOfLabels}). %In all tasks except income, our labeled data are higher resolution than the grid cells in our rectangular grid. In these cases, we aggregate labeled data up to the grid cell level. 

The raw forest cover, elevation, population density and nighttime lights data are provided natively as rasters with higher spatial resolution than our custom grid. For these tasks, we perform aggregation by calculating the mean of all labeled pixels with centroids that fall within the imagery grid cell. The resulting labels indicate mean forest cover, mean elevation, mean population density, and mean nighttime lights across the image grid cell.

Our road length data are provided as high-resolution spatial line segments. To aggregate these data to the image grid cell, we calculate the sum of road length segments within each image. The resulting labels indicate the total length of recorded roads that fall within an image grid cell. 

Our housing price data are available as individual geocoded house sales. We aggregate these geocoded prices to the image grid cell by taking the average housing price per square foot across all sale prices that fall within the extent of the image. The resulting labels indicate the average housing price per square foot across all observed houses within a grid cell.

Our income data are provided at the block-group level (see Section \ref{sec:labeldata} for details). In some parts of the U.S., these block-groups are larger in total area than our image grid cells. However, in other regions, block-groups are smaller than our image grid cells. To treat both cases consistently, we aggregate incomes to the grid cell level by taking the weighted average of block-group incomes, where the weights are the area of intersection between the image grid cell and the block-group polygons. These weights are normalized to unity for each grid cell. The resulting labels indicate the area-weighted average median income across the grid cell.

Future users of a production-scale version of \methodname\ would employ label data of arbitrary format and resolution. The above approaches provide guidelines for how to match various forms of label data to the pre-computed image feature grid, but other methods may be used. In the simplest case, for example, sparse point data could be directly matched to the nearest grid cell centroid.

     \subsection{Featurization of satellite imagery}
     \label{sec:featurization}
         
\paragraph{\textbf{Notation}} In our context, the input variable $z$ is a set of satellite images $\mb{I}$, each corresponding to a physical location, $\ell$. We use brackets to denote indexing into images, with colons denoting sub-regions of images (e.g. $\Iell[i,j]$ is the $(i,j)^{th}$ pixel of image $\Iell$, $\Iell[i:i+M, j:j+M]$ is the square sub-image of size $M \times M$ starting at pixel $(i,j)$.) Because images have a third dimension (spectral bands), a colon $\Iell[i,j,:]$ denotes all bands at pixel $(i,j)$.  Indexing into non-image objects is denoted with subscripts (e.g. the $k^{th}$ element of vector $\x$ is denoted as $\xk$ and the $k^{th}$ patch in a set of patches $\mathbf{P}$ is denoted as $\Pk$). We denote inner products with angular brackets $\langle \cdot, \cdot \rangle$  and the convolution operator with $\conv$.

\paragraph{Connection to the kitchen sinks framework} The \textit{random kitchen sink} featurization used in \methodname\ relies on a nonlinear mapping $g(z; \Theta_k)$, where $z$ is an input variable and $\Theta_k$ is a randomly drawn vector. Here, we describe the implementation details of this featurization in the context of satellite imagery. Connecting our implementation and notation to the framework of random kitchen sinks, the random variables $\Theta_k$ are instantiated as the values of a random patch $\Pk$ and the bias $b_k$. The input variable $z$ is an image $\Iell$, and $g(z; \Theta_k)$ represents the convolution of the patch over the image, followed by addition of the bias $b_k$ and application of a element-wise ReLU function and an average pool, as described in the Methods of the main article and detailed below. %we construct the features for \methodname\ in greater detail.

% %% OLD VERSION in case people hate my changes: 
% \paragraph{\textbf{Notation}} We use brackets to denote indexing into images, with colons denoting sub-regions of images (e.g. $\Iell[i,j]$ is the $(i,j)^{th}$ pixel of image $\Iell$, $\Iell[i:i+M, j:j+M]$ is the square sub-image of size $M \times M$ starting at pixel $(i,j)$.) Because images have a third dimension (i.e. spectral bands), a colon $\Iell[i,j,:]$ denotes all bands at pixel $(i,j)$. %\footnote{This is consistent with python syntax for indexing into matrices, except that we use a 1-index rather than a 0-index to denote the first element along a dimension.} 
% Indexing into non-image objects is denoted with subscripts (e.g. the $k^{th}$ element of vector $\x$ is denoted as $\xk$ and the $k^{th}$ patch in a set of patches $\{\Pk\}$ is denoted as $\Pk$). Angular brackets $\langle \cdot, \cdot \rangle$ denote dot products and $\conv$ denotes the convolution operator.

 \begin{figure}[t]
        \centering
        \includegraphics[width=0.9\textwidth]{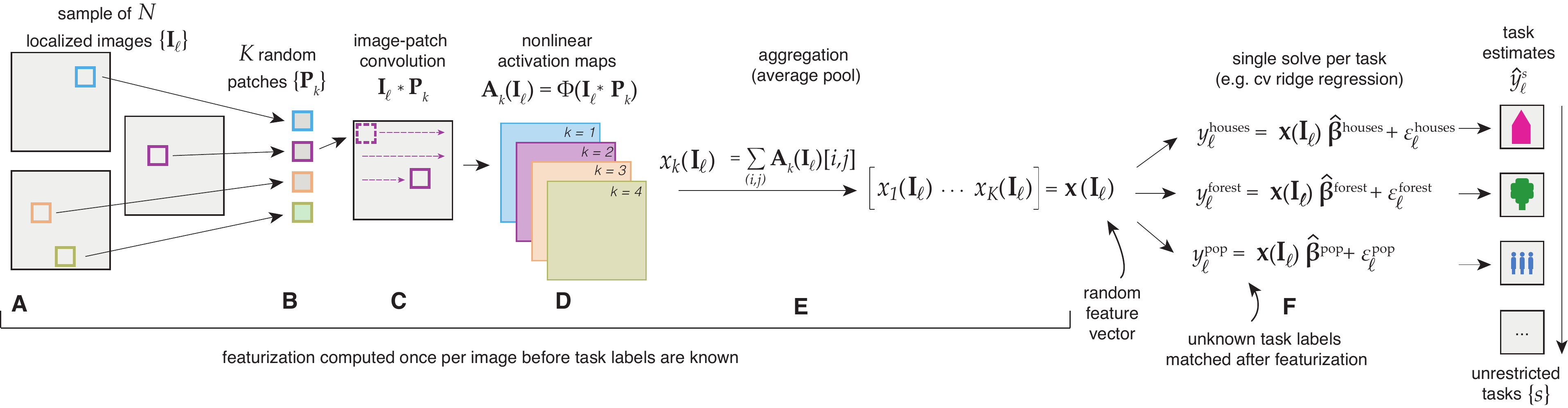}
        \caption{\textbf{\methodname\ process from featurization to multi-task prediction.} Given a large sample of $N$ satellite images (A), a random sample of $K$ patches (B) are drawn. (C) These $K$ random patches $P_k$ are convolved over each image $\Iell$ and (D) passed through a nonlinear function $\phi(\cdot)$ to generate $K$ activation maps. (E) Pixel-specific activations are pooled across each image to generate one set of $N \times K$ features that are stored and distributed to all users. (F) The same random feature vector $x$ is used in cross-validated ridge regression across many distinct tasks, after labeled and geo-referenced data $y_\ell$ is matched to features from each image $\Iell$ (as shown in Fig. 1B of the main text). (G) Models trained via ridge regression can be used to generate predictions across unrestricted tasks for any location with satellite imagery. }
        \label{fig:featurization}
\end{figure}

\begin{figure}[t]
        \centering
        \includegraphics[width=.5\textwidth]{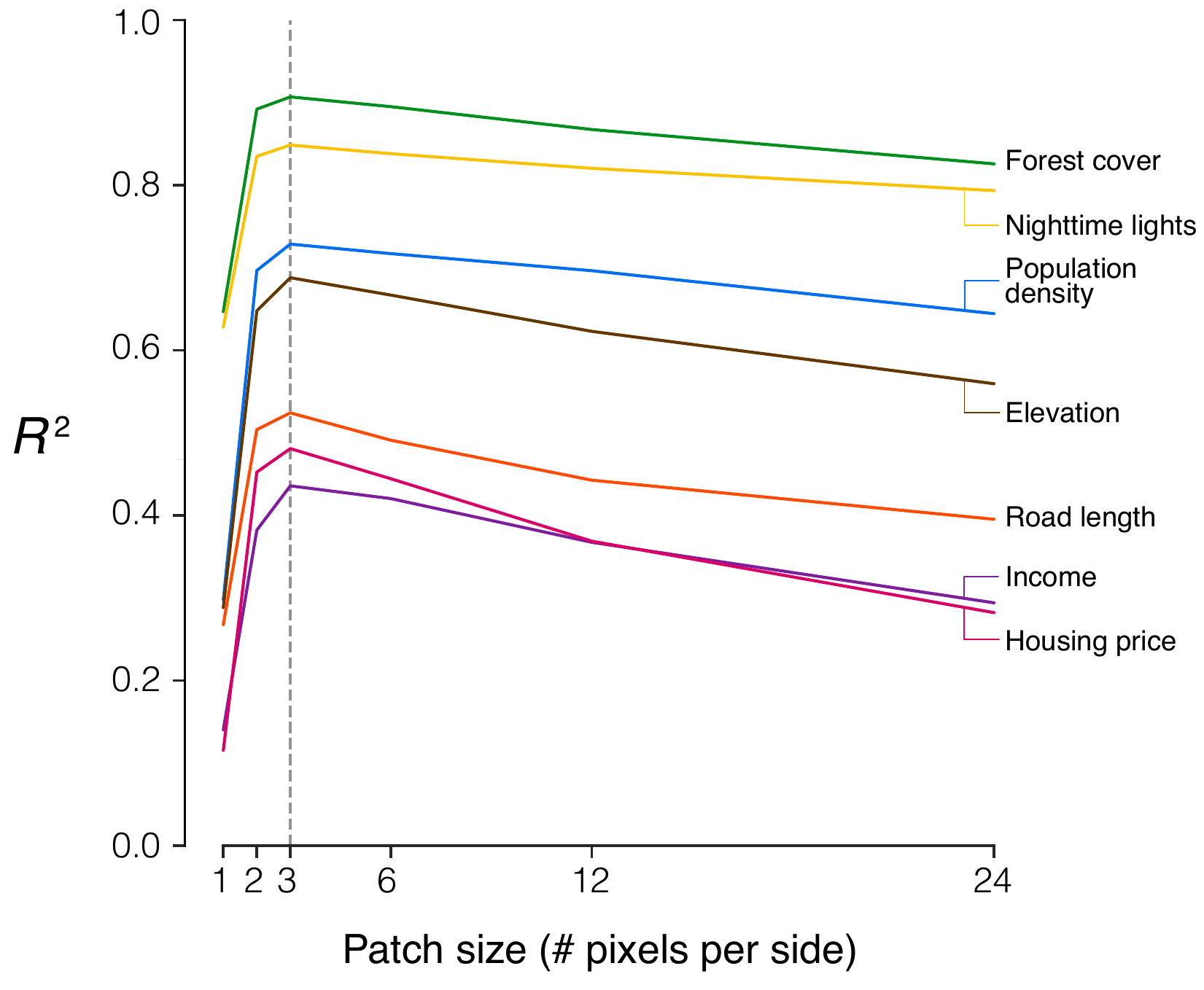}
        \caption{\textbf{Performance by patch size.} Featurization in \methodname\ relies on convolving an $M\times M\times 3$ patch $\Pk$ across satellite images. $M$ indicates the width in pixels of each sub-image patch, and the third dimension indexes the 3 spectral bands used throughout the analysis in this paper (an analogous approach can be applied to hyperspectral data). This figure shows, for each task, test set $R^2$ for patch sizes $M=1, 2, 3, 6, 12$ and $24$, using $K=2,048$ features for each $M$. The dotted gray line indicates the benchmark model used throughout the paper, with $M=3$. %Forest cover, elevation, and population density tasks use a uniform at random sample of training and test set images, while nighttime lights, road length, income, and housing price use a population weighted sample.
        }
        \label{fig:patch_size}
\end{figure}

\paragraph{\textbf{Methodological Details}} Fig.~\ref{fig:featurization} depicts our featurization process. As described in Section \ref{sec:labeldata} and \ref{sec:grid_and_sampling}, we begin with two sets (uniform and population-weighted samples) of $N=100,000$ satellite images, each of which measures $640 \times 640 \times 3$ pixels (the third dimension represents the visible red, green, and blue spectral bands). We then coarsen the images to $256 \times 256 \times 3$ pixels to reduce computation. Next, we draw $K/2 = 4,096$ small sub-image ``patches'' of size $M \times M \times 3$ uniformly at random from the 80,000 images that comprise our training and validation set, and calculate the negative of each patch to get another $4,096$ patches (Fig.~\ref{fig:featurization}A,~\ref{fig:featurization}B).
Our chosen specification sets $M=3$,
%and utilizes three bands (RGB), 
so that each patch $\Pk$ is of dimension $3 \times 3 \times 3$ (see Fig.~\ref{fig:patch_size} for performance in experiments using different patch sizes).

We then ``whiten" each patch by zero components analysis (ZCA), a common pre-processing routine in image processing~\cite{krizhevsky2009learning}. ZCA whitening pre-multiplies each patch by a transformation such that the resulting empirical covariance matrix of the whitened patches is the identity matrix. %This whitening step can be thought of as de-correlating the set of features, such that the resulting features are non-redundant\todomisc{IB: Is this definitely true? Or would it be more accurate to say something like: "...de-correlating the pixels within a $M \times M \times 3$ patch, which has been demonstrated in practice to offer a better-performing feature space (maybe needing citation)."}. A complementary motivation for whitening is that it fixes the first and second moments of the distribution of within-patch pixel distributions to match a standard Gaussian, while leaving the higher-order moments unconstrained (so that they may still match the original distribution of patches). [TBH i think this is still too much musing cause whitening is pretty standard. IB: I agree. I think the below sentences are enough].
We then convolve each patch $\Pk$ over each of the $N$ images (Fig.~\ref{fig:featurization}C) to obtain a set of $254 \times 254 \times 1$ pixel matrices for each image $\Iell$\footnote{To improve efficiency of the featurization process, our implementation calculates the inner product of patch and image only for the original $K/2$ patches. We then create an additional $K/2$ values equal to the negative of each of the original inner products.}.% these matrices are equivalent to the \emph{linear} activation map $\Theta_k z$.
  During the convolutions each $3\times 3 \times 3$ sub-image $\Iell[i:i+M,j:j+M,:]$ is also whitened according to the same whitening matrix as is applied to the patches.\footnote{In practice, we apply the whitening operator as a right multiplication to the original $8192 \times 27$ whitened patch matrix in order to reduce computation.} 
We then apply a pixel-wise nonlinearity operator $\Phi$ to each resulting matrix to obtain $K$ \emph{nonlinear} activation maps $\Ak (\Iell) = \Phi(\Pk \conv \Iell + \mb{b}_k)$ for each image $\Iell$
(Fig.~\ref{fig:featurization}D) so that the $(i,j)^{th}$ pixel of the $k^{th}$ activation map is defined as
\begin{align*}
    \Ak (\Iell) [i,j] = 
    \Phi(\langle \Iell [i:i+M,j:j+M, :], \Pk \rangle + {b}_k),
\end{align*}

where $b_k$ is a bias term from the constant bias matrix $\mb{b}_k$, in which every element is equal to $b_k = 1$. We use $\Phi(\Iell; \Pk, \mb{b}_k ) = \reLu(\Pk \conv \Iell+\mb{b}_k) := \max\{\Pk \conv \Iell + \mb{b}_k, 0\}$ as the nonlinear operator. We then aggregate across the image by taking the average of the nonlinear activation maps (Fig.~\ref{fig:featurization}E). The combination of the nonlinear operator $\Phi(\cdot)$ and average pooling composes the function $g(\cdot)$ above, and creates a scalar value for each patch $k$ and image $\ell$ pair:
\begin{align}
\label{eq:feat_vector}
    \XkIell = \frac{1}{254^2}\sum_{i=1}^{254} \sum_{j=1}^{254} \Ak(\Iell)[i,j]
\end{align}
Stacking these scalars across all $K$ patches provides the resulting $K$-dimensional feature vector, $ \XIell := \begin{bmatrix}
\mb{x}_{1}(\Iell) &\mb{x}_{2}(\Iell)& ... & \mb{x}_{K}(\Iell)
\end{bmatrix} \in \mathbb{R}^K$. This featurization thus embeds the original image $\Iell$ into a $K$-dimensional feature space, which can then be mapped to many different outcomes using task-specific models ($s$) implemented by researchers ($r$): $y_\ell^{s,r} = \XIell{\bm\beta}^{s,r} + {\epsilon}_\ell^{s,r} $, as illustrated in Fig.~\ref{fig:featurization}F. This linear relationship between labels and features may express a relationship between labels and image pixels that is highly nonlinear because the features themselves are nonlinear with respect to the images.

\paragraph{\textbf{Patch size and sampling}} 
We approximate the idealized complete convolutional basis, which contains features for all patch sizes, with the simpler truncated basis where we use only a single patch size. Throughout our main analysis, we use a $3 \times 3 \times 3$ patch size for $\Pk$. While larger patches may, in principle, enable the detection of image features with larger spatial structure, we find that, in practice, patch size $M=3$ performs best across all seven tasks (Fig.~\ref{fig:patch_size}). This finding suggests that most information contained within satellite imagery of this resolution can be represented by local-level image structure, and that the inclusion of ``non-local'' relationships reduces the efficiency of the function approximator by introducing more degrees of freedom. This empirical finding is consistent with previous applications of kitchen sink features \cite{Simonyan2015}. 

We draw patches randomly from the empirical distribution of $M \times M \times 3$ patches from our training data set of satellite images. Drawing patches from the empirical distribution, rather than generating them randomly, allows us to sample efficiently from the distribution of sub-images we will encounter in the sample.  This patch selection process is almost identical to the filter selection methods described in refs. \cite{Coates2011, recht2019imagenet, agarwal2013least}. 
%We select patches randomly from the distribution of our satellite images, crather than designing a tailor-made set of patches for several reasons. First, randomly drawing patches from images is conceptually simple, easy to implement, and computationally inexpensive. Second, it performs well in practice \todovaishaal{Is there somewhere we can cite that choosing patches from the set of images performs well (besides ourselves)?}. This is likely because, with enough random draws, the patches span the landscape of informative sub-images. Third, random patch generation prevents us from being in any way task-specific with our feature generation. Rather than explicitly looking for patches that correspond to trees, buildings, or roads, we assembled the set of patches before choosing the set of tasks to study. 
It may be valuable for future research to explore whether \methodname\ performance and computational efficiency could be improved through patch selection algorithms. For example, one goal in selecting patches-based features is to promote relative sparsity in the resulting patch-based features, as in ref. \cite{Romero2016}. However, any attempt to tailor patch selection or featurization to a particular task of interest requires sacrificing the generalizability of this task-agnostic featurization. It remains an open question whether a non-randomly selected set of basis patches could potentially achieve similar (or greater) performance than what we present here when applied to arbitrary new tasks.

\paragraph{Alternative interpretations relating \methodname\ to kernels and CNNs}
The methods summary describes how \methodname's convolutional random features enable nonparametric approximations of nonlinear functions through an embedding in a rich basis that expresses local spatial relationships.  Here, we provide two alternative interpretations of the approach, the first relating to kernel methods and the second relating to convolutional neural networks. We believe these interpretations can provide useful lenses to consider why \methodname\ works, and may also be helpful to researchers thinking about related problems. 

First, one could interpret the design of \methodname\ as if we were attempting to design a computationally tractable approximation to implementing a ridge regression using a convolutional kernel and the kernel trick. Under this interpretation, one could arrive at the same design of \methodname\ using the following logic: (i) Design a kernel that allows us to describe the ``similarity'' of every image to every other image in the sample. (ii) For any new task, we want to use a kernel regression to predict the unobserved labels of new images based on their similarity to all other images---specifically,  predicted labels would be a weighted sum of all observed labels using weights determined by this kernel-based measure of image similarity, i.e. the kernel trick. (iii) Unfortunately, calculating such a kernel exactly would be computationally intractable on a data set as large as the one we use, so instead use convolutional kitchen sinks (i.e. the featurization in \methodname) to approximate the desired kernel regression. This last step follows from prior work demonstrating two concepts. First, random features can approximate the lifted feature space induced by well-known kernels~\cite{Rahimi} as the number of random features increases. Second, convolutions of random patches drawn from joint Gaussian distributions has been proven to approximate, in the limit, a kernel in which every sub-image from one image is compared with every sub-image from another using an arc-cosine distance function~\cite{Daniely2016}. Thus, convolutions with random patches should, in the the limit, approximate a kernel that compares every sub-image with every other sub-image in the sample. However, because our distribution of patches is drawn from training imagery, rather than from Gaussian distributions, there is not an analytical expression that is known for the kernel being approximated by \methodname\ in the limit. 

The above logic would arrive at a design essentially the same as \methodname, although it is not our preferred motivation or interpretation of why \methodname\ works because it is a more complicated rationale than is needed. Ref. \cite{Rahimi2008} showed that the existence of an associated kernel is not necessary for performance using kitchen sinks. Rather, it is simply the embedding of an input in a descriptive basis that provides the predictive skill, the insight that motivates our preferred---and we think simpler---interpretation presented in the main text. Nevertheless, the interpretation of \methodname\ in the context of kernels motivates one way to understand the mechanism through which \methodname\ achieves predictive skill at low computational cost. Namely, it enables the approximation of a nonparametric kernel regression, using some (unknown) fully convolutional kernel that is sufficiently rich to represent meaningful similarity between images but costly enough to prohibit a direct application of the kernel trick.

An additional way to contextualize \methodname\ is in terms of its computational elements. In particular, \methodname\ uses image convolutions and nonlinear activation operations common to convolutional neural networks (CNNs)~\cite{alber2017}. Indeed, \methodname\ is mathematically identical to the architecture one would arrive at if one designed a very shallow and very wide CNN without using backpropogation and instead using random filters. Specifically, \methodname\ could be viewed as a two-layer CNN that has an 8,192-neuron wide hidden layer with untrained weights that are randomly initialized by drawing from sub-images in the sample, and that uses an average-pool over the entire image. In contrast to the conventional CNN approach of optimizing weights (via backpropogation), the random initialization with no subsequent optimization significantly reduces training time and avoids numerical challenges associated with non-convex optimization procedures (such as vanishing gradients). Thus, in the main text, we do not frame \methodname\ as a CNN because \methodname\ does not exploit the primary benefits of a deep CNN, since \methodname\ lacks intermediate layers and does not implement backpropogation. Nonetheless, some readers may find this description more intuitive, and, as mentioned in the main article, we believe that the high performance of \methodname\ might motivate the design of CNN architectures that share some of these computational elements.

Because deep CNNs are a state-of-the-art tool for SIML tasks, we provide further comparisons of \methodname\ performance and cost relative to this benchmark in Sections~\ref{sec:cnn_benchmark}~and~\ref{sec:cost_analysis}, respectively.

    \subsection{Data separation practices and cross-validation}
         \label{sec:data_separation_uncertainty}
         We split our data into a 20\% holdout test sample and an 80\% training and validation sample. Within the training and validation sample, we perform 5-fold cross validation in our primary analysis, splitting the training and validation sample into 5 sets of 80\% training data (64\% of full sample) and 20\% validation data (16\% of full sample), such that the validation sets are disjoint.

\paragraph{\textbf{Creating the holdout test set}} Before any of the label data are touched, we remove a hold-out test set that is chosen uniformly at random from the entire sample, consisting of $20\%$ of the original data. The analysis and diagnostic procedures that follow use only the remaining $80\%$ of the observations. The held-out test set is only used once, for the purposes of comparison to the validation set performance in Table~\ref{SI:testSetTable}. It is important to keep these data untouched until this point to ensure that our final performance results do not suffer from over-fitting.

\paragraph{\textbf{Tuning hyperparameters}} We choose the optimal $\lambda$ in Eq.~\eqref{eq:regularization} for each outcome through 5-fold cross-validation over the training and validation sample. Specifically, $\lambda$ is chosen to maximize average performance ($R^2$) across 5 folds, from a list of candidate values.\footnote{We choose these candidate values so as to ensure the chosen optimal $\lambda$ is not the minimum or maximum of all $\lambda$s supplied.}
For tasks with the same sampling scheme (i.e. UAR versus population-weighted sampling), the folds are consistent across tasks, so that each of the five folds comprises the same set of locations across the tasks.
 
\paragraph{\textbf{Using cross-validation to measure model robustness}} In addition to being a principled way of selecting hyperparameters, cross-validation gives us a notion of how robust our model is to changes in the training and validation samples. Since each of the 5 validation sets is disjoint and randomly selected, the empirical spread of performance across folds gives us a notion of the variability of our model when applied to new data sets from the same distribution. Understanding this variation is one way of understanding the performance of our model; it gives us a notion of variance of aggregated performance (e.g. $R^2$ over the entire sample, for a given set of hyperparameters).  A useful aspect of \methodname's low computational cost of model training, however, is that it enables researchers to calculate the variance of individual predictions by bootstrapping. %although we leave this type of testing to future work. 
         
     \subsection{Training and testing the model}
     \label{sec:training_testing_model}
         %\paragraph{\textbf{Primary specification}}

%For each outcome, we begin with 100,000 observations of grid cell labels and image features sampled over the continental U.S. We pre-process these data by setting aside a 20\% validation set (see Sec.~\ref{sec:data_separation_uncertainty} for details on how we split our samples) and dropping any missing values. Our training set size thus becomes $N$ = 80,000 for forest cover, 80,000 for elevation, 54,375 for population, 80,000 for nighttime luminosity, 73,102 for income, 80,000 for roads, and 58,729 for housing. We model outcomes whose label distribution is approximately log-normal by taking a log transformation of the labels, adding 1 before logging to avoid dropping labels with an initial value of zero (see Section \ref{sec:log_level} for performance in logs and levels for all tasks). \footnote{Since housing price per square foot is always non-negative, for that variable we use just a log transformation.}

In our primary model (results shown in Fig. \ref{Fig:MapArray} of the main text) we solve for grid cell labels as a linear function of random convolutional features using a ridge regression model and a cross-validation procedure. 
To obtain training and validation sets, we follow the data separation practices outlined in Sec.~\ref{sec:data_separation_uncertainty}, and drop any observations with missing values.
The resulting combined training and validation set sizes are $N$ = 80,000 for forest cover, 80,000 for elevation, 54,375 for population density, 80,000 for nighttime lights, 73,102 for income, 80,000 for road length, and 80,420 for housing price. 

\begin{figure}[!t]
        \centering
        \includegraphics[width=.6\textwidth]{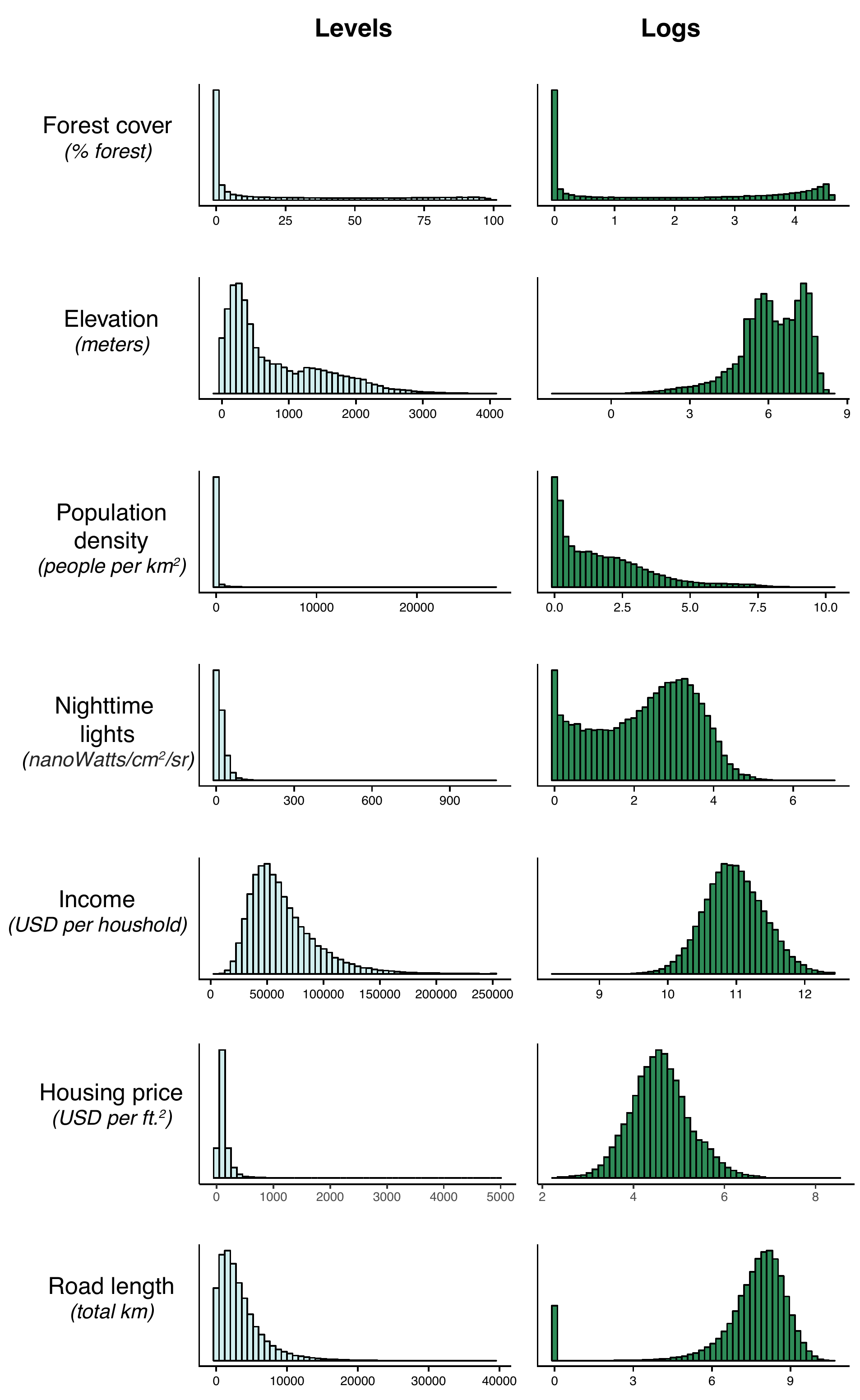}
        \caption{\textbf{Distribution of outcome variables in levels and logs.} Histograms show the distribution of each outcome variable over all sampled image grid cells (approximately 100,000 observations, depending on data availability). Forest cover, elevation, and population density are sampled uniform at random across the continental US, while all other variables are randomly sampled with population weighting. The first column shows the distribution in levels, and the second in logs. For elevation, population density, nighttime lights, and road length, logs were taken after adding 1 to the raw values, given the propensity of zero values in these outcomes.}
        \label{fig:log_level_hists}
\end{figure}

Population density, nighttime lights, and housing price have label distributions that are approximately log-normal (Fig. \ref{fig:log_level_hists}), so we take a log transformation of the labels. We add 1 before logging to avoid dropping labels with an initial value of zero (see Section \ref{sec:log_level} for performance in logs and levels for all tasks).\footnote{Since housing price per square foot is always positive, for that variable we use just a log transformation.}

With these labels and features in hand, we regress each outcome $y_{\ell}^s$ for each task $s$ on features $\Xell$ as follows: 
\begin{align}
    y_{\ell}^s = \XIell \bm\beta^s + \epsilon_\ell^s 
    \label{Eq:GeneralProblemSI}
\end{align}
We solve for $\bm\beta^s$ by minimizing the sum of squared errors plus an $l_2$ regularization term: 
\begin{equation}
\underset{\bm\beta^s}{\min} \frac{1}{2}|| y_\ell^s - \XIell \bm\beta^s ||^2_2 + \frac{\lambda^s}{2}||\bm\beta^s||^2_2 
\label{eq:regularization}
\end{equation} 

We use ridge regression across all outcomes to demonstrate the generalizability of using a single set of image features across many simple regression models. Further, this standardized methodology facilitates comparison of performance and sensitivity across tasks. We note that other modeling choices could potentially improve fit (e.g. using a hurdle model for zero-inflated outcome distributions such as road length); we leave such task-specific explorations for future research. 

In visual display of results and calculation of performance metrics such as $R^2$, we clip our predictions for each task at the minimum and maximum values observed in the labeled data. %Figs. \ref{Fig:MapArray} and \ref{fig:us_error_scatter} show performance and error structure of these predictions, respectively, across the US for all seven tasks.

\begin{figure}[th!]
        \centering
        \includegraphics[width=.8\textwidth]{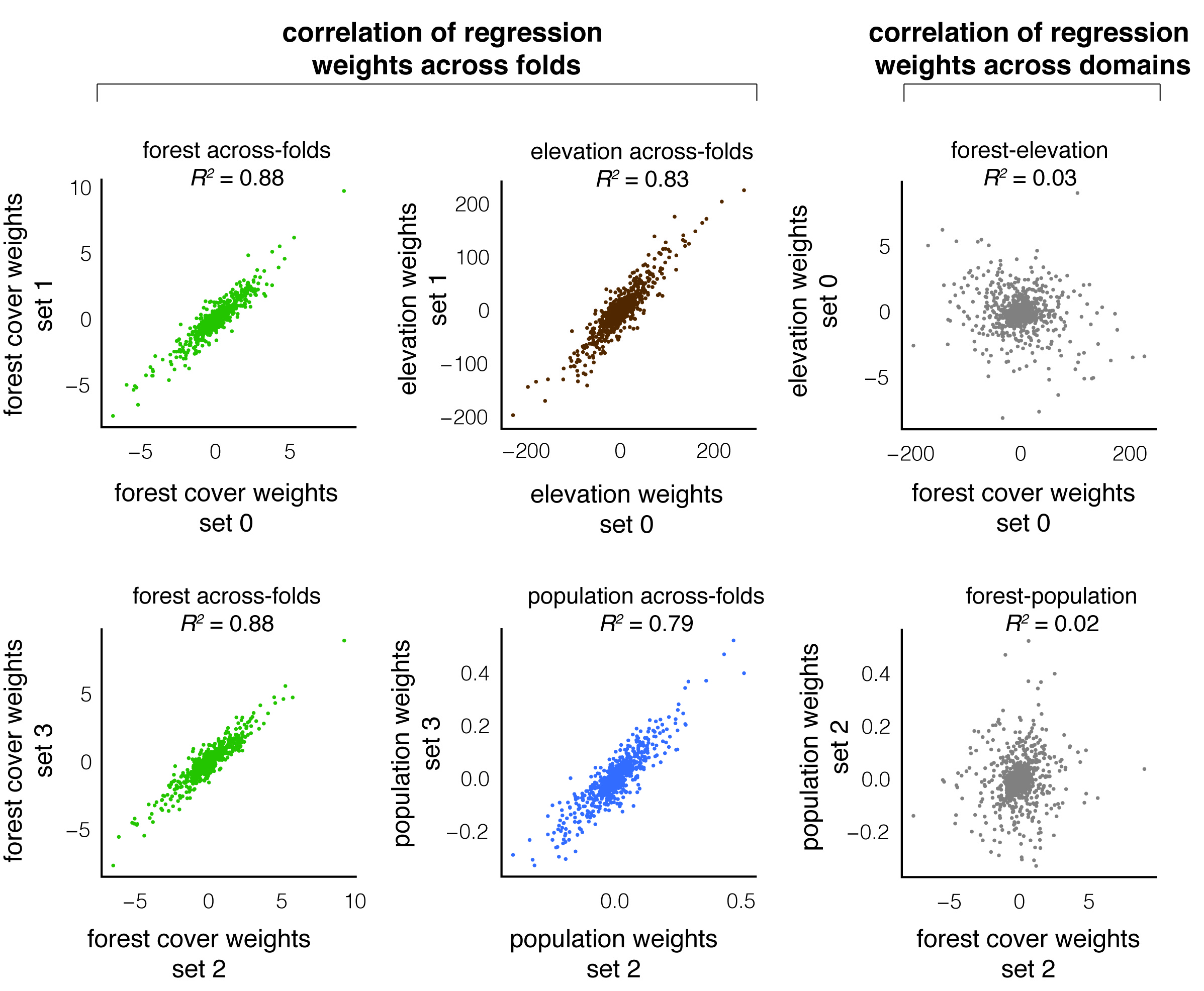}
        \caption{\textbf{Regression weights across folds within a task vs. across tasks within a fold.} All scatterplots indicate regression weights for forest cover, elevation and/or population density. Each point depicts estimated coefficent values for the $k$th feature ($\bm\beta^s_k$) when trained on either different samples or different labels.  In the across-fold examples (first two columns), we learn weights for disjoint training and validation splits for the same task via cross-validation in which one fold acts as the training set and the other as the validation set. Values corresponding to each axis are the regression weights when that fold is the training set (e.g. the top left scatter shows $\{\bm\beta_k^{forest1},\bm\beta_k^{forest0}\}$), and indicate a strong correlation across regression weights from different folds. In the across-task examples (last column), regression weights are shown for the same training and validation sets for two distinct tasks (e.g. the top right scatter shows $\{\bm\beta_k^{elevation0},\bm\beta_k^{forest0}\}$). We see that there is virtually no correlation in regression weights across tasks, demonstrating that predictions across tasks lie in orthogonal subspaces of the feature space. Across all examples here, we set the number of random features to $K = 1,024$.%For consistency across comparisons, $R^2$ is calculated on standardized regression weights, which have been demeaned and divided by their standard deviations.
        }
        \label{fig:correlation_of_betas}
\end{figure}

The resulting weights (i.e. regression coefficients) $\hat{\bm\beta}^s$ obtained from estimation of Eq.~\eqref{Eq:GeneralProblemSI} indicate, along with the variance of the features, which features $k$ (derived from random patch $\Pk$) capture meaningful information for prediction in each task. Fig.~\ref{fig:correlation_of_betas} demonstrates that the recovered weights are stable across cross-validation folds within a task. The first two columns show standardized weights that are estimated from disjoint training and validation splits for the same task.\footnote{For consistency across comparisons, $R^2$ is calculated on standardized regression weights, which have been demeaned and divided by their standard deviations. The number of random features is set to $K = 1,024$ for visual display purposes.} Values corresponding to each axis are the regression weights estimated when the corresponding fold composes the training set. High $R^2$ values indicate a strong correlation between regression weights from \textit{different folds within a single task} (forest cover, elevation, and population density are shown), demonstrating that similar linear combinations of features are selected by the regression model, even when the sample of training images changes. This suggests that specific sets of patches consistently contain valuable information in predicting outcomes for a specific task. However, different combinations of patches are useful for different tasks, and we find no correlation in the weights recovered \textit{between tasks}. For example, in the last column of the figure, we show that regression weights that are recovered for forest cover and elevation (top right) are essentially orthogonal as are regression weights recovered for forest cover and population (lower right). In these two plots, regression weights are shown for the same training and validation sets, but for two distinct tasks. Sets of features that are relevant for prediction in one task appear to be irrelevant for another, as there is virtually no correlation in regression weights. 

\paragraph{Intuition} The consistency of weights recovered in \methodname\ across folds within a task, and the orthogonality of weights recovered within a fold but across tasks, provides some intuition for why \methodname\ provides consistent results and generalizes across a very large (potentially infite) number of potential tasks. The rich featurization $\XIell$ locates image $\Iell$ in a very high-dimensional ($K$-dimensional) feature space.  Solving for $\bm\beta^s$ in Eq.~\eqref{Eq:GeneralProblemSI} then identifies the $K$-dimensional vector $\bm\beta^s$ that points in the direction of most rapid ascent (the gradient vector) for labels $y^s$, when the position of images $\XIell$ are projected onto this vector. Because the feature space is so large --- our baseline model has an 8,192-dimensional feature space --- there are a vast number of orthogonal gradient vectors that can be drawn through this space along which images can be organized for different tasks. The left and center panels of Fig.~\ref{fig:correlation_of_betas} illustrate that similar $K$-dimensional gradient vectors $\bm\beta^s$ are selected when solving for the same task but using different samples (each point depicts an element of the vector $\bm\beta^s$). The right panels shows that for different tasks, the gradient vectors are orthogonal and point in completely unrelated directions in the feature space. This orthogonality means that predictions $\hat{y}^s$ for different tasks will be independent of one another, even though both are constructed as linear combinations of the same set of features.

     \subsection{Primary model test set performance, robustness to functional form, and spatial distribution of errors}
     \label{sec:test_set_performance}
         
Here, we describe how we test for overfitting to the training and validation set in our primary model, test for primary model performance robustness to alternative functional forms, and characterize the spatial distribution of primary model error. 

\paragraph{\textbf{Performance in a holdout test set}}

To test for overfitting, we evaluate the performance of our primary model on a randomly sampled 20\% holdout set. These data were never used for model selection and were only touched at the end of our analysis to check for overfitting. To conduct this test, for each outcome, we use cross-validation within the training set to determine the outcome-specific optimal $\lambda$. We then retrain the model on the full training set using this optimal $\lambda$, and evaluate this model on the holdout test set. We find that performance in the test set is nearly identical to that of the validation set (Table~\ref{SI:testSetTable}), which indicates that our models were not overfit to the data. For some performance metrics, such as the maps in the main text, we present validation set performance (instead of the test set) because the sample is larger and the performance is unchanged.

\begin{table}[htbp] \centering 
\begin{tabular}{@{\extracolsep{5pt}} lcc} 
   & Cross-validation & Test set  \\ 
   \textbf{\textit{Task}} & $R^2$ & $R^2$  \\ 
\hline \\
 Forest cover & $0.91$ & $0.91$ \\ 
 Elevation & $0.68$ & $0.68$ \\ 
 Population density & $0.73$ & $0.72$ \\ 
 Nighttime lights & $0.85$ & $0.85$ \\
 Income & $0.45$ & $0.45$ \\ 
 Road length & $0.52$ & $0.53$ \\ 
 Housing price  & $0.50$ & $0.52$ \\ 
\end{tabular} 
\caption{\textbf{Model performance in the hold out test set.} For each outcome, we use 5-fold cross-validation within the training/validation set using 80\% of our labeled data to optimally select task-specific hyperparameters in ridge regression (i.e. $\lambda$). We then retrain each model on the full training set using this optimal $\lambda$. Performance on the validation set (column 1) is compared to that of the held out test set (column 2).}
\label{SI:testSetTable}
\end{table}

\paragraph{\textbf{Robustness of model to alternative functional forms}}
\label{sec:log_level}

Throughout the main text, we report primary model performance in each task from a model estimated with labels that are either logged (e.g. population density), or in levels (e.g. forest cover). The decision regarding functional form for each task was made based on the underlying distribution of labels across our image grid cells. Many outcomes, such as housing prices, display exceptionally skewed distributions that approximate log-normality (see Fig.~\ref{fig:log_level_hists}). For these outcomes, we take the natural log of the image grid cell values in model training and testing. Table~\ref{tab:logvlevel} shows model performance for all tasks under both the levels and logs functional forms.\footnote{In tasks where negative values or zeros are present (e.g. forest cover, elevation, and nighttime lights), we drop negative values and add one to zero values before taking logs for this test.} Tasks with highly skewed distributions, such as population density, housing price per square foot, and nighttime lights have substantially higher performance ($R^2$ increases by 10-64\%) after being logged. Tasks whose labels display much less skew in levels, such as road length, income, and elevation show small to modestly reduced performance (4-21\%) when their outcomes are modeled in logs. %\todo{CHECK WHETHER THIS IS TRUE ABOUT THE UNDERLYING DISTRIBUTIONS! If so, calculate the actual \%s or say what the take away is for this section. I think its that things perform better when their outcomes are normally distributed but that may not be the case -- need to see histograms.}

\begin{table}[ht]
\iftoggle{arxiv}{}{\vspace{-1cm}}
\centering
\begin{tabular}{lcc}
& Log model  & Levels model \\
 \textbf{\textit{Task}} & $R^2$  & $R^2$ \\
 \hline
&& \\
Forest cover & 0.90 & \textbf{0.91} \\
Elevation & 0.58 & \textbf{0.68} \\
Population density &\textbf{0.73}& 0.56 \\
Nighttime lights & \textbf{0.85}& 0.77 \\
Income & 0.43 & \textbf{0.45} \\
Road length &0.41& \textbf{0.52} \\
Housing price  & \textbf{0.50}& 0.44 \\
&&
\end{tabular}
  \caption{ \textbf{Model performance across tasks and functional forms.} All $R^2$ values indicate performance using the optimal hyperparameter $\lambda$ after 5-fold cross-validation. In the log model, the outcome variable is defined as the natural logarithm of the original labeled data (e.g. natural log of the average forest cover over an image gridcell). In the levels model, the outcome variable is simply the level of the aggregated labeled data, as defined in Section \ref{sec:label_agg}. Values in bold are reported in the main text.}
\label{tab:logvlevel}
\end{table}

% \begin{figure}[th!]
 \begin{figure}[th]
        \centering
        \includegraphics[height=.85\textheight]{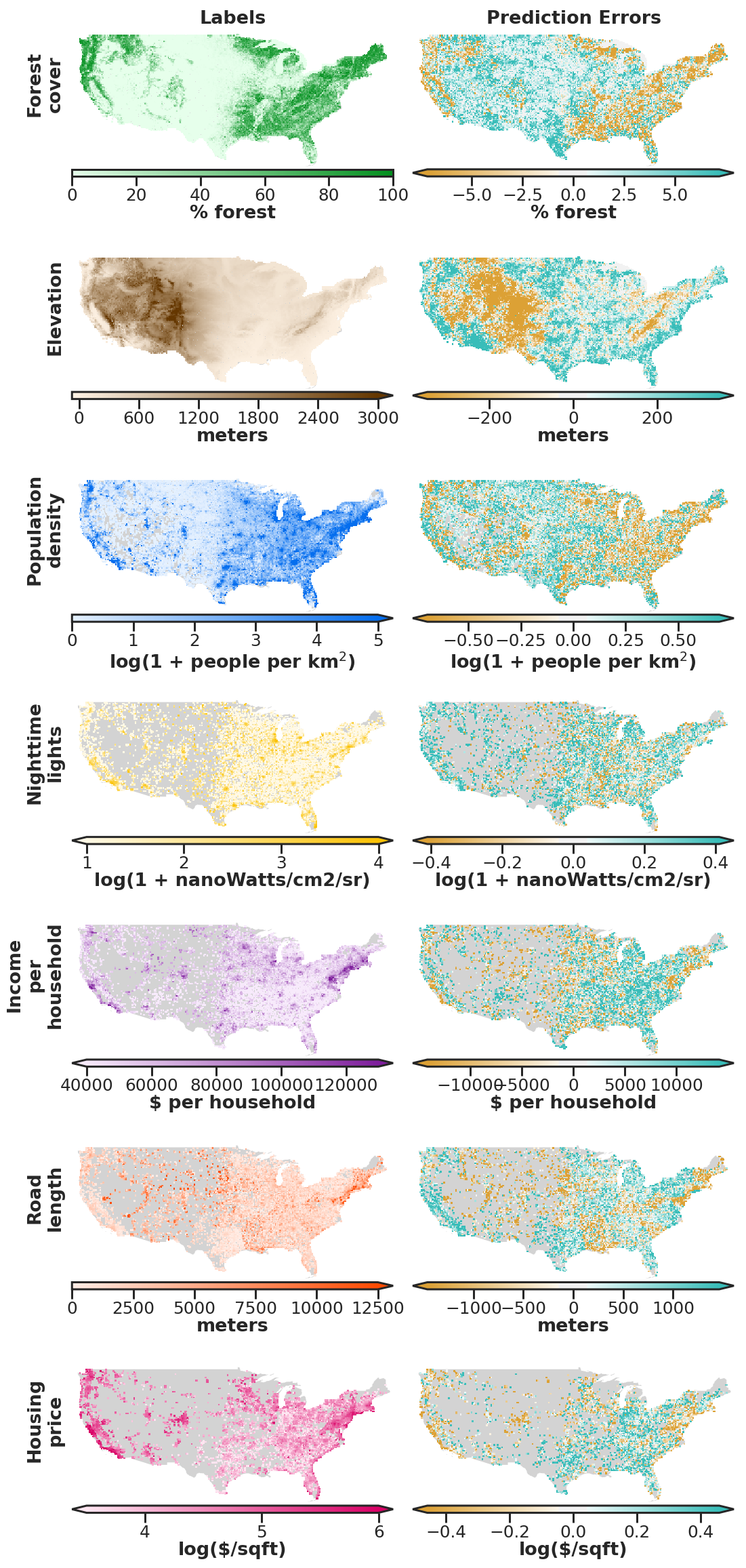}
        \caption{\textbf{Labels and prediction errors over space for each task.} Left maps: $\sim$80,000 observations used for training and validation, aggregated up to 20km x 20km cells for display (precise number of observations varies by task based on data availability; see Section \ref{sec:training_testing_model}). Right maps: prediction errors from concatenated validation set estimates from 5-fold cross-validation for the same $\sim$80,000 grid cells, identically aggregated for display.}
        %Prediction error color bars for each task span 1 standard deviation away from zero. The first three tasks are uniformly sampled across space, while the bottom four tasks are sampled using population weights. Grey areas are not sampled in experiment.}
        \label{fig:us_error_scatter}
\end{figure}

\paragraph{\textbf{Spatial distribution of errors}}

Fig.~\ref{fig:us_error_scatter} shows the distribution of errors over space, for the model predictions presented in Fig.~\ref{Fig:MapArray}. 
The model systematically over-predicts low values and under-predicts high values across all tasks. This is likely due to our choice of ridge regression, which favors predictions that tend toward the mean due to the $\ell_2$ penalty. 
The structured correlation of errors across space suggests that there is substantial room for model improvement, potentially from including task specific knowledge. For example, our models of housing price and elevation could likely, respectively, be improved by adding in information about school districts --to address clustering of house price error in parts of big cities -- or location -- to help identify large areas of high elevation such as the Rocky Mountains. We recognize that there exists substantial room for task-specific model performance, which we leave for future research. 
Further, discontinuities in the error structure over political boundaries can help identify inconsistency in label quality. For example, the sharp increase in road length prediction error moving across the border from Louisiana to Texas suggests that the raw data labeling in these two states may differ methodologically, which introduces error into the label, and in turn, the model.

\newpage\clearpage

     \subsection{Altering the number of features and training set size}
     \label{sec:secondary_analysis_nk}
         To better understand factors that could improve the primary model, we test the sensitivity of its performance to the number of features and the training set size (results shown in Fig.~\ref{Fig:Waffle} in the main text). Understanding the returns to additional features and observations enables better optimization of model performance given cost constraints.

Since features in \methodname\ are generated randomly, there is no theoretical reason to select a specific number of features. To test the sensitivity of the primary model performance to the number of features, we train a model identically to our primary specification (Section \ref{sec:training_testing_model}) except that we vary the number of features across the values $\{100, 200, 500, 1000, 2000, 4096, 8192\}$ (Fig.~\ref{Fig:Waffle}A). For each set of features and each task, we conduct 5-fold cross-validation to recover the optimal hyperparameter $\lambda$.

%We find that increasing the number of features increases performance with diminishing marginal returns. 

%-- but comes at the cost of additional computation. Featurization costs scale linearly with the number of features, as discussed in more detail in Section~\ref{sec:cost_analysis}. (min 81\% for income, max 96% for nighttime luminosity).

Notably, using only 100 features recovers a substantial amount of the variation across tasks. Of the tasks, the least variation is recovered for income ($R^2$ using 100 features is 81\% of $R^2$ using 8,192 features) and the most variation is retained in nighttime lights ($R^2$ using 100 features is 96\% of $R^2$ using 8,192 features). This suggests that in computation or memory-limited settings, fewer features could be used with only minor losses in performance. On the other hand, even with 8,192 features, performance does not fully flatten out (on a logarithmic scale). This suggests that performance could be improved further by increasing the number of features past $K = 8,192$. At the limit of our testing, a doubling of $K$ from $4,096$ to $8,192$ led to a largest performance increase of $0.026\ R^2$ for income and a smallest of $0.010\ R^2$ for forest cover.

To test the sensitivity of primary model performance to the number of training samples, we train a model identical to our primary specification (with 8,192 features) except with a varying size of training set (from 500 to 64,000 images) (Fig.~\ref{Fig:Waffle}A).\footnote{The same per-fold validation sets are used for each iteration of this analysis as well as for the primary analysis and for the test of model performance sensitivity to the number of features.} In cases where the training set has fewer than 64,000 total observations due to missing data (e.g. population density, income, road length and housing price), we use the full training data set to construct our largest training sample.

Similarly to increasing the number of features, increasing the training set size increases model performance with diminishing marginal returns. Notably, models trained on only 500 observations recover at minimum 56\% (road length) of performance relative to $N = 64,000$ and at maximum 87\% (forest cover), excluding income and housing price, which require larger samples to attain performance. This suggests that, for all but the most difficult SIML tasks, \methodname\ may be useful even when label collection is very costly. For the tasks with the best $R^2$ performance (forest cover, nighttime lights), performance plateaus out as the number of training observations approaches $64,000$. However, for the remaining five tasks, these results show that more training data could substantially increase performance further. The range of performance gain from increasing $N = 32,000$ to $64,000$ is bounded below by forest cover ($.005\ R^2$) and above by road length ($.027\ R^2$).

     \subsection{Testing generalizability across space and comparison to kernel-based interpolation.} 
     \label{sec:secondary_analysis_spatial}
         \begin{figure}[t]
        \centering
        \includegraphics[width=0.6\textwidth]{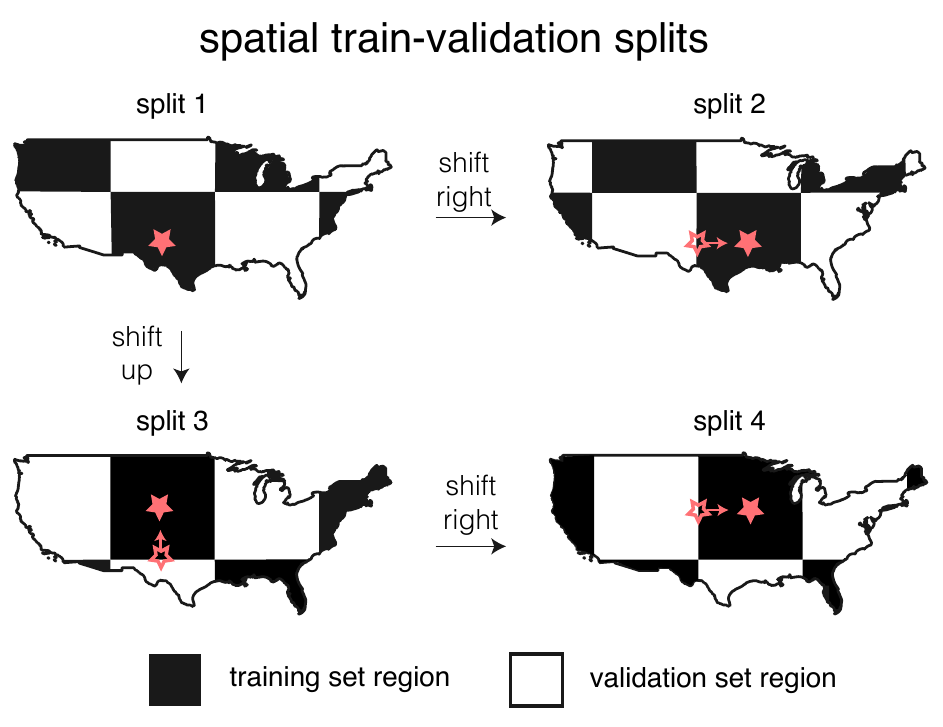}
        \caption{\textbf{Illustration of the procedure to systematically shift train and validation sets in space when assessing the performance of \methodname over regions with no ground-truth data.} To assess the ability of \methodname\ to generate meaningful predictions when extrapolating across large spatial distances, we conduct a ``checkerboard'' experiment (Section \ref{sec:secondary_analysis_spatial}, Fig. 3B-C of the main text) in which the training set (``black squares'') and validation set (``white squares'') are separated by increasingly large distances. The length of a square in each experiment is $\delta$, measured in degrees. This figure demonstrates the four different train/validation splits that are created by shifting a given spatial checkerboard (split 1) by $\delta/2$ to the right (split 2), $\delta/2$ up (split 3), and both  simultaneously (split 4).}
        \label{fig:jitter_explanation}
\end{figure}

To understand the ability of our model to predict outcomes in large contiguous regions with no ground truth, we design an experiment where we evaluate models using training and validation sets that are increasingly far away from each other in space. Specifically, we iteratively create a grid over the US with a side length of $\delta$ degrees and then use this grid to divide the training and validation dataset ($N = 80,000$) into spatially disjoint sets of roughly equal size. We create these disjoint sets by assigning observations that lie in every other box within the grid to the train set and test set, respectively, creating a checkerboard pattern with the train set and test set, as shown in Fig.~\ref{Fig:Waffle}B. We vary the width $\delta$ of each square in the grid range across the values of $\{0.5, 1.5, 2, 4, 6, 8, 10, 12, 14, 16\}$ degrees (roughly 40 to 1400 km) in sequential runs of the experiment. As $\delta$ increases, validation set observations become on average farther away from the training set points. This distance makes prediction on the validation set more difficult, because observations in the validation set are now likely to be less similar to those in the training set. We learn the model on the training set using ridge regression. To assess the stability of this performance, we offset the checkerboard and re-run the above analysis four times -- once in the original location and then three more times -- shifting the grid up, right, and both up and right by half the width of the grid (see Fig.~\ref{fig:jitter_explanation}). The $\ell_2$ regularization term, $\lambda$, is selected to maximize average performance in the four validation sets, as we would select in in a standard cross-validation procedure.

The performance plotted in Fig.~\ref{Fig:Waffle}C is the performance on the the resulting validation sets. We find that across most tasks, performance degrades only slightly as the distance between training observations and testing observations increases. This suggests that \methodname\ is indeed learning image-label mappings that transfer across spatial regions.

\paragraph{\textbf{Comparison of \methodname\ to kernel-based spatial interpolation}}
%motivation

In these experiments we demonstrate that \methodname\ outperforms spatial interpolation (or extrapolation, depending on geometry) -- a commonly used simple technique to fill in missing data (Fig.~\ref{Fig:Waffle}C). This suggests that \methodname, and SIML generally, exploits the spectral and structural content of information within an image to generate predictions at national scale that extend beyond what can be captured by geographic location alone. 

%valuable measurements relative to spatial interpolation at a national scale, and that \methodname\ is basing predictions on the spectral and textural content of information of the image, rather than simply learning to predict outcomes by identifying where they are in space.

%This suggests that \methodname, and SIML generally, can provide valuable measurements relative to spatial interpolation at a national scale, and that \methodname\ is basing predictions on the spectral and textural content of information of the image, rather than simply learning to predict outcomes by identifying where they are in space. 

%We compare the performance of our satellite model to a baseline model using Gaussian interpolation in this spatial extrapolation experiment to better understand the relative value of satellite imagery in measuring outcomes at a national scale. There are two dimensions to this. First, \methodname, and SIML generally, is useful only if it outperforms existing simpler techniques to measure and fill in missing data, such as spatial extrapolation \todomisc{[CITE]}. Second, it could be the case that \methodname\ is not learning to measure outcomes, but rather learning to predict outcomes by learning where images are in space. If \methodname\ outperforms interpolation than it is likely capturing information other than the location of the image. 

We compare \methodname to kernel-based spatial interpolation using a Gaussian Radial Basis Function (RBF) kernel, a simple and general widely used approach.  In this approach, the value for a point in the validation set at location $\ell_v \in \mathbb{R}^2$ is predicted to be a weighted sum of the values of all the points in the training set $\ell_t$, as follows:

\begin{align*}
\hat{y}_v^s = \frac{\underset{\ell_t \in [\textrm{Train}]}{\sum} y_t^s w(\ell_t, \ell_v)}{\underset{\ell_t \in [\textrm{Train}]}{\sum} w(\ell_t, \ell_v)};
\hspace{4em} w(\ell_t, \ell_v) = e^{-\frac{1}{2\sigma^2}\|\ell_t - \ell_v\|^2}
\end{align*}
Here, $w$ is the weight assigned to each observation in the training set based on kernel values that are indexed to distance, such that $w$ decreases as the distance between the point being predicted and the point in the training set increases. We select $\sigma$ -- the parameter that determines the rate at which $w$ degrades with distance -- to maximize average performance on the validation set across all four spatially-offset runs, similar to how we tune $\lambda$ in the spatial extrapolation experiment described above. The optimal value of the bandwidth parameter $\sigma$ will depend on the task at hand, as well as the average distance from points in the validation set to points in the training set. To ensure comparability, spatial interpolation based predictions and performance are computed for the exact same samples as used for \methodname\ in each checkerboard partition.

%results
%We find that in all tasks except elevation, \methodname\ outperforms spatial interpolation. This shows that \methodname\ is able to measure outcomes better than interpolation -- and thus is of practical value -- and that \methodname\ is capturing information about an image other than its location. In the case of elevation, spatial interpolation outperforms \methodname. This makes sense because elevation is both smoothly varying and continuous across space. Even for elevation, however, the perturbation analysis ranges are overlapping for $\delta \geq 12$.    

%Looking at the performance of spatial interpolation and random features across space also tells us much about the nature of the seven tasks of study, and their amenability to being remotely sensed. \todomisc{expand - im thinking like forestcover, nl, and income are all looking pretty good across space, NL in particular is also way better than interpolation. }
        
     %\subsection{Image-2-Vec Properties}
      %   \inputt{img2vec.tex}

     \subsection{{Label super-resolution}}
     \label{sec:superresolution}

 \begin{figure}[h]
        \centering
        \includegraphics[width=.85\textwidth]{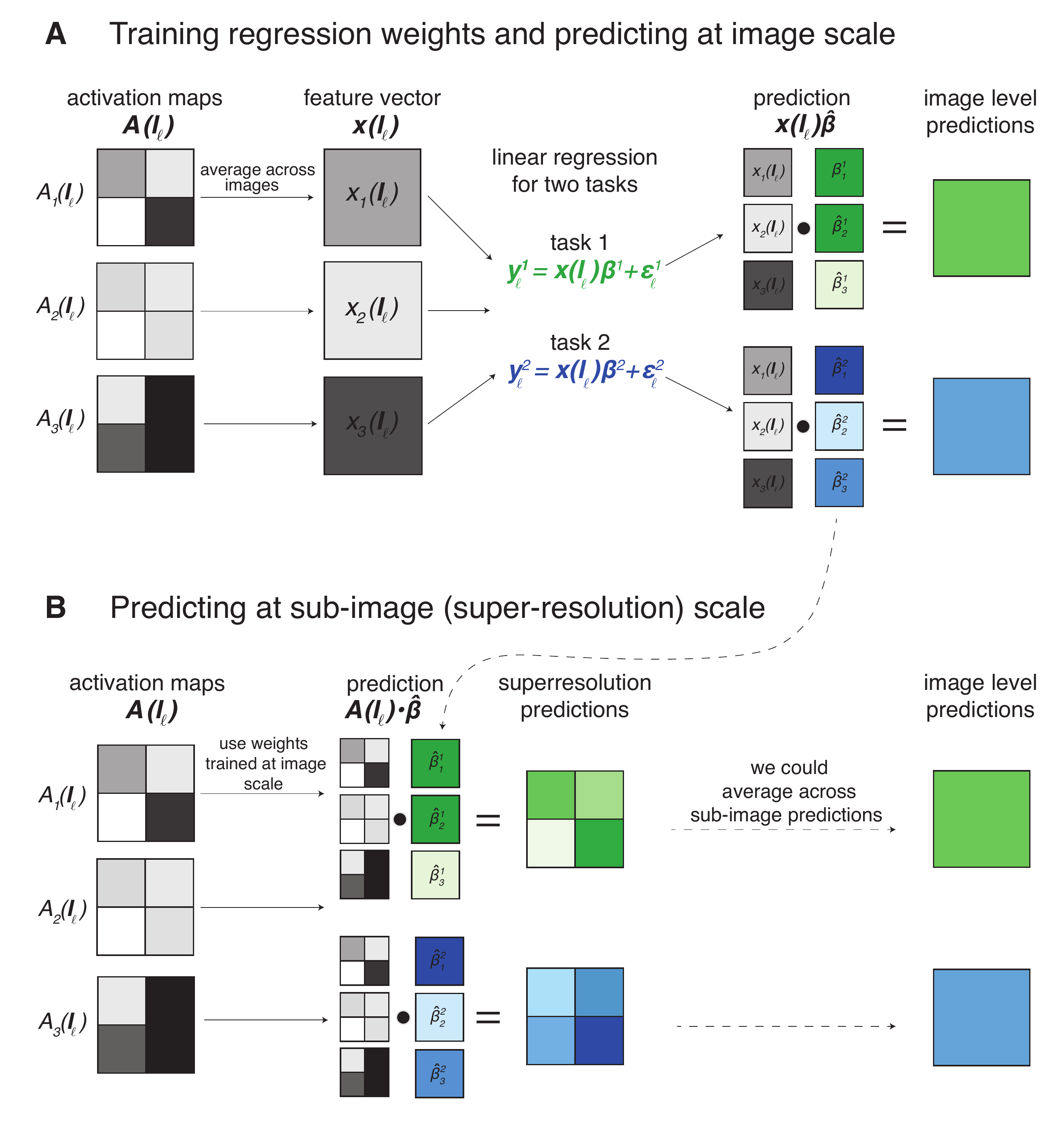}
        \caption{\textbf{Illustration of the procedure to construct predictions at image resolution and label super-resolution.} Panel A illustrates the standard \methodname\ prediction pipeline. After convolution with random patches, nonlinear activation maps $\AkIell$ are averaged across images to construct a set of image-level features $\XkIell$ used in linear regression to generate predictions at image-scale (Section \ref{sec:featurization}). Panel B illustrates how the weights trained using labels and features at image-scale in panel A can be used to generate predictions at resolutions higher than the images and labeled data, achieving predictions at label super-resolution. The scalar product of the entire activation map $\AkIell$ and the estimated weights vector $\hat{\bm\beta}$ generates label super-resolution predictions at any desired sub-image scale larger than pixel-level. The last column of panel B illustrates the fact that label super-resolution predictions, when averaged across an image, are identical to predictions generated from the standard process in panel A.}
        \label{fig:superresolution_cartoon}
\end{figure}
%\clearpage
%\newpage

 \begin{sidewaysfigure}[h]
        \centering
        \includegraphics[width=\textwidth]{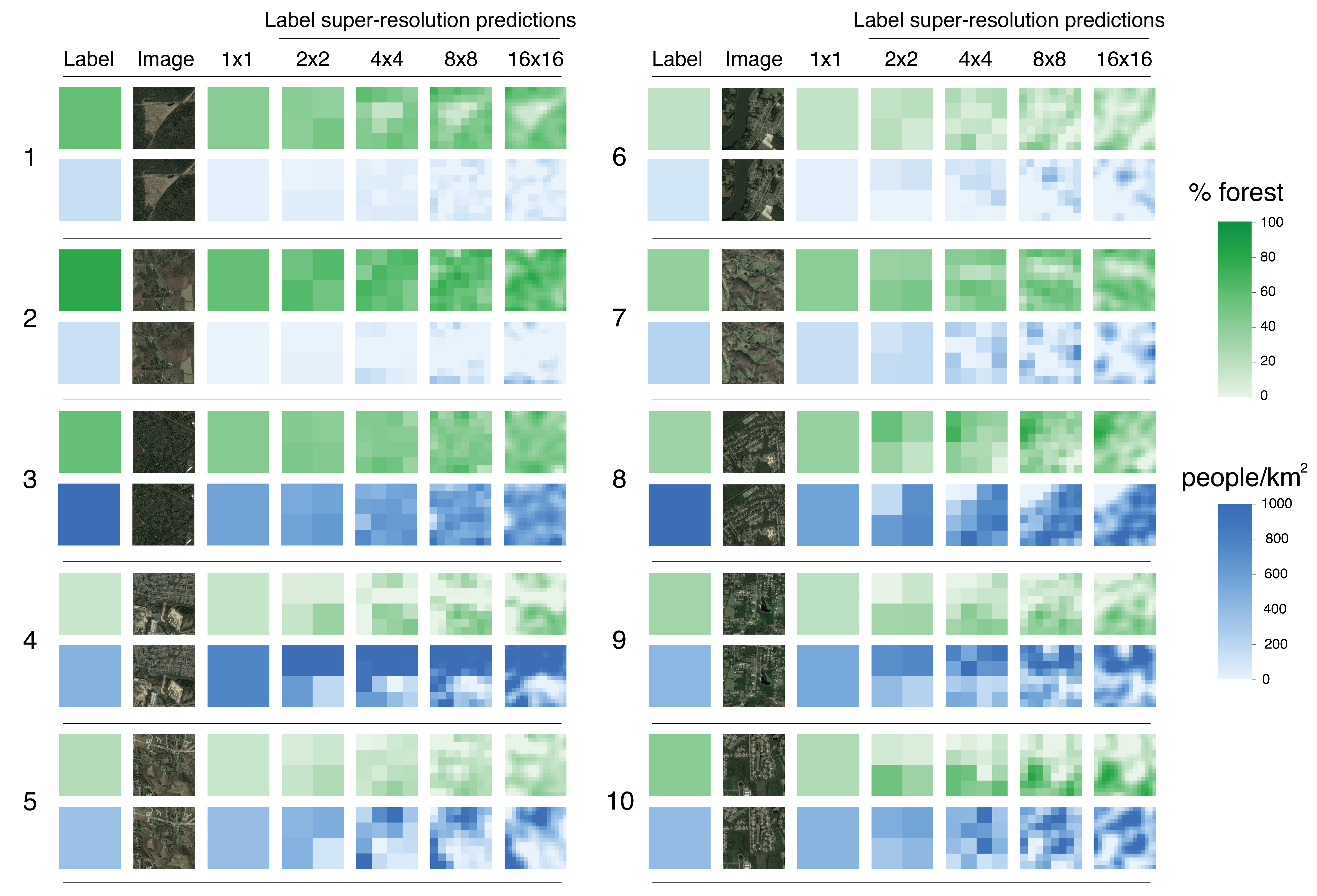}
        \caption{\textbf{Label super-resolution performance across ten randomly selected images.} Each set of images indicate the image-level labels (column 1), the image itself (column 2) and predicted outcomes from \methodname\ at increasing levels of label super-resolution (columns 3-7). These ten examples were selected uniformly at random from images in which our labels indicated at least $10\%$ total forest cover and at least $100$ people/km$^2$.}
        \label{fig:superresolution_all10}
\end{sidewaysfigure}

As discussed in the methods summary, the featurization method in \methodname\ exploits the fact that many image-level outcomes of interest are linearly decomposable across sub-image regions. This is done by 
%extracting structured information at all sub-image regions and 
creating image-level features that are averages of statistics from all sub-image regions. Because these features are ultimately used in linear regression, a natural property of this approach is that weights estimated in this linear regression can be used not only to generate predictions of outcome variables at the image-scale, but also at the scale of any sub-image region. As satellite imagery are available at increasingly high spatial resolution, this ``label super-resolution'' property is both practical and powerful, enabling researchers to generate novel predictions at higher resolution than available ground truth data.

This section gives mathematical justification for a simple method to use \methodname\ to predict outcomes of interest at a finer resolution than available labeled data. 
%We also explain the label super-resolution properties of \methodname\ visually (Extended Data Fig.~\ref{fig:superresolution_cartoon}), and document the empirical performance of label super-resolution. 
We display the label super-resolution properties of \methodname\ visually, and quantitatively document the empirical performance of this label super-resolution approach.

\textbf{Why MOSAIKS naturally achieves super-resolution for label predictions} Given an image-label pair $\{\Iell, y_\ell^s\}$, the goal of label super-resolution is to resolve which sub-regions of the image $\Iell$ contribute to high or low values of $y_\ell^s$. Recall that for image $\Iell$, feature vector $\XIell$ is a $K$ dimensional vector, where each scalar element $\XkIell$ of  $\XIell$ is an average across the pixels of the image of the values obtained by convolving sub-regions of the image with patch $\Pk$. As in Section \ref{sec:featurization}, denote by $\X$ the full random feature matrix in $\RR^{N \times K}$, so that $\X_{\ell k}$ denotes the $k^{th}$ element of the feature vector describing image $\Iell$. By Eq.~\eqref{eq:feat_vector}, we can decompose the feature elements as: 
\begin{align*}
\X_{\ell k} := \XkIell &= \frac{1}{254^2}\sum_{i=1}^{254}\sum_{j=1}^{254} \Ak(\Iell)[i,j]  
\end{align*}
 %\todoesther{Is there a reason we have lower and upper case X's here? - yeah so we define the features in terms of vectors $x_k$ in the featurization section so it matches there. Honestly it would make more sense if superscript parenthesies the ks on little x and big A to avoid overloading but ppl thought that was ugly.}
where $\Ak$ is the activation map associated with patch $\Pk$. Since we are using a linear model to form predicted values, we can trace these values back to subregions of the original image.
When we perform a linear regression for task $s$, the resulting regression weights are a vector $\hat{\bm\beta}^s \in \RR^K$ such that the scalar $\hat{\beta}^s_k$ describes the relative weight of feature $k$ in the image-scale predictions. The prediction of outcome $s$ using image $\Iell$ thus decomposes as:
\begin{align*}
    \hat{y}_\ell^s &= \X_\ell \hat{\bm\beta}^s \\
    &= \sum_{k=1}^{K} \X_{\ell k} \cdot \hat{\beta}^s_k \\
    & =  \sum_{k=1}^{K} \left(\frac{1}{254^2} \sum_{i=1}^{254} \sum_{j=1}^{254} \Ak(\Iell)[i,j]\right) \cdot \hat{\beta}^s_k \\
    &= \frac{1}{254^2} \sum_{i=1}^{254} \sum_{j=1}^{254}
    \underbrace{\left(  \sum_{k=1}^{K}  \hat{\beta}^s_k \cdot \left(\Ak(\Iell)[i,j] \right) \right )}_{\text{super-resolution prediction}}
\end{align*}
where the third line follows from substituting $\X_{\ell k}$ according to Eq.~\eqref{eq:feat_vector}.
Therefore, we can associate with each pixel indexed by $(i,j)$ a predicted super-resolution value:
\begin{align}  
    \label{eq:per_pixel_contribution}
    \hat y_{\ell,(i,j)}^s = \sum_{k=1}^{K}  \hat{\beta}^s_k \cdot \left(\Ak(\Iell)[i,j] \right)
\end{align} 
which is that pixel's predicted label value, and thus its contribution to the overall predicted image-level label value $\hat{y}_\ell$ for $\Iell$. We use a Gaussian filter to smooth these per-pixel predictions to enforce spatial consistency and reduce variance of the high-resolution predictions, using a kernel bandwidth of $\sigma=16$ pixels. These smoothed pixel-level predictions can be average-pooled to larger sub-image scales as shown in Fig. \ref{Fig:Global}B. The procedure to construct label super-resolution predictions, and a comparison of it to the precedure to construct image-level predictions, is illustrated in Fig.~\ref{fig:superresolution_cartoon}. 
     
%For different outcomes $s$, we get different weight vectors $\hat{\bm\beta}^s$, so that we can perform this pixel-wise component analysis for each task, so long as the sub-image-scale features (i.e. activation maps) are saved prior to averaging them to the grid cell level. 

Fig.~\ref{fig:superresolution_all10} demonstrates empirical performance of Eq.~\eqref{eq:per_pixel_contribution} using ten examples of this approach at label super-resolutions on both the forest cover and population density outcomes. The ten images were randomly selected from the union of observations with forest cover $> 10\%$ and population density $ > \textrm{100 people/km}^2$ to ensure that all images considered had a non-negligible value for each variable.\footnote{To ensure that weights decomposed as a sum, as in Eq.~\eqref{eq:per_pixel_contribution}, we used level values (i.e. not log-transformed) for population density labels in  Fig.~\ref{fig:superresolution_all10}.}

In our formulation, super-resolution label predictions are easily estimable during featurization. Consider again the per-pixel contributions of  Eq.~\eqref{eq:per_pixel_contribution}. An alternative way to express this is
\begin{align*}
   \hat y_{\ell,(i,j)}^s = \left( \sum_{k=1}^{K}  \hat{\beta}^s_k \cdot\Ak(\Iell)\right)[i,j] 
\end{align*}
That is, label super-resolution estimates are just a linear combination of the activation maps $\AkIell$ weighted by $\hat{\beta}^s_k$ (see Fig.~\ref{fig:superresolution_cartoon}). Every time we featurize a new image $\Iell'$, we must perform the step of computing the $K$ activation maps $\{ \Ak(\Iell')\}_{k=1}^K$ (Fig.~\ref{fig:featurization}~D). 
Therefore, if we already have a suitable regression weight vector $\hat{\bm\beta}^s$ for task $s$, for any new images $\Iell'$ that we featurize, we can compute the super-resolution label predictions $\sum_{k=1}^{K}  \hat{\beta}^s_k \cdot\Ak(\Iell)$ as weighted combinations of the activation maps at negligible additional cost, prior to pooling, in the existing featurization pipeline. %\todomisc{link to new figure for superres} Alternatively, label super-resolution may simply require a second pass over the image data.

\newpage\clearpage

\begin{figure}[!h]
    \centering
    \includegraphics[width=.4\textwidth]{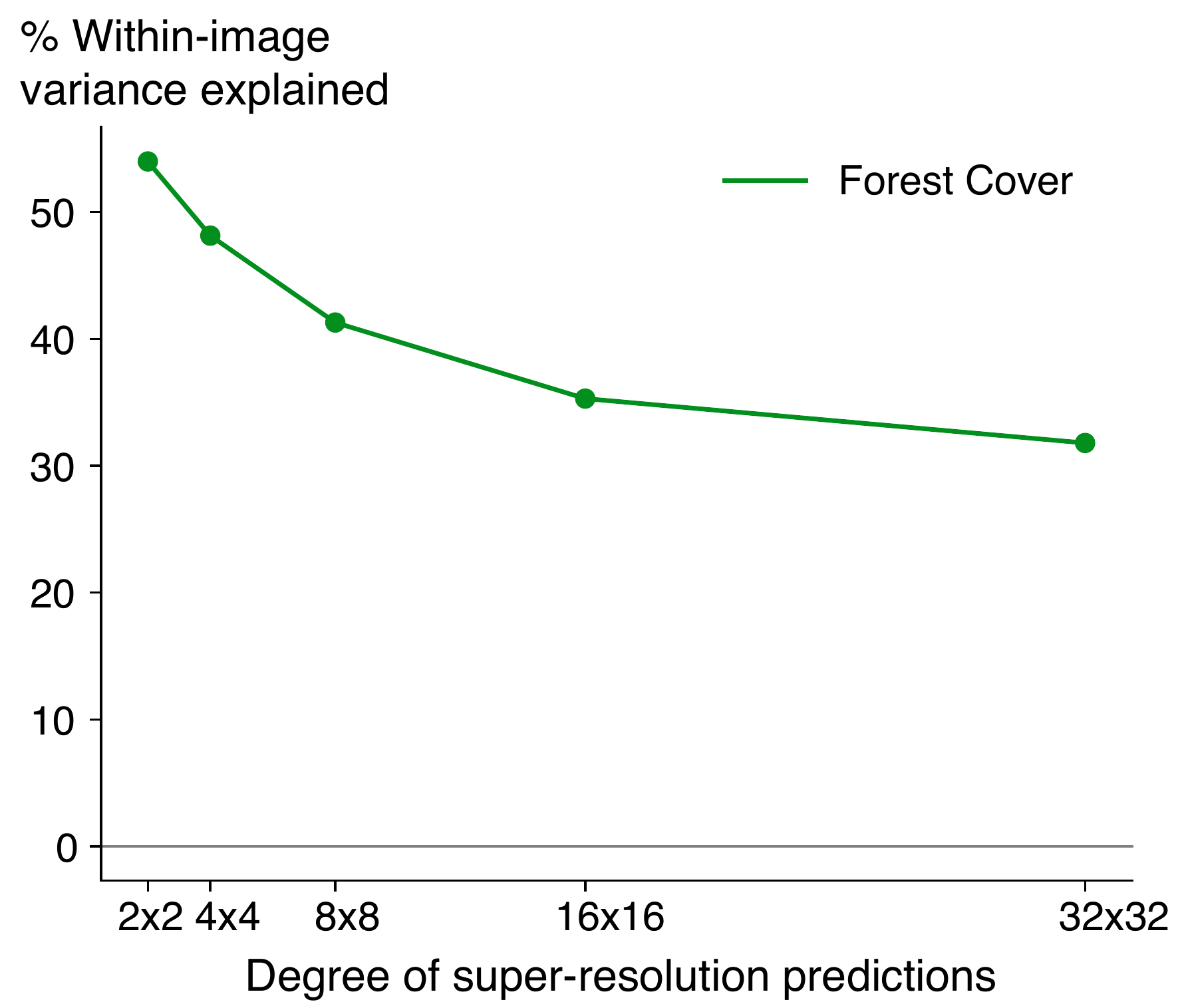}
    \caption{\textbf{Systematic evaluation of \textit{within-image} $R^2$  recovered in the forest cover task.}}
    \label{Fig:superreslineplot}
\end{figure}

\textbf{Evaluating label super-resolution performance}

To systematically evaluate the ability of \methodname\ to accurately predict outcome labels at super-resolution, we evaluate the within-image label variation that \methodname's label super-resolution predictions accurately explain. We use forest cover for this test because the raw label resolution is substantially finer than the grid cell used to construct labels (see Section \ref{sec:grid_and_sampling} and Fig.~\ref{fig:aggregationOfLabels}), so we are able to attach ``true'' labels to super-resolution predictions within each image. In our main analysis, we construct grid cell forest cover labels by averaging fine-resolution raw forest cover data (see Section \ref{sec:label_agg}). Here we leverage the fine resolution of the raw data to compare label super-resolution performance of a model trained on aggregated labels but tested on high-resolution raw forest cover data.

Specifically, we learn regression weights $\hat{\bm{\beta}}^s$ using a ridge regression applied to image-level labels from the full U.S. UAR sample ($N=100,000$). We do so for multiple regularization parameters, $\lambda$, then make label super-resolution predictions on a ``validation'' set of 1,000 images\footnote{Note that none of the pixel-level values in this validation set are used in the ridge regression, but the corresponding image-level labels are in-sample.}. We use the R2 score for 32x downscaled predictions (32x32 predictions per image) to choose optimal values for $\lambda$ and $\sigma$ (the Gaussian filter length scale).\footnote{The optimal $\lambda=1e5$ is higher than that chosen to optimize image-level predictions (Fig.~\ref{Fig:MapArray}), likely due to increased noise in sub-image predictions.} Next, we use the weights derived from our image-level ridge regression, along with the corresponding optimal $\lambda$ and $\sigma$, to make label super-resolution predictions for 16,000 additional images drawn randomly from the full set of 100,000 (excluding the 1,000 used for choosing hyperparameters). Lastly, we aggregate these pixel-level predictions to coarser sub-image scales, where increasing aggregation (lower label super-resolution factor) reduces noise in the predictions at the cost of lower resolution.

We assess the performance of label super-resolution at a variety of scales by calculating the percent of the variance of the raw within-image forest cover labels that can be explained by the super-resolution label predictions at each scale. For example, to assess the performance of $2\times 2$ label super-resolution predictions, we average predictions from the $254 \times 254$ label super-resolution predictions by quadrants, resulting in four predicted values (twice the original resolution).\footnote{For the analysis, we clip the images and predictions to 224 x 224 pixels so they are evenly divisible by a 32x super-resolution factor.} We perform the same per-quadrant average for the raw fine-resolution forest cover labels. We demean both the within-image predictions and labels to eliminate across-image variation, thereby focusing this test on the ability of the predictions to explain residual within-image variation. We then concatenate these within-image predictions and labels across the $N = 16,000$ images, so that the resulting $R^2$ value reported is the percent of super-resolution label variance explained by label super-resolution predictions, across $64,000 = 16,000 \cdot 2^2$ label-prediction pairs. 

The resulting performance of label super-resolution predictions at different scales is shown in Fig.~\ref{Fig:superreslineplot} for width scales of $2 \times 2$, $4 \times 4$, $8 \times 8$, $16 \times 16$, and $32 \times 32$. We test up to $w=32$ because the native width of the forest cover labels ($\sim~$30m) is just under $1/32$ the width of the original image ($\sim~$1km). Label super-resolution predictions are trained only on the aggregate label at the image-level. Nonetheless, as Fig.~\ref{Fig:superreslineplot} shows, we are able to explain over $50\%$ of the within-image label variations at $2 \times 2$ super-resolution, and over $30\%$ of the variation using $32 \times 32$ super-resolution grids.

%The native resolution of the forest cover labels is about $8$ pixels per side (about $1/32 \times$ width of the image). %so due to alignment reasons the lowest super-resolution prediction scale we test is $8$ pixels ($1/32$ width of the original image). 

\textbf{Comparisons to other within-image prediction algorithms} 
The derivation leading to Eq.~\eqref{eq:per_pixel_contribution} has a very similar form to the derivation of class activation mapping in \cite{zhou2016learning}. Similar to our goal of label super-resolution, class activation mapping identifies image sub-regions that contribute to the overall prediction for that image. Class activation mapping usually refers to finding discriminative regions of an image that help explain a binary classification decision; we differ from this in our objective of predicting regression values at finer-resolution than the image-sized labels. We use the term ``label super-resolution'' (also used in~\cite{malkin2018label}) to further distinguish our approach from \emph{image} super-resolution methods in image processing and microscopy, which increase the resolution of the image itself, rather than the associated labels. 

A approach to \methodname's label super-resolution predictions are methods specifically designed for pixel-level classification, or \emph{semantic labelling} of satellite imagery \cite{Firat2014,Volpi2017}. 
However, these approaches make use of sub-image labels for training, as opposed to our setting, where only one label per image (per task) is provided. For example, \cite{malkin2018label} studies the case of weakly supervised image segmentation, predicting land cover at finer resolution than the provided labels, which are already at sub-image resolution.
Some such semantic labelling approaches use a downsample-then-upsample approach inspired by auto-encoders \cite{vincent2010stacked} to learn lower-dimensional latent representations which are then up sampled to image-size prediction maps from which per-pixel classifications can be made. The upsampling procedure introduces more parameters to be tuned during model training, as well as additional computational cost in producing predictions. We again contrast this complex machinery with the simplicity of \methodname\space's approach, which calculates label super-resolution predictions as a weighted sum of activation maps.

\textbf{Conditions where label super-resolution is most easily interpretable} 
 The  linear decomposition of Eq.~\eqref{eq:per_pixel_contribution} holds when using labels that represent the average or sum of values within a grid cell, such as forest cover, elevation, population density, nighttime lights, income, or road length. However, it does not hold exactly when values are transformed nonlinearly after aggregation (e.g. $\log (\sum{y}) \neq \sum{\log (y)}$).\footnote{This issue could be addressed -- in the case of logged variables -- if one obtained a geometric mean image-level outcome rather than an arithmetic mean.} In these cases, the interpretation of label super-resolution estimates requires care. Another case in which the interpretation of the sub-image predictions is difficult is when an image-level characteristic is not directly the sum of sub-image parcels. For instance, when predicting mean housing price in a grid cell, a manicured park might contribute to a higher value, yet that component of the image does not, in itself, have any associated housing price. In this case, we 
 %cannot decompose our image-level outcome as the simple sum of sub-image-level outcomes, unless we 
 we would
 interpret the sub-image predictions as ``contributions to grid cell mean housing price'' (similar to the class activation maps of~\cite{zhou2016learning}) rather than the more natural interpretation as simply ``a finer resolution prediction of housing price.''

     \subsection{Global model}
     \label{sec:global_analysis}
         \begin{figure}[t]
    \centering
    \includegraphics[width=.7\textwidth]{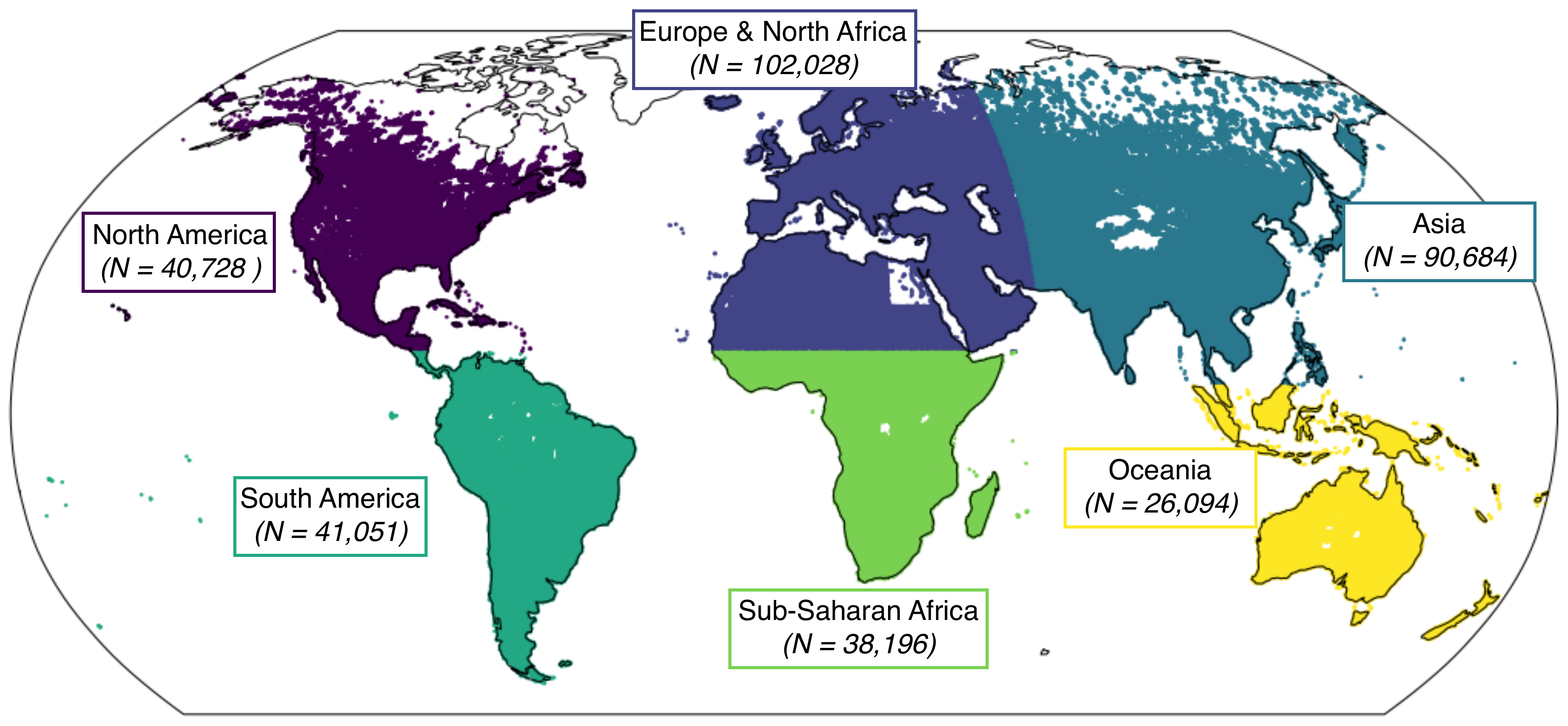}
    \caption{\textbf{Continent samples used to solve four tasks at global scale.} \methodname\space predictions at global scale are generated from six separate cross-validated ridge regressions using random convolutional features. Each continent model is trained on 80\% of the sample size shown ($N$).} 
    \label{fig:continentsplit}
\end{figure}

For our global analysis, we create a global grid, composed of roughly 420 million cells just over 1km$^2$ in size, using an identical structure to that described in Section \ref{sec:grid_and_sampling} for the US. To obtain observations for our global analysis, we sub-sample 1,000,000 cells from this grid, sampling UAR from non-marine grid cells. This relatively sparse sampling of global data is due to the cost of obtaining imagery data.

%Again, we discard any grid cells that fall entirely into the ocean, but maintain grid cells covering inland lakes and seas.

One of the difficulties in sub-sampling from the global grid is that there are many grid cells where 
\iftoggle{arxiv}{we do not have images available}
{no Google imagery is available}
(there are negligibly few missing images in the US grid). After discarding grid cells with missing imagery from our original sample of 1,000,000 observations, we are left with $N = 498,063$ valid observations. 
After removing observations for which labeled data are missing for any of the tasks we analyze at global scale (forest cover, elevation, population density, and nighttime lights),
we are left with $N=423,476$ observations, which we use to train/validate (80\%, $N = 338,781$) and test (20\% $N = 84,692$) the model. % Note: Tamma updated these numbers 12/20/2019, following dimensions in Vaishaal's new feature matrix (N=498,133), after manually removing duplicate features (N=498,063), and then after removing obs with missing labels (N=423,474). 

%We note that there is likely spatial correlation across grid cells in the probability of missing imagery, such that the resulting sample is no longer a truly UAR over space. However, there is no \emph{a priori} means of identifying regions of the world with missing data, limiting our ability to constrain our random sampling to locations without missing imagery. As just (XX \%) of our sampled images were unavailable, it is highly unlikely that this influences our overall model performance. 

When generating features ($K = 2,048$) for our global model, we conduct featurization as described in Section \ref{sec:featurization}. Note that to create the global features we use patches drawn randomly from the \emph{global} sample of images, not just from within the US.

When training the global model, we follow the approach outlined in Sec. \ref{sec:training_testing_model}, solving for grid cell labels as a linear function of the random convolutional features using ridge regression and cross-validation to tune the regularization parameter $\lambda$. However, recovered regression weights are likely to differ across regions of the globe due to heterogeneity in image quality and in visual signal of task labels or their derivatives. Therefore, we divide our global sample into six continental regions before solving each task. The continents (and sample sizes used for training and testing each continent-specific model) are shown in Fig. \ref{fig:continentsplit}.

Modeling heterogeneity using the continents shown in Figure \ref{fig:continentsplit} leads to meaningful gains in performance over ignoring continent effects. As shown in the main text, the approach accounting for heterogeneity generates $R^2$ values of 0.85, 0.45, 0.62, and 0.49, for forest cover, elevation, population density, and nighttime lights, respectively (Fig. \ref{Fig:Global}). In contrast, a global model that pools all observations across the globe and solves for a single linear function of random convolutional features generates $R^2$ values of 0.80, 0.26, 0.48, and 0.41, for the same tasks.

    \subsection{Generalizing to other ACS variables}
    \label{sec:ACS}
        Here we demonstrate the ability of \methodname\ to generalize rapidly across a range of new variables by replicating our primary analysis (i.e. that in Fig. \ref{Fig:MapArray}) for 12 variables from the American Community Survey, the source we use in the main text to measure income across the continental US. This survey is conducted annually across the US, tracking a diverse range of socioeconomic outcomes, from housing information to income and education. For this exercise, we select variables from the ACS that span a range of diverse outcomes and which seem likely to have at least some visible signal in daytime satellite imagery. We report performance for all tested variables. 

We calculate grid cell level labels from census block group level ACS data using the the same method used for ACS income label construction outlined in Sec.~\ref{sec:label_agg}. The resulting labels represent the area-weighted average value of the outcome across the grid cell. The ACS variables we predict are listed and described in Table \ref{tab:ACS}.

% We find that the performance of \methodname\ differs substantially across variables (Fig.~\ref{fig:acs}). \methodname\ explains over 40 percent the variation in building age, house value, number of rooms in a house, household income and per capita income, but explains less than a tenth of the variation in the percent of household income spent on rent. This pattern of predictability shown in Fig.~\ref{fig:acs} is intuitive, with outcomes likely to have obvious visible features, such as age and value of housing, performing well, while outcomes without a clear visible signal, such as the percent of household income dedicated to rent, performing poorly. 
Patterns in performance across tasks could be explained by the hypothesis that some outcomes exhibit more visible features, such as age and value of housing, while other outcomes exhibit less clear visible signal, such as the percent of household income dedicated to rent. 

This exercise shows the ease of generalizing \methodname\ to new contexts. In total, training these twelve predictive models took less than 45 minutes on a workstation with ten cores (IntelXeon CPU E5-263) (Table \ref{table:cost_table}). Given the similarity in performance between \methodname\ and other state of the art approaches documented in Sec.~\ref{sec:cnn_benchmark}, \methodname\ offers a relatively quick and easy way to determine how predictable a variable might be from high resolution visible satellite imagery.

\begin{table}
    \centering \small \vspace{-2cm}
\begin{tabular}{|p{3cm}|p{1.2cm}|p{9.8cm}|}
\hline
  \textbf{Name} & \textbf{Code} & \textbf{Description} \\
 \hline
 Travel time to work & B08303 &  “Travel time (minutes) to work refers to the total number of minutes that it usually took the worker to get from home to work during the reference week. The elapsed time includes time spent waiting for public transportation, picking up passengers in carpools, and time spent in other activities related to getting to work.” \\
 \hline
Percent Bachelor’s Degree & B15003 & Calculated as the number of people over 25 with only bachelor’s degrees (i.e. not masters or doctorate) divided by the total number of people over 25. \\ 
 \hline

Median Household Income & B19013 & Median household income in the past 12 Months (2015 inflation-adjusted dollars) \\
 \hline

Per Capita Income & B19301 & Per capita income in the Past 12 Months (2015 inflation-adjusted dollars) \\
 \hline

Percent below poverty level & C17002 & Calculated as the number of people 15 years or older whose income fell below the poverty level divided by the total number of people 15 years or older. \\
 \hline

Percent food stamp/snap & B22010 & Percent household received food stamps/snap in the past 12 months. \\ 
 \hline

Median income & B25071 &  Gross rent as a percentage of household income in the past 12 months (dollars) \\
 \hline

Number of housing units & B25001 & “A housing unit may be a house, an apartment, a mobile home, a group of rooms or a single room that is occupied (or, if vacant, intended for occupancy) as separate living quarters. Separate living quarters are those in which the occupants live separately from any other individuals in the building and which have direct access from outside the building or through a common hall. Both occupied and vacant housing units are included in the housing unit inventory. Boats, recreational vehicles (RVs), vans, tents, railroad cars, and the like are included only if they are occupied as someone's current place of residence.” \\ 
 \hline

Percent vacant & B25002 & “A housing unit is vacant if no one is living in it at the time of interview.” \\
 \hline

Structure age & B25035 & Data reported is the median year structure built. We calculate structure age as 2015 – median year structure built. \\
 \hline

Number of rooms & B25017 & “For each unit, rooms include living rooms, dining rooms, kitchens, bedrooms, finished recreation rooms, enclosed porches suitable for year-round use, and lodger's rooms. Excluded are strip or pullman kitchens, bathrooms, open porches, balconies, halls or foyers, half-rooms, utility rooms, unfinished attics or basements, or other unfinished space used for storage.” \\
 \hline

Median house value & B25077 & For owner-occupied housing units. \\

\hline

\end{tabular}
\caption{\footnotesize \textbf{Description of variables from the American Community Survey (ACS) used in the analysis (Fig. \ref{Fig:Global}).} Quoted descriptions of variables are from: \text{https://censusreporter.org/topics/table-codes/}.}
\label{tab:ACS}
\end{table}

%\begin{figure}[!ht]
%    \centering
%    \includegraphics[width=.75\textwidth]{Figs%    \caption{\textbf{Generalizing \methodname\ to additional variables from the American Community Survey.} Out-of-sample performance ($R^2$) shown for all tasks in the main text (top panel), as well as 11 additional variables from the American Community Survey (bottom panel), the source used for household income throughout the main text.}
%    \label{fig:acs}
%\end{figure}

 %    \subsection{{Comparison to other Remote Sensing Methods}}
 %    \label{sec:rs_method_comparisons}
 %        \inputt{rs_method_comparisons.tex}

 \section{Comparisons to other models}
    \phantomsection
    Here, we compare the predictive performance and computational cost of \methodname\ to other approaches in the literature. 

     \subsection{Benchmarking performance}
     \label{sec:cnn_benchmark}

\begin{figure}[ht]
        \centering
        \includegraphics[width=.65\textwidth]{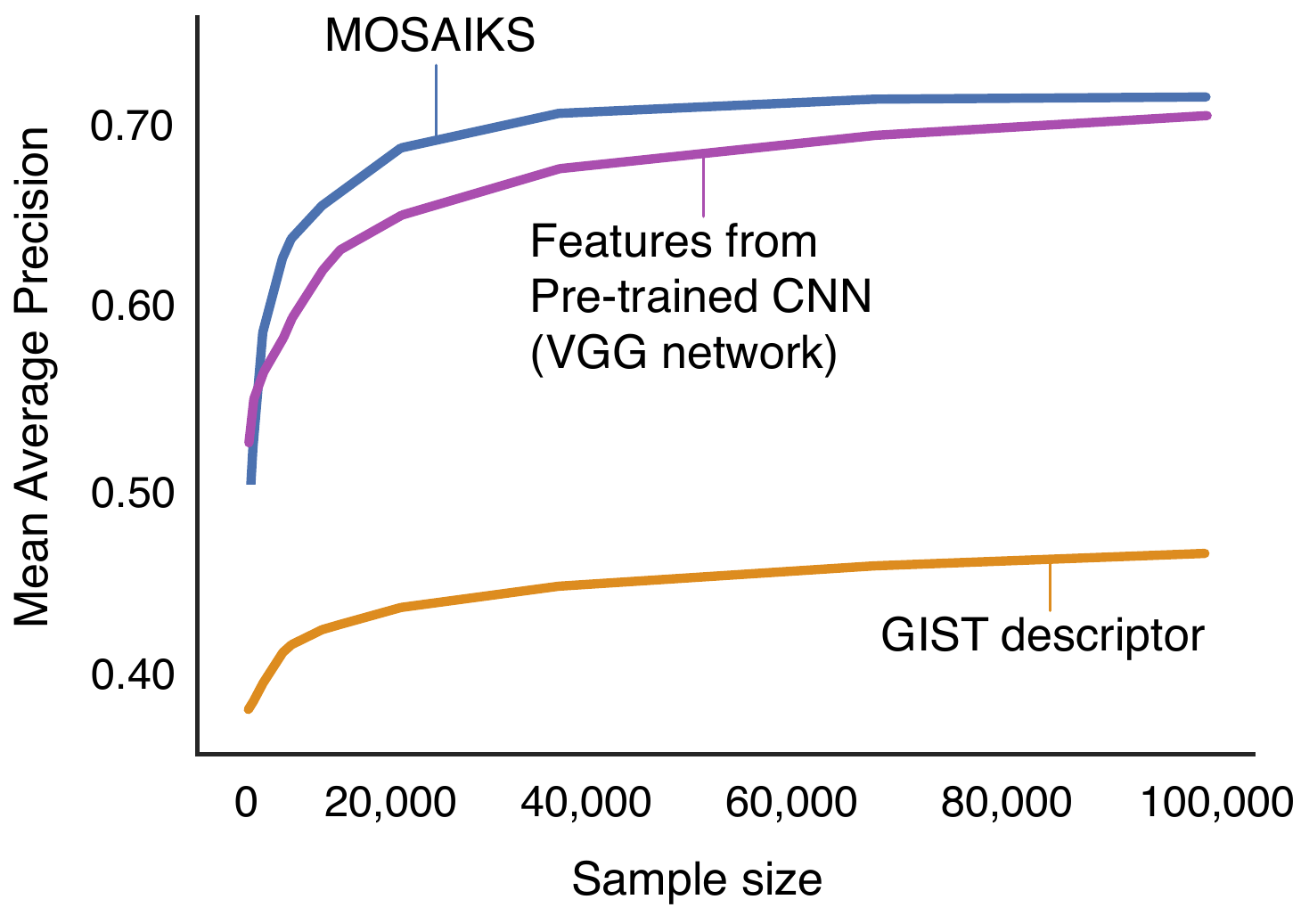}
        \caption{\textbf{Comparison of out-of-sample performance across feature extraction techniques as a function of sample size.} Mean Average Precision is shown for the out-of-sample prediction of housing price class (low, medium, high) for all single-family home sales after 2010 in Arizona, as a function of training sample size. Three feature extraction techniques are compared: \methodname (blue), a pre-trained CNN (VGG, purple), and the GIST descriptor (orange). Figure reproduced from~\cite{Bolliger2017}.
        }
        \label{fig:gist_comp}
\end{figure}

Convolutional neural networks (CNNs) have become the default ``gold standard'' in many image recognition tasks \cite{krizhevsky2012imagenet}, and are increasingly used in remote sensing applications \cite{Jean2016,Volpi2017,Gechter2018a,Perez2017,Robinson2017,penatti2015deep,Jean2019,Zhong2017,Hu2019,Maggiori2017}. Simultaneously, alternative generalizable and computationally efficient pipelines have been developed that incorporate unsupervised featurization and/or a classification or regression algorithm \cite{Cheng2017,Cheriyadat2014,penatti2015deep,Inglada2017,PerezSuay2017}. \methodname\ is low-cost and generalizable like these latter models; however, unlike these other models, it offers accuracies for regression problems competitive with that of leading CNN architectures. Here we quantitatively assess the predictive performance of \methodname\ relative to (a) a CNN trained end-to-end with the outcomes of interest, (b) a similarly cheap, unsupervised featurization used in place of random convolutional features in the \methodname\ infrastructure and (c) a transfer learning approach. For (b), we use the features generated by the last hidden layer of a pre-trained variant of the CNN (trained on natural imagery). This common approach is unsupervised in that the weights of the CNN are not trained using the labels of the outcome of interest, and such an approach has been shown to have better predictive performance than many other unsupervised featurization algorithms (e.g. GIST, SIFT, Bag of Visual Words) on satellite image tasks \cite{Cheng2017}. Previous analyses show through direct comparison that our methodology significantly outperforms ridge regression models using GIST features \cite{Bolliger2017}. Fig~\ref{fig:gist_comp} (reproduced from~\cite{Bolliger2017}) demonstrates this comparison, describing out-of-sample performance for the prediction of housing price class for homes in Arizona.

These exercises compare \methodname\ performance to that of models suited to the data availability of different prediction domains (abundant within the U.S., and relatively scarce at dispersed locations globally). A fine tuned CNN is expected to perform well in the United States, where data are relatively high quality and the sample size is large (nearly a hundred thousand observations); whereas the transfer learning approach is designed to perform well in regions where the training data are more coarse and the sample sizes are smaller (hundreds of observations). The ability of \methodname\ to perform on par with these approaches in each setting demonstrates its generalizability.

\paragraph{Comparison to a deep convolutional neural network and an alternative unsupervised featurization}
First, we compare the performance of \methodname\ to that of a tuned Residual Network (ResNet)\cite{He2016} -- a common, versatile deep network architecture used in recent satellite-based learning tasks \cite{Perez2017}. We train this network \textit{end to end} to predict outcomes in all seven tasks across the continental US, using as input the same imagery used by \methodname.

Specifically, we train an 18-layer variant of the ResNet Architecture pre-trained on ImageNet using stochastic gradient descent to minimize the mean squared error (MSE) between the predictions and labels with an initial learning rate of $0.001$ with a single decay milestone at 10 epochs, and momentum parameter of $0.9$. We train the model for 50 epochs, at which point performance approaches an asymptote. The optimal values for learning rates were tuned on a validation set for the task of predicting population density. We employ a standard train/test split of 80\%/20\%, matching our approach when evaluating \methodname.

Second, we compare \methodname\ performance to a similarly cheap, unsupervised featurization generated by the last hidden layer of a pre-trained variant of the CNN used above, trained on natural imagery. To execute this comparison, we use the features from the last layer of a 152-layer variant of the ResNet Architecture, and then run ridge regression on these features for each task.

Fig.~\ref{Fig:Waffle}A in the main text compares the holdout accuracy of \methodname\ to both alternative approaches, demonstrating that \methodname\ (dark bars) achieves performance competitive with ResNet (middle bars) across all seven tasks, while providing substantially greater performance than ridge regression run on features from the pre-trained CNN (lightest bars).

% \begin{figure}[!ht]
%         \centering
%         \includegraphics[width=.95\textwidth]{Figs_ExtData_SI/cnn_comparison_2panel_clean.pdf}
%         \caption{\textbf{Comparison of test accuracy between \methodname, ResNet, and a regression model using features from a pre-trained CNN.} Panel A shows task-specific performance of \methodname\ (dark bars), in contrast to: an 18-layer variant of the ResNet Architecture (ResNet-18) trained end-to-end for each task (middle bars); and an unsupervised featurization that uses the last hidden layer of a 152-layer ResNet variant that was pre-trained using natural imagery in combination with ridge regression (lightest bars). Panel B shows the performance of ResNet-18 by task and training epoch, demonstrating that all tasks reached an asymptote after 400 epochs. Dark lines indicate the cumulative maximum performance by epoch, while light lines indicate epoch-specific performance. %In all results shown, forest cover, elevation, and population density tasks use a uniform at random sample of training and test set images, while nighttime lights, road length, income, and housing price use a population weighted sample.
%         }
%         \label{fig:cnn_comparison}
% \end{figure}

\begin{figure}[!ht]
        \centering
        \includegraphics[height=.9\textwidth]{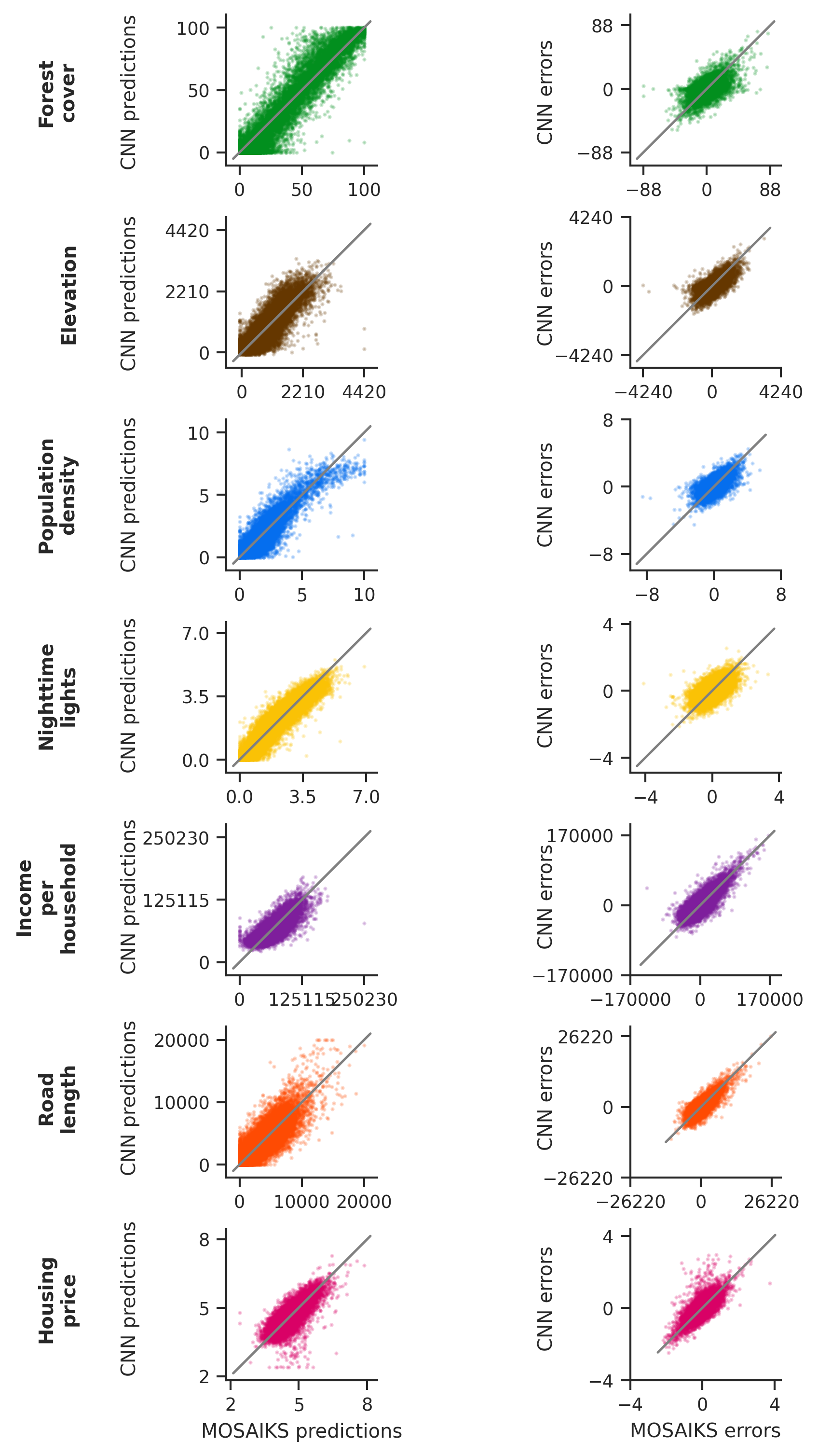}
        \caption{\textbf{Comparison of predictions and prediction errors between \methodname\ and the ResNet-18 CNN.} The left column shows the relationship between predictions generated by \methodname\ ($x$-axis) and predictions generated by the ResNet-18 CNN ($y$-axis). The right column shows the relationship between prediction errors from \methodname\ ($x$-axis) and prediction errors from the CNN ($y$-axis). In both plots, each point indicates one grid cell ($\sim$1km$\times$1km) in the holdout test set; the test set sample size is approximately 20,000 for each task, although sample sizes vary somewhat due to data availability across tasks (Section \ref{sec:training_testing_model}). 
        }
        \label{fig:cnn_comparison}
\end{figure}

\paragraph{Comparison to a transfer learning approach}

We also compare the performance of MOSAIKS to that of a transfer learning approach in which nighttime lights observations are used to tune a CNN that was pre-trained on ImageNet. The tuned CNN is then used to extract features from the satellite images and a linear model is trained to predict the outcome of interest. This approach leverages a large number of nighttime lights observations to better learn how to extract information  from satellite imagery that is meaningful to tasks that may be reflected in nighttime lights (e.g. wealth). Comparing the performance of \methodname\ to that of transfer learning tests the value of learning these features from nighttime lights, relative to the unsupervised featurization of \methodname.

We compare the performance of MOSAIKS to the transfer learning approach by replicating a subset of the analyses in \cite{Jean2016} and \cite{Head2017}. Using \methodname, we predict wealth, electricity, mobile phone ownership, education, bed net count, female body mass index, water access, and hemoglobin level in Haiti, Nepal and Rwanda; we additionally predict child weight percentile, child height percentile and child weight for height percentile in Rwanda. These variables are recorded at geo-located ``cluster'' locations by the Demographic and Health Survey (DHS); the survey methodology is detailed in \cite{Jean2016}. The variables and countries we provide performance metrics for were chosen based on the facility of obtaining and matching images and labels from the original authors and their replication code bases. We report performance for all tested variables and countries.

In this analysis, we use two \methodname-based models to predict the DHS cluster labels. First, we use only the \methodname\ random convolutional features (indicated as RCF), which we calculate for each image as detailed in Sec.~\ref{sec:featurization}, and then average over the 100 images associated with each DHS cluster (see \cite{Jean2016} and \cite{Head2017} for details on matching images to clusters; we use the same matching approach as the original authors). In a second model, denoted MOSAIKS-NL below, we use the \methodname\ random convolutional features along with features based on nighttime lights. The nighttime light features for each cluster are counts of the number of nightlight values that fall within a set of 19 bins, as well as the minimum, mean and maximum of the values within the image. Bins were evenly spaced on a log scale from a luminosity of 0.1 to 500 (in units of nanoWatts/cm$^2$/sr). We average nighttime light features for all 100 images associated with each DHS cluster, as we do for RCF.

We show results for the MOSAIKS-nighttime lights model for two reasons. First, it presents the most fair comparison to the transfer learning approach, which also leverages both nighttime lights and daytime imagery. Second, it demonstrates the ability of \methodname\ to seamlessly combine information from different sensors – by appending their features in a linear model – to make predictions. 

\paragraph{Training a model that uses features from multiple sensors}

\begin{figure}[!ht]
        \centering
        \includegraphics[width=.95\textwidth]{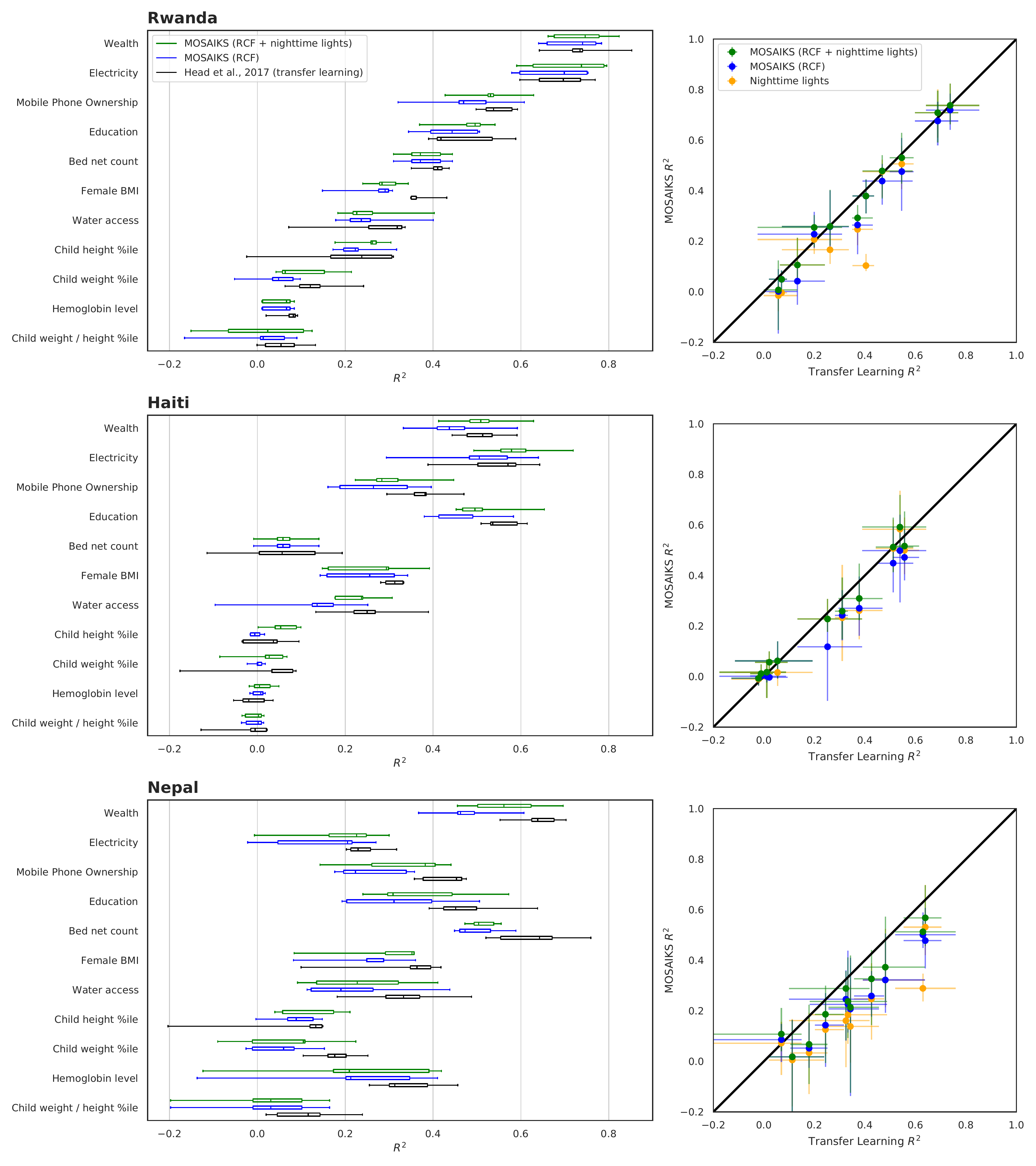}
        \caption{\footnotesize \textbf{Comparison of accuracy between \methodname\ and a transfer learning model.} Box plots (left) show task-specific performance of \methodname\ models (RCF in blue and RCF + nighttime lights in green) in contrast to a transfer learning model (black). Box and whiskers show the performance over the 5 cross-validation folds. Scatter plots (right) show the performance of \methodname\  models, as well as a nighttime lights-only model (orange) versus the transfer learning model performance. Each point in the scatter is the average $R^2$ over the 5 cross-validation folds, while whiskers indicate the full range of performance across folds.
        }
        \label{fig:headRep}
\end{figure}

A key benefit of the \methodname\ approach is that it can easily combine information from multiple sensors. Recall that to train a model when using only the RCF from visual imagery we regress the outcome $y_{\ell}^s$ for each task $s$ on features $\Xell$ as follows: 
\begin{align*}
    y_{\ell}^s = \XIell \bm\beta^s + \epsilon_\ell^s 
\end{align*}
And solve for $\bm\beta^s$ by minimizing the sum of squared errors plus an $l_2$ regularization term: 
\begin{equation*}
\underset{\bm\beta^s}{\min} \frac{1}{2}|| y_\ell^s - \XIell \bm\beta^s ||^2_2 + \frac{\lambda^s}{2}||\bm\beta^s||^2_2 
\end{equation*} 
To include features from an additional sensor, $S_\ell$, such as nighttime lights, one simply generates a new set of features, $\mathbf{z}(\mathbf{S}_\ell)$, -- using the RCF algorithm or any other unsupervised featurization approach -- and includes the features in the regression model, giving: 
\begin{align*}
    y_{\ell}^s = \XIell \bm\beta^s + \mathbf{z}(\mathbf{S}_\ell) \bm\gamma^s +  \epsilon_\ell^s 
\end{align*}
Then, one solves for $\bm\beta^s$ and $\bm\gamma^s$ by minimizing the sum of squared errors plus individual regularization terms for each sensor: 
\begin{equation*}
\underset{\bm\beta^s, \gamma^s}{\min} \frac{1}{2}|| y_\ell^s - \XIell \bm\beta^s - \mathbf{z}(\mathbf{S}_\ell) \bm\gamma^s ||^2_2 + \frac{\lambda_1^s}{2}||\bm\beta^s||^2_2 + \frac{\lambda_2^s}{2}||\bm\gamma^s||^2_2
\end{equation*} 
Regularizing the features from each sensor separately enables the model to treat features from individual sensors differently, which we found improves model performance. We implement a model that combines RCF and features from nighttime lights in Fig.~\ref{fig:headRep}. Features from additional sensors could be added to the model in a similar way.

\paragraph{Interpretation of test accuracy comparisons}
Note that the performance of these models represents a reasonable lower bound on potential performance; some task-specific enhancements could be used to improve predictive power for each of these methods. For example, more layers could be added to ResNet or alternative architectures could be tested for specific tasks. In the case of \methodname\ and the pre-trained ResNet features, more flexible regression models could be used to estimate the relationship between features and labels, such as increasing $K$, using a nonlinear model, or leveraging a hurdle model in tasks with a large number of zero observations. While these task-specific changes may marginally improve performance of any of these approaches, prior research on similar image recognition tasks suggests further gains for the ResNet are likely to be minimal \cite{Zoph2016NeuralLearning}.
While the similarity of performance in Fig.~\ref{Fig:Waffle}A is perhaps surprising, it is also encouraging for further research. This comparison suggests that wide, shallow networks using local-level features (analogous to random convolutional features) are as descriptive as more complex, highly optimized CNN architectures for satellite remote sensing, across many tasks.

The similarity in performance of \methodname\ and ResNet across tasks (Fig.~\ref{Fig:Waffle}A) is consistent with a hypothesis that both approaches are reaching the limit of information that is provided by satellite imagery for predicting the outcomes we test. Fig. \ref{fig:cnn_comparison} provides further evidence of such a hypothesis, as the test set predictions and prediction errors are highly correlated across ResNet and \methodname\ in all tasks studied. A human prediction baseline has not been established, but could provide insight into whether there is substantial room for improvement in skill for each of these tasks, although we suspect many of these tasks will be difficult for non-expert humans (e.g. nightlights or house prices).
%In general it is difficult to assert that any method is extracting the maximum information available from the a data source, but similarity in performance across tasks indicates that there may in fact be limits as to the predictability of certain tasks via features based on satellite imagery alone.  As SIML becomes more widely used for remote sensing, understanding the nature of observability of certain tasks will be increasingly important.

%\textbf{Comparison to a leading unsupervised featurization algorithm}

     \subsection{Comparing costs}
     \label{sec:cost_analysis}
     %\inputt{cost_analysis_new.tex}
     
%\newpage\clearpage
%\vfill
\begin{table}[t]
\centering
\begin{tabular}{lcc}
     & \textbf{ResNet} & \textbf{\methodname} \\
 \textbf{\textit{Component}} & Time (GPU)  &  Time \\
 \hline
Training set featurization ($N=80$k) & \multirow{4}*{$\sim$ \bf{7.9 hours}}  & $\sim$ 1.2 hours (GPU) \\
\cline{3-3}
Model training  &&  $\sim$ {2.8 seconds (GPU)}  \\
 && $\sim$ {50 seconds (10 cores)} \\
 && $\sim$ \bf{1.8 minutes (laptop)} \\
\cline{2-2} \cline{3-3}
Holdout set featurization ($N=20$k) & \multirow{4}*{$\sim$ \bf{40 seconds}} &$\sim${18 minutes (GPU)} \\
\cline{3-3}
Holdout set prediction& &$<$ {0.01 seconds (GPU)} \\
 && $\sim$ {0.1 seconds (10 cores)} \\
 && $\sim$ \bf{0.7 seconds (laptop)} \\
\hline
Total cost to ecosystem & {$\sim$ \bf{7.9 hours}} & $\sim$ 1.5 hours (GPU) \\
\cline{2-2} \cline{3-3}
Total cost to user& \bf{{$\sim$ \bf{7.9 hours}}} & $\sim$ {2.8 seconds (GPU)} \\
 && $\sim$ {50.1 seconds (10 cores)} \\
 && $\sim$ \bf{1.8 minutes (laptop)} \\
\end{tabular}
  \caption{ 
   \label{table:cost_table} 
   \textbf{Wall-clock times of components of \methodname\ compared with a fine-tuned CNN.} Bold times are those that a practitioner using each method would incur (assuming \methodname\ users have access to a standard laptop only). Model training time includes training \emph{after} tuning for a single task for both ResNet and \methodname. \methodname\ was run using $K$=8,192 features. ResNet operations were run on an Amazon EC2 p3.2xlarge instance with a Tesla V100 GPU and 60GB of onboard RAM. Cost of computation on this machine is roughly $\$3/hr$. \methodname\ operations are shown for runs on this same GPU, a local workstation with ten cores (Intel Xeon CPU E5-263), and a standard laptop (MacBook Pro).} 
\end{table}
%\vfill
% % \begin{table}[H]
% \centering
% \begin{tabular}{lcc}
%      & \textbf{ResNet} & \textbf{\methodname} \\
%  \textbf{\textit{Component}} & Time  &  Time \\
%  \hline
% Training set featurization ($N=80$k) & \multirow{2}*{$\sim$ \bf{2.7  days}}  & $\sim$ 1.2 hours \\
% \cline{3-3}
% Model training  &&  $\sim$ \bf{3.5 minutes}  \\
% \cline{2-2} \cline{3-3}
% Holdout set featurization ($N=20$k) & \multirow{2}*{$\sim$ \bf{40 seconds}} &$\sim${18 minutes} \\
% Holdout set prediction& &\bf{0.1 seconds} \\
% \hline
% Total cost to ecosystem & {$\sim$ \bf{2.7 days}} & $\sim$ 1.6 hours \\
% Total cost to user& \bf{{$\sim$ \bf{2.7  days}}} & \bf{$\sim$ 3.6 minutes} \\
% \end{tabular}
%   \caption{ 
%   \label{table:cost_table} 
%   \textbf{Wall-clock times of components of \methodname\ compared with a fine-tuned CNN.} Bold times are those that a practitioner using each method would incur. Model training time includes training after tuning for a single task, both for ResNet and for \methodname. \methodname\ times use the full $K=8,192$ features. ResNet operations were run on an Amazon EC2 p3.xlarge instance with a Tesla V100 GPU and 16GB of onboard RAM. Cost of computation on this machine is roughly $\$3/hr$. \methodname\ operations were run on a 10-core CPU with 95GB of RAM.} 
% \end{table}

In practice, high computational costs can limit the use of SIML methods -- especially when resources are scarce, such as in government agencies of low-income countries \cite{Haack2016} or research teams and NGOs with limited budgets. Specifically designed to address this challenge, \methodname\ scales across many research tasks
by decoupling featurization from task selection, model-fitting, and prediction. The computationally costly step of featurization is done centrally on a fast computer with a graphics processing unit (GPU); individual practitioners need only download the pre-computed features, merge on labels for the task they select, and run regressions. Because features are created and stored by a central entity, the research community makes use of a cached set of computations, reducing the overall computational burden of widespread SIML and any external social costs generated by these computations \cite{strubell2019energy}. Additionally, this decoupling of task-agnostic computations from task-specific computations allows practitioners to run more diagnostic analyses on their tasks, such as those presented in Fig.~\ref{Fig:Waffle} of the main text.

From the perspective of a user who can access pre-computed \methodname\ features to train and validate a new task, we find that \methodname\ is $\sim250\times$ to $10,000\times$ faster than a state-of-the-art neural net architecture (ResNet), depending on the computational resources available to a \methodname\ user (Table~\ref{table:cost_table}). Moreover, \methodname\ performance is competitive with the ResNet on all tasks we have studied (Fig \ref{Fig:Waffle}A). From the perspective of the entire computational ecosystem, which bears the cost of image featurization in addition to model training and testing, we find that \methodname\ is $5.3 \times$ faster than the ResNet when solving a single task. The relative efficiency of \methodname\ grows with the number of tasks studied because \methodname\ features can be reused across tasks. 
For the ResNet, the times in Table~\ref{table:cost_table} reflect our wall-clock time on a single Amazon EC2 instance for a single task, so that the time costs are similar to that of introducing a single new domain \textit{ex post}. For \methodname, Table~\ref{table:cost_table} includes wall-clock times on three different computational platforms, as users may have access to different resources. We show times using the same GPU as we use for the ResNet comparisons, times on a local workstation with ten cores (Intel Xeon CPU E5-263), and times on a standard laptop (MacBook Pro). For both ResNet and \methodname, we report in Table \ref{table:cost_table} model training time \emph{after} using cross-validation to select optimal hyperparameters. For \methodname, model training time on the local workstation with 10 cores is $\sim$6.8 minutes when including cross-validation to select penalization parameters in ridge regression. 
The ecosystem-wide costs of featurization per task shown in Table \ref{table:cost_table}  decline as \methodname\ becomes more widely adopted, because features can be cached centrally and distributed without modification to multiple users who are training and/or testing SIML in common locations.

We considered only one CNN architecture, which we chose because of its use in previous remote sensing applications~\cite{Jean2016}. We did not attempt to innovate in neural net architectural design or algorithms. While one could pursue targeted innovations in neural networks for remote sensing, such as in ref. \cite{Zhong2017}, we emphasize that our method is currently orders of magnitude faster for the user than off-the-shelf fine-tuned CNN methods (Table~\ref{table:cost_table}), does not require a GPU for prediction, and achieves competitive prediction performance (Fig.~\ref{Fig:Waffle}A). There is recent work that aims to train networks to learn a ``common representation" that can generalize across tasks, but this is a subject of ongoing research~\cite{Ruder2017}, requires the tasks to be known in advance, and has yet to be demonstrated or evaluated at scale. %Future work might investigate targeted deep models to improve prediction accuracy across tasks at reasonable speeds that make them accessible to a diverse base of researchers. -- I don't think we need this. There's a lot that could be done in the future - JP

%\footnote{Table \ref{table:cost_table} shows wall-clock times for ResNet and \methodname\ on a GPU. However, future users of \methodname\ are more likely to train and test their models on a standard laptop. We find that the total cost to the user of training and testing a new task on a standard laptop is approximately 6 minutes, less than 2$\times$ the value shown for a GPU.}

 %    \subsection{Production-Scale Implementation and API} 
 %    \label{sec:API}
 %        \inputt{api.tex}

% %\printbibliography[resetnumbers=false]
 %\end{refsection}

 %\section{Todos}
 %\listoftodos

%%%%%%%%%%%%%%%%%%%%%%%%%%%%%%%%%%%%%%%%%%%%%%%%%%%%%%%%%%%%%%%%%%
                % Extended Data: SUPP FIGURES %
%%%%%%%%%%%%%%%%%%%%%%%%%%%%%%%%%%%%%%%%%%%%%%%%%%%%%%%%%%%%%%%%%%
%  \pagebreak
% %\clearpage
% \section*{Extended Data}
% % commented this out so it compiles b/c there are problems with the figure refs

% \inputt{supplementary_figures.tex}

% \clearpage
%\end{document}

\end{document}